\documentclass[1p,preprint]{elsarticle}
\usepackage{url}  %Required
\usepackage{graphicx}  %Required
\usepackage{latexsym} 
\usepackage{multirow}
\usepackage{graphicx}
\usepackage{wrapfig}
\usepackage{amsmath}
\usepackage{mathtools}
\usepackage{xspace}
\usepackage[pdftex]{color}
\usepackage{floatflt}
\usepackage{graphicx}
\usepackage{pict2e}
\usepackage{amsthm}

 \author[1]{Dalal Alrajeh}
 \ead{dalal.alrajeh@ic.ac.uk}

 \author[2]{Hana Chockler}
 \ead{hana.chockler@kcl.ac.uk}  

 \author[3]{Joseph Y.~Halpern}
 \ead{halpern@cs.cornell.edu}  
 
\address[1]{Department of Computing,  Imperial College London,  UK}
\address[2]{Department of Informatics, King's College London, UK}
\address[3]{Computer Science Department, Cornell University, USA}

\newcommand{\Scal}{{\cal S}}

\renewcommand{\phi}{{\varphi}}

\newcommand{\BT}{\mbox{{\it BT}}}
\newcommand{\BS}{\mbox{{\it BS}}}
\newcommand{\ST}{\mbox{{\it ST}}}

\newcommand{\BSs}{{\it BS}}

\newcommand{\CA}{\mbox{{\it C}}}
\newcommand{\DV}{\mbox{{\it DV}}}
\newcommand{\FS}{\mbox{{\it FS}}}
\newcommand{\IS}{\mbox{{\it IS}}}
\newcommand{\RV}{\mbox{{\it RV}}}
\newcommand{\RA}{\mbox{{\it A}}}

\newcommand{\AP}{\mbox{{\it AP}}}
\newcommand{\PLS}{\mbox{{\it PLS}}}

\newcommand{\fppar}{$\mbox{FP}^{{\rm NP}}_{||}$}
\newcommand{\ppar}{$\mbox{P}^{{\rm NP}}_{||}$}
\newcommand{\sPD}{\mbox{{\scriptsize{\it PD}}}}

\newcommand{\sIL}{\mbox{{\scriptsize{{\it IL}}}}}

\newcommand{\sTT}{\mbox{{\scriptsize{{\it AA}}}}}

\newcommand{\sAS}{\mbox{{\scriptsize{{\it AS}}}}}

\newcommand{\sDE}{\mbox{{\scriptsize{{\it D}}}}}
\newcommand{\sMC}{\mbox{{\scriptsize{{\it C}}}}}
\newcommand{\sUW}{\mbox{{\scriptsize{{\it W}}}}}
\newcommand{\sFM}{\mbox{{\scriptsize{{\it FM}}}}}
\newcommand{\sLM}{\mbox{{\scriptsize{{\it LM}}}}}
\newcommand{\sGC}{\mbox{{\scriptsize{{\it CG}}}}}
\newcommand{\sCM}{\mbox{{\scriptsize{{\it CM}}}}}
\newcommand{\sAM}{\mbox{{\scriptsize{{\it AM}}}}}
\newcommand{\sSC}{\mbox{{\scriptsize{{\it SC}}}}}
\newcommand{\sRG}{\mbox{{\scriptsize{{\it RG}}}}}
\newcommand{\sCB}{\mbox{{\scriptsize{{\it CB}}}}}

\newcommand{\PD}{\mbox{{\it PD}}}

\newcommand{\IL}{\mbox{{\it IL}}}

\newcommand{\TT}{\mbox{{\it AA}}}
\newcommand{\DP}{\mbox{{\it D}}}

\newcommand{\DO}{\mbox{{\it PD}}}
\newcommand{\AS}{\mbox{{\it AS}}}

\newcommand{\DE}{\mbox{{\it D}}}
\newcommand{\MC}{\mbox{{\it C}}}
\newcommand{\UW}{\mbox{{\it W}}}
\newcommand{\FM}{\mbox{{\it FM}}}
\newcommand{\LM}{\mbox{{\it LM}}}
\newcommand{\GC}{\mbox{{\it CG}}}

\newcommand{\CM}{\mbox{{\it CM}}}
\newcommand{\AM}{\mbox{{\it AM}}}
\newcommand{\SC}{\mbox{{\it SC}}}
\newcommand{\RG}{\mbox{{\it RG}}}
\newcommand{\CB}{\mbox{{\it CB}}}

\newcommand{\bd}{\begin{definition}}
\newcommand{\ed}{\end{definition}}
\newcommand{\be}{\begin{enumerate}}
\newcommand{\bi}{\begin{itemize}}
\newcommand{\ee}{\end{enumerate}}
\newcommand{\ei}{\end{itemize}}

\newcommand{\U}{{\cal U}}

\newcommand{\V}{{\cal V}}
\newcommand{\R}{{\cal R}}

\newcommand{\stam}[1]{}
\newcommand{\bft}{{\bf true}\xspace}

\newcommand{\commentout}[1]{}

\newcommand{\fullv}[1]{#1}
\newcommand{\shortv}[1]{\commentout{#1}}

\newtheorem{theorem}{Theorem}[section]
\newtheorem{definition}[theorem]{Definition}

\newtheorem{lemma}[theorem]{Lemma}
\newtheorem{proposition}[theorem]{Proposition}
\newtheorem{example}[theorem]{Example}

\newtheorem{claim}{Claim}

\newcommand{\bbox}{\vrule height7pt width4pt depth1pt}

\newcommand{\clm}{\begin{claim}}
\newcommand{\eclm}{\end{claim}}
\newcommand{\union}{\cup}
\newcommand{\M}{{\cal M}}

\newcommand{\ra}{\rangle}
\newcommand{\<}{\langle}
\newcommand{\thm}{\begin{theorem}}
\newcommand{\pro}{\begin{proposition}}
\newcommand{\ethm}{\end{theorem}}
\newcommand{\prf}{\noindent{\bf Proof.} }

\definecolor{darkred}{rgb}{0.65,0,0}

\newcommand{\SCICH}{\textit{SCICH}}

\newcommand{\F}{{\cal F}}

\newcommand{\satt}{\models}

\newcommand{\Pa}{\mathit{Par}}
\newcommand{\Compat}{\mathit{Compat}}
\newcommand{\lem}{\begin{lemma}}
\newcommand{\elem}{\end{lemma}}
\newcommand{\epro}{\end{proposition}}

\newcommand{\dfn}{\begin{definition}}
\newcommand{\edfn}{\end{definition}}

\newcommand{\eprf}{\bbox\fullv{\vspace{0.1in}}}

\newcommand{\xam}{\begin{example}}
\newcommand{\exam}{\end{example}}

\begin{document}
%
%dalal22
 \title{Combining Experts' Causal Judgments\tnoteref{t1}} \tnotetext[t1]{A preliminary version of the paper appeared in
    \emph{Proceedings of the Thirty-Second AAAI Conference on
      Artificial Intelligence (AAAI-18)}.}

%dalal22
%joe32: in what package is this macro defined?  I cut it for now,
%since it causes me problems
%\newpageafter{Combining Experts' Causal Judgments}

\begin{abstract}
  Consider a policymaker who wants to decide which intervention to
perform in order to change a currently undesirable situation.
The policymaker has at her disposal a team of experts, each with their 
own understanding of the causal dependencies between different factors
contributing to the outcome. The policymaker has varying degrees of
confidence in the experts' opinions. She wants
to combine their opinions in order to decide on the most effective
intervention. 
We formally define the notion of an effective intervention, and then
consider how experts' causal judgments can be combined in order to
determine the most effective intervention.  
We define a notion of two causal models being \emph{compatible}, and
%dalal22
%joe32: while I'm happy to use ``merge/merged'' instead of
%``combine/combined''; but then we need to be consistent throughout
%the paper.  There are about 50 occurrences of
%``combine/combined/combining''.  For now, I changed this back.
%dalal23
%show how compatible causal models can be combined.  We  then use
show how compatible causal models can be merged.  We  then use
%show how compatible causal models can be merged.  We  then use
%joe33
%it as the basis for combining experts causal judgments.
it as the basis for combining experts' causal judgments.
%dalal20
We also provide definitions of partial compatibility and causal model 
decomposition to cater for cases when models are incompatible. 
%dalal:6 remove notion of case study
%We illustrate our approach on two case studies.
We illustrate our approach on a number of real-life examples.

\end{abstract}
\maketitle
\section{Introduction}

%joe1: complete new introduction.  I don't think that we were telling
%the right story. See below for comments on the old introduction.  
%joe3
%Consider a policymaker who is trying to decide which intervention(s),
Consider a policymaker who is trying to decide which intervention,
%joe3: added
that is, which actions,
should be implemented in order to bring about a desired outcome,
%joe3: you're not supposed to abbreviate e.g. or i.e. in the main
%text; only in parens, according to standard tyle manual
%e.g., 
%%.  For example, she may want to implement interventions that would 
%prevent violent behavior in prisons or  reduce famine mortality in
such as preventing violent behavior in prisons or reducing famine mortality in
some country.   
%hana added text about interventions
% dalal: breaks the flow. Suggest including later as our definition of
% intervention
%joe3: I don't think this really adds, and the discussion of causality
%will only confuse people who don't know the definition.  Cut.
%An intervention in this context is an action or a set of actions that 
%would bring 
%about a desired outcome. In other words, we are interested in actions
%that \emph{change} 
%the current situation; in contrast to the concept of structural
%causality, we are not 
%concerned with actions that contribute to the outcome, but need a
%``reinforcement'' in order to change the final outcome.
%
%We assume that the policymaker has access to various experts who can
% dalal
The policymaker has access to various experts who can
%joe3
%advise her.  Some  perhaps experts may be (in the policymaker's view)
advise her on which interventions to consider.  Some  experts may be (in the policymaker's view)
more reliable than others; they may also have different areas of
expertise;
or may have perceived alternative factors in their analysis.  
%hana10
The goal of the policymaker is to choose the best intervention, taking into account
the experts' advice. 
%The policymaker must somehow combine these expert's advice
%to reach her decision.
%
%dalal: should add here questions illustrating the difficulty as at the moment
% an immediate answer would be pick the most reliable one which is not 
% what we're arguing for
%joe7: I hope that the next paragraph illustrates some of the issues;
%I'm not sure what we can put here

There has been a great deal of work on combining experts'
%joe20
%probabilistic judgments.  (Genest and Zidek \citet{GZ}
%joe32: I made lots of changes to deal with Elsevier's use of \cite
%and \citet
%probabilistic judgments.  (Genest and Zidek \citeyear{GZ}
probabilistic judgments.  (\citet{GZ}
provide a somewhat dated but still useful overview; 
%joe3
%Dawid~\shortcite{Daw87} and Fenton et~al.~\citet{FNB16}, among others,
%joe20: \shortcite => citeyear
%joe32
%Dawid~\citeyear{Daw87} and Fenton et~al.~\citeyear{FNB16}, among others,
\citet{Daw87} and \citet{FNB16}, among others,
%joe2: it's not really that relevant
%give a Bayesian analysis that will be relevant to our work.)  However,
give a Bayesian analysis.)  
%hana interventions again
%joe12: line shaving 
%In this work, however, we are interested in combining experts'
We are interested in combining experts'
%joe3
%judgment specifically in the context of deciding on the best
%intervention. Hence, 
judgments in order to decide on the best intervention. Hence,
%However,
%since we are interested in \emph{interventions}, 
%dalal2:
we need more than
probabilities.  We need to have a causal understanding of the
situation.  Thus, we assume that the experts provide the policymaker
with \emph{causal models}. 
%(we review the definition of causal models in Section~\ref{sec:background}).  
In general, these models may
involve different variables (since the experts may be focusing on
different aspects of the problem).  Even if two models both include
variables $C_1$ and $C_2$, they may disagree on the relationships
between them.  For example, one expert may believe that $C_2$ is
independent of $C_1$ while another may believe that $C_1$ causally
depends on $C_2$.
Yet somehow the policymaker
%joe12: we don't actually combine the models if there's an
%inconsistency, so this is misleading
%must combine (the information in) these causal models to reach her
%decision.  How should she do this?
wants to use the information in these causal models to reach her
decision.  

%hana8 rewrote the paragraph, adding more references; original below
%joe10: reinserted original sentence and did further rewriting
Despite the clear need for causal reasoning, and the examples in the
% dalal8: two close uses of clear
%literature and in practice where experts are clearly working with
literature and in practice where experts work with
%dalal8:
%causal models (see, e.g., \citet{CFKL15,Sampson:2013}),
%joe32
%causal models (e.g., \citet{CFKL15,Sampson:2013}),
causal models (e.g., \cite{CFKL15,Sampson:2013}),
there is surprisingly little work on combining causal judgments.
Indeed, the only work that we are aware of
%joe20
that preceded our work 
is that of
%joe14
%Bradley, Dietrich, and List \shortcite{BDL14},
%joe32
%Bradley, Dietrich, and List \citeyear{BDL14} (BDL from now on),
\citet*{BDL14} (BDL from now on),
who prove an impossibility result.
Specifically, they describe certain arguably reasonable desiderata,
%dalal23 I assume here again that they go straight into (not)combining
% if I am mistaken then there are a couple of places where the term
% combined when discussing BDL should be replaced with merged
%joe33: for consistency, I think we should say ``merge'' here, since
%in our terminology, we merge models and combine opinions
%and show that there is no way of combining causal models so as
and show that there is no way of merging causal models so as
to satisfy all their desiderata.  They then discuss various weakenings
of their assumptions to see the extent to which the impossibility can
be avoided, none of which seem that satisfactory.
%hana17 added the arxiv paper
%joe20: moved this up two paragraphs, since it's more closely related
%to BDL, not causal discovery.
%A recent work by Zennaro and Ivanovska examines the problem of
%\emph{fairness} of the aggregated model expressing 
%the opinions of different experts on data.
Following the conference version of our paper,
Zennaro and Ivanovska \citeyear{ZI18} examined the problem of
%dalal23 similarly here I didnt replace this with merge
%joe33: but I did :-)
%combining causal models where the combined model must satisfy a
merging causal models where the merged model must satisfy a
%joe32
%fairness requirement (although the individual expert's model may not
fairness requirement (although the individual experts' models may not
be fair).  They proposed a way of combining models based on ideas of BDL.
%joe20: added my paper with Meir Friedenberg, which received positive
%reviews from KR, so may end up there
Friedenberg and Halpern \citeyear{FH18} also considered the same
problem of combining causal model of experts, but allowed for the
possibility that experts disagree on the causal structure of
variables due to having different focus areas.

%joe10: new material.  I don't think we should use up valuable real
%estate in the introduction to go into great deails.  Also, there are
%many more papers than these three.
%dalal8:
%There is also a great of work on a closely related problem
%joe12
%There is also much  work on a closely related problem
There is also much  work on the closely related problem
of \emph{causal discovery}: constructing a single causal model 
from a data set.
%joe11
%These causal models have been constructed using a variety of
%techniques to find the model that
A variety of techniques have been used to find the model that
best describes how the data is generated (see,
%joe12: ``e.g.'' inside parens; ``for example'' outside
%for example, \citet{CH10,CH12,HEJ14,TS11,TT15}; Triantafillou andn
%joe32
%e.g., \citet{CH10,CH12,HEJ14,TS11,TT15}; Triantafillou and
%Tsamardinos \citeyear{TT15} provide a good overview of work in the
e.g., \cite{CH10,CH12,HEJ14,TS11,TT15}; 
\citet{TT15} provide a good overview of work in the
area).

Of course, if we have the data that the experts used to generate their
models, then we should apply the more refined techniques of the work
%joe15
%on causal discovery.   However, while the causal model constructed by
on causal discovery.   However, while the causals model constructed by
experts are presumably based on data, the data itself is typically no
%joe12
%longer available.  The models represent the distillation
longer available.  Rather, the models represent the distillation
of years of experience, obtained by querying the experts.
%We assume that the causal models are given to us by experts and
%reflect the experts'  knowledge and opinions as well as the
%datasets. This setting is especially relevant 
%in situations where we do not have sufficient data to generate causal models
%automatically; rather, the causal models provided by experts are constructed
%based on their expertise.

%joe10: our ``series'' has length two ...
%We provide a series of gradually weaker conditions under which
%two causal models $M_1$ and $M_2$ are \emph{compatible}.
%dalal11: I think we should a sentence about what we provide, something like the following
In this paper, we present an approach to combining experts' causal models 
when sufficient data for discovering the overall causal model is not available. 
The key step in combining experts' causal models lies in defining when
two causal models are \emph{compatible}.  Causal models can be
%dalal23
%combined only if they are compatible.
merged only if they are compatible.
%dalal15:
%joe14: this seems like a non sequitur.  It has nothing to do with
%comes before or after.  We don't even mention deterministic models here.
%I was happy with your later comment on deterministic models
%(Our focus is on deterministic causal models.)
%joe12
%We start with the notion of strong compatibility, where the conditions say,
We start with a notion of \emph{strong} compatibility, where the conditions say,
among other things, that if both $M_1$ and $M_2$
involve variables $C_1$ and $C_2$, then they must agree on the causal
relationship between $C_1$ and $C_2$.  But that is not enough.
Suppose that in both models $C_1$ depends on $C_2$, $C_3$, and $C_4$.
Then in a precise sense, the two models must agree on \emph{how} the
dependence works, despite describing the world using possibly different
%joe10
%sets of variables.  Roughly speaking, we are able to do this when,
sets of variables.  Roughly speaking, this is the case when,
for every variable $C$ that the two models have in common, we can
designate one of the models as being ``dominant'' with respect to $C$, and
use that model to determine the relationships for $C$.
When $M_1$ and $M_2$ are compatible, we are able to
%dalal23
%construct a combined model $M_1 \oplus M_2$ that can be viewed as
construct a merged model $M_1 \oplus M_2$ that can be viewed as
%joe14
%satisfying all of Bradley, Dietrich, and List's desiderata.
satisfying all but one of BDL's desiderata (and we argue that the one
it does not satisfy is unreasonable).  
%dalal22
% I think at this point we need to say what information the joined
% compatible models have  
% and what they enable in terms of analysis.
% we talk about combined models in the sense of joining compatible ones (for which we dont consider probabilities)
% and also talk about combined in terms of incompatible ones with probability... maybe better to distinguish the 
% the term combined in these two settings?

%joe32*: added
In a precise sense, all conclusions that hold
%dalal23
%in either of the models $M_1$ and $M_2$ also holds in the combined model (see
in either of the models $M_1$ and $M_2$ also hold in the merged model (see
Theorem~\ref{oplusproperties}(e) and Theorem~\ref{oplusproperties1}(e)).
%dalal23
%By combining in this way, the combined model takes advantage of the
In this way, the merged model takes advantage of the
information supplied by all the experts (at least, to the extent that
%joe35
%the experts are compatible), and can go beyond what we can do with
the experts' models are compatible), and can go beyond what we can do with
either of the individual models (e.g., considering interventions that
%dalal23
%simulatenously act on variables that are in $M_1$ but not in $M_2$ and
simultaneously act on variables that are in $M_1$ but not in $M_2$ and
variables that are in $M_2$ but not in $M_1$).

%joe10: added paragraph break
%dalal22: maybe too strong statement
%This set of constraints is very restrictive, and, as we show on
%joe32: we haven't said anything about constraints
%This set of constraints may be too restrictive, and, as we show on
The set of constraints that need to be satisfied for models to be
compatible is quite restrictive; as we show on
real-life examples, 
%joe12: let's tone it down
%models are usually not compatible in this sense.
models are often not compatible in this strong sense.
%joe10
We thus define two successively more general notions of compatibility,
%dalal22: are the two generalisations referring to I4' and partial
%compatibility?
%joe32: yes
%dalal23
%and demonstrate, by means of examples, how this allows for the combining
and demonstrate, by means of examples, how this allows for the merging
a wider class of models, and reasoning about interventions under less
strict conditions. 
%dalal22
%But even with this more general notions, we may find that not all the
%joe32
%But even with this more general notions, some
But even with these more general notions, some 
%dalal20: should say compatible rather than incompatible.e 
%experts' models are incompatible.  In that case, we simply place a
%dalal22
%experts' models are compatible. 
%joe35
%experts' models maybe still incompatible due to disagreements about some 
experts' models may still be incompatible due to disagreements about some 
parts of the model, even though possible interventions to be considered do not affect
those parts of the model.  
%dalal22: brought forward decomposition and pushed the discussion on combining 
% to follow the arguments  in the paper
%joe32
%We, therefore, introduce a notion of causal model decomposition to
We therefore introduce a notion of causal model decomposition to 
allow policymakers to
``localize'' the incompatibility between models,and
%joe32
%combine parts of the models so as to indicate an effective intervention.
%dalal23
%combining the parts of the models that can be combined.
merge the parts of the models that are compatible.

%dalal22 added this introductory part
%joe32: we don't use standard techniques to combine, but to assign
%probability 
%Having set out the formal foundation for (in)compatibility, we then
%show how   models may be combined  using relatively standard
%techniques, based on the perceived reliability of the experts who
%proposed them.
%dalal23
%Having set out the formal foundation for combining causal models, we
Having set out the formal foundation for merging causal models, we 
show how   probabilities can be assigned to different reasonable ways
of combining experts' causal models based  on the perceived reliability
of the experts who proposed them, using relatively standard techniques.
The policymaker will then have a probability on causal
models that she can use to decide which interventions to
implement.  Specifically, we can use the probability on causal models to
compute the probability that an intervention is efficacious.
%joe32
%Combining that with the cost of implementing the intervention, she can
Combining that with the cost of implementing the intervention, the
policymaker can 
compute the most effective intervention.  As we shall see, although we
%joe2
%work with the same causal strutures used to define causality,
work with the same causal structures used to define causality,
interventions are different from (and actually simpler to analyze
than) causes.

%dalal20
%dalal22 moved this earlier
%We also introduce a notion of causal model decomposition to 
%allow policymakers to
%``localize'' the incompatibility between models, and
%combine parts of the models so as to indicate an effective intervention.

%dalal22
We draw on various examples from the literature (including real-world
scenarios involving complex sociological phenomena) to illustrate
%joe32
%this, such as
%crime prevention scenarios \citet{Sampson:2013} and radicalization 
our approach, including
crime-prevention scenarios \cite{Sampson:2013}, radicalization in prisons
%joe32
%\citet{Wikstrom:2017}. We thus  believe that our approach provides a 
\cite{Wikstrom:2017}, and child abuse~\cite{Marinetto:2011}.

 These examples reinforce our  belief that our
approach provides a  
useful formal framework that
can be applied to the determination of appropriate interventions for
%joe32
%policy making. 
policymaking. 

%We believe that our approach provides a useful formal framework that
%can be applied to the determination of appropriate interventions in
%real-world scenarios involving complex sociological phenomena, such as
%crime prevention scenarios \citet{Sampson:2013} and radicalization 
%\citet{Wikstrom:2017}.   
%
%hana9
%joe11: As I said, I think we can (and probably should) include the
%full version with our submission).
\shortv{
  Proofs and detailed descriptions of
  %joe11
  some of
  the examples in the paper are
deferred to the  
%joe11: As I said, I think we can (and probably should) include the
%full version with our submission).  This also saves a line
%full paper, due to lack of space;an anonymized version of the full
%paper (with proofs) can be found at
%https://www.dropbox.com/s/53467mtxoqor97j/inter-ventions.pdf?dl=0.
%dalal13:
%full paper (included with the submission), due to lack of space.
%joe14; should we post the full paper and give a url, or should we
%just assume that they'll find it.
full paper, due to lack of space.
}

%joe10: left for full paper
\fullv{
The rest of the paper is organized as follows. Section~\ref{sec:background} 
provides some  background material on causal
models.  We formally define our notion of intervention and compare it
to causality in Section~\ref{sec:intervention}.
We discuss our concept of compatibility and how causal
%dalal23
%models can be combined in Section~\ref{sec:combining}. We discuss how the
models can be merged in Section~\ref{sec:combining}. We discuss how the
notions of interventions and of compatible models can be used by the
policymakers 
to choose optimal interventions in Section~\ref{sec:experts}. Finally,
%hana26
we summarize our results and outline future directions in
Section~\ref{sec:conclusions}. 
%%we illustrate these 
%%concepts on two case studies in Section~\ref{sec:casestudies}.
%in Section~\ref{sec:casestudies}, we illustrate these 
%concepts using two legal cases. 
}

%joe1: Do any of the papers below consider interventions in the same
%sense that we do.  If they are relevant, we should cite them, but we
%need to explain their relevance.  For example, transportability isn't
%so obviously relevant.

\section{Causal Models}\label{sec:background} 

%joe10*: I looked over this section.  Unfortunately, I don't see what
%to cut.  This is already the ``short'' version.  The trouble is that
%we need the definition of causal models, and we've already cut the
%definition of causality.
%joe1: no need to define causality; we're not working with it.
In this section, we review the definition of causal
%joe32
%models introduced by Halpern and Pearl \citeyear{HP01b}.
models introduced by \citet{HP01b}.
%joe32
%The material in this section is largely taken from \citet{Hal48}.
The material in this section is largely taken from \cite{Hal48}. 
%\subsubsection{Causal structures}

%The HP approach assumes that the world is described in terms of
We assume that the world is described in terms of 
variables and their values.  
Some variables may have a causal influence on others. This
influence is modeled by a set of {\em structural equations}.
It is conceptually useful to split the variables into two
sets: the {\em exogenous\/} variables, whose values are
determined by 
factors outside the model, and the
{\em endogenous\/} variables, whose values are ultimately determined by
the exogenous variables.  
%hana9
\fullv{
For example, in a voting scenario, we could have endogenous variables
that describe what the voters actually do (i.e., which candidate they
vote for), exogenous variables 
that describe the factors
that determine how the voters vote, and a
variable describing the outcome (who wins). 
}
The structural equations
%joe15: there is no outcome
%describe how the outcome is
describe how these values are 
%hana9
\shortv{determined.}
\fullv{determined (majority rules; a candidate
wins if $A$ and at least two of $B$, $C$, $D$, and $E$ vote for him;
etc.).
}

Formally, a \emph{causal model} $M$
is a pair $(\Scal,\F)$, where $\Scal$ is a \emph{signature}, which explicitly
lists the endogenous and exogenous variables  and characterizes
their possible values, and $\F$ defines a set of \emph{(modifiable)
structural equations}, relating the values of the variables.  
A signature $\Scal$ is a tuple $(\U,\V,\R)$, where $\U$ is a set of
exogenous variables, $\V$ is a set 
of endogenous variables, and $\R$ associates with every variable $Y \in 
\U \union \V$ a nonempty set $\R(Y)$ of possible values for 
$Y$ (that is, the set of values over which $Y$ {\em ranges}).  
%joe2: it's a joint paper :-)
%For simplicity, I assume here that $\V$ is finite, as is $\R(Y)$ for
For simplicity, we assume here that $\V$ is finite, as is $\R(Y)$ for
every endogenous variable $Y \in \V$.
$\F$ associates with each endogenous variable $X \in \V$ a
function denoted $F_X$
%joe7: adde
(i.e., $F_X = \F(X)$)
such that $F_X: (\times_{U \in \U} \R(U))
\times (\times_{Y \in \V - \{X\}} \R(Y)) \rightarrow \R(X)$.
This mathematical notation just makes precise the fact that 
$F_X$ determines the value of $X$,
given the values of all the other variables in $\U \union \V$.
%hana9
\fullv{
If there is one exogenous variable $U$ and three endogenous
variables, $X$, $Y$, and $Z$, then $F_X$ defines the values of $X$ in
terms of the values of $Y$, $Z$, and $U$.  For example, we might have 
$F_X(u,y,z) = u+y$, which is usually written as
$X = U+Y$.   Thus, if $Y = 3$ and $U = 2$, then
$X=5$, regardless of how $Z$ is set.%
\footnote{The fact that $X$ is assigned  $U+Y$ (i.e., the value
of $X$ is the sum of the values of $U$ and $Y$) does not imply
that $Y$ is assigned $X-U$; that is, $F_Y(U,X,Z) = X-U$ does not
necessarily hold.}  
}  

%joe3: reinstated this paragraph
The structural equations define what happens in the presence of external
interventions. 
Setting the value of some variable $X$ to $x$ in a causal
model $M = (\Scal,\F)$ results in a new causal model, denoted $M_{X
\gets x}$, which is identical to $M$, except that the
equation for $X$ in $\F$ is replaced by $X = x$.

%joe2: reordered, and made the dependencies context free
The dependencies between variables in a causal model $M$
%joe12: redundant M
%$M$
can be described using a {\em causal network}\index{causal
  network} (or \emph{causal graph}),
%joe2: cut
%For each context $\vec{u}$,
%there is a graph on the variables in $\V$,
%with an edge going from $X$ to $Y$ if
%$Y$ depends on $X$ in context $\vec{u}$.  Note that the roots of the
%causal network are labeled by exogenous variables.  
whose nodes are labeled by the endogenous and exogenous variables in
$M = ((\U,\V,\R),\F)$, with one node for each variable in $\U \cup
\V$.  The roots of the graph are (labeled by)
the exogenous variables.  There is a directed edge from  variable $X$
to $Y$ if $Y$ \emph{depends on} $X$; this is the case
if there is some setting of all the variables in 
$\U \union \V$ other than $X$ and $Y$ such that varying the value of
$X$ in that setting results in a variation in the value of $Y$; that
is, there is 
a setting $\vec{z}$ of the variables other than $X$ and $Y$ and values
$x$ and $x'$ of $X$ such that
%joe12: put inside \shortv, to remove space before footnote
%$F_Y(x,\vec{z}) \ne F_Y(x',\vec{z})$.
\fullv{$F_Y(x,\vec{z}) \ne F_Y(x',\vec{z})$.}  
%hana9 added a footnote to replace a paragraph below for the conference version
\shortv{
%joe12: put inside
$F_Y(x,\vec{z}) \ne F_Y(x',\vec{z})$.%  
%dalal16:
%\footnote{In many papers in the literature (e.g., \citet{BDL14,Sampson:2013})
%joe32
  %  \footnote{In many papers (e.g., \citet{BDL14,Sampson:2013}),
    \footnote{In many papers (e.g., \cite{BDL14,Sampson:2013}),
%a causal model is defined simply by a causal graph indicating the
a causal model is defined  by a causal graph indicating the
%dalal11: remove the extra whether
%dependencies, perhaps with an indication of whether whether a change
dependencies, perhaps with an indication of whether a change
has a positive or negative effect. Our models are more expressive, since the
equations typically provide much more detailed information regarding
%dalal11:
%the dependence between variables.} 
the dependency between variables.}
}
A causal model  $M$ is \emph{recursive} (or \emph{acyclic})
if its causal graph is acyclic.%
%joe39
\footnote{Halpern \citeyear{Hal48} calls this \emph{strongly
    recursive}, and takes a recursive model to be one where, for each
  context $\vec{u}$, the dependency graph is acyclic (but it may be a
  different acyclic graph for context, so that in one context $A$ may
  be an ancestor of $B$, while in another, $B$ may be an ancestor of
  $A$).}
It should be clear that if $M$ is an acyclic  causal model,
then given a \emph{context}, that is, a setting $\vec{u}$ for the
exogenous variables in $\U$, the values of all the other variables are
determined (i.e., there is a unique solution to all the equations).
%hana9
\fullv{   
We can determine these values by starting at the top of the graph and
%joe3
%working our down.  In this paper, following the literature, we
working our way down.
}
%joe39
%In this paper, following the literature, we restrict to recursive models.
In this paper, following the literature, we restrict to acyclic models.
%dalal22: added example as requested by a reviewer

\fullv{   
%joe32: I'm not convinced that this is the best place for the
%example, although I don't feel strongly about it.  I added the next
%sentence 
The following example, due to \citet{Lewis00}, describes a simple
causal scenario. 
  \begin{example}\label{rock1}
  %dalal23
  \normalfont
Suzy and Billy both pick up rocks and throw them at a bottle. Suzy's rock gets there first, shattering
the bottle. Since both throws are perfectly accurate, Billy's would have shattered the bottle had it not
been preempted by Suzy's throw.
Consider a model  having
an exogenous variable $U$ that encapsulates whatever background factors
cause Suzy and Billy to decide to throw the rock
(the details of $U$
do not matter, since we are interested only in the context where $U$'s
%joe32: rewrote
%value is such that both Suzy and Billy throw), a three binary
value is such that both Suzy and Billy throw).  Thus, $U$ has four
possible values, depending on which of Suzy and Billy throw.   We also
have three binary variables: $\ST$ for Suzy 
throws,  $\BT$ for Billy throws, and  $\BS$ for bottle shatters.
%joe32:
$\ST = 1$ means ``Suzy throws'';  $\ST=0$ means that she does not.  We
interpret $\BT=1$, $\BT=0$, $\BS=1$, and $\BS=0$ similarly.
%joe32
%$\ST=1$ if Suzy throws; $\ST = 0$ if she does not.  $\BT$ and $\BS$
The values of $\ST$ and $\BT$ are determined by the
%joe32: not quite right
%context with $F_{\ST}(\vec{u})$ is such that $\ST = 1$ if Suzy
%throws, and $\ST = 
%0$ if she doesn't, and similarly $F_{\BTs}(\vec{u})$ is such that $\BT
%0$ if she doesn't. and similarly $F_{\BTs}(\vec{u})$ is such that $\BT
%%= 1$ if Billy throws, and $\BT = 0$ if he doesn't.
context.
The value of $\BS$
is determined by the equation $F_{\BSs}(\vec{u},\ST, \BT) = \ST \vee
\BT$. 
%joe32
%The model is depicted in Figure~\ref{fig0}.
The causal graph corresponding to this model is depicted in
Figure~\ref{fig0}.%
%joe32: moved up from below and put in footnote.  This is really a
%side-comment. 
%
%Note that a more elaborate model may include variables that capture
%temporal ordering over whose rock hits the  bottle first.
%Since the problem we address in this paper
%is orthogonal, we refer the reader to \cite{XXXX} for further details
%on representing temporal ordering  of events in causal models. 
\footnote{We could also model this problem using time-indexed variables that
explicitly model the temporal order of events; see \cite{Hal48,HP01b}.
%dalal23
%The issues raised bythis example could be made using either approach.}  
The issues raised by this example could be made using either approach.}  
%dalal23
\eprf
\begin{figure}[h]
\begin{center}
%joe32: let's make it a bit larger, to be consistent with the other diagrams.
  %\setlength{\unitlength}{.07in}
  \setlength{\unitlength}{.15in}
\begin{picture}(8,8)
%dalal23 thickened lines and increase circle size to 0.4
\thicklines
\put(3,0){\circle*{.4}}
\put(3,8){\circle*{.4}}
\put(0,4){\circle*{.4}}
\put(6,4){\circle*{.4}}
\put(3,8){\vector(3,-4){3}}
\put(3,8){\vector(-3,-4){3}}
\put(0,4){\vector(3,-4){3}}
\put(6,4){\vector(-3,-4){3}}
\put(3.7,-.2){$\BS$}
\put(-2,3.8){$\ST$}
\put(6.15,3.8){$\BT$}
\put(3.4,7.8){$U$}
\end{picture}
\end{center}
%joe32
%\caption{A  model for the rock-throwing example.}\label{fig0}
\caption{The causal graph for the rock-throwing example.}\label{fig0}
\end{figure}
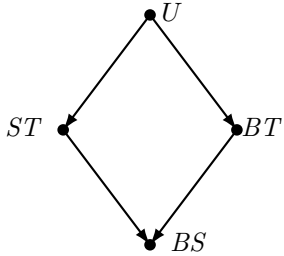
\end{example}
%hana26 need to add a reference for the above (temporal ordering of
%events in causal models) 
}

%hana9 keep for the full version?
\fullv{
%joe3: added
%joe32
  %  In many papers in the literature (e.g., \citet{BDL14,Sampson:2013})
    In many papers in the literature (e.g., \cite{BDL14,Sampson:2013})
a causal model is defined simply by a causal graph indicating the
%dalal22 repeated whether
%dependencies, perhaps with an indication of whether whether a change
dependencies, perhaps with an indication of whether  a change
has a positive or negative effect; that is, edges are annotated with
$+$ or $-$, so that an edge from $A$ to $B$ annotated with $+$ means
that an increase in $A$ results in an increase in $B$, while
if it is annotated with a $-$, then an increase in $A$ results in a
decrease in $B$ (where what constitutes an increase or decrease is
%dalal22
%determined by the model).  Our models are more expressive, since the
determined by the model).  Examples of these are shown in Section \ref{sec:combining}.
Our models are more expressive, since the
equations typically provide much more detailed information regarding
%dalal22
%the dependence between variables; the causal graphs capture only part
the dependence between variables (as shown in Example \ref{rock1}); the causal graphs capture only part
of this information.  Of course, this extra information makes
%dalal23
%combining models more difficult (although, as the results of
merging models more difficult (although, as the results of
%joe14
%Bradley, Dietrich, and List \shortcite{BDL14} show, the difficulties in
BDL show, the difficulties in
%dalal23 we need to be consistent in the wat we decribe BDL14's work
% either "merge models" or "combine experts' opinions"
%joe33: let's use our terminology
%combining models already arise with purely qualitative graphs).
merging models already arise with purely qualitative graphs).
}

To define interventions carefully, it is useful to have a language in
which we can make statements about interventions.
Given a signature $\Scal= (\U,\V,\R)$, a \emph{primitive event} is a
%joe20*: should we allow causal formulas to include exogenous
%variables (so that X=x makes sense for X \in \U, as does
%{Y <- y]\phi for Y \in \U.  It is straightforward to give semantics
%to such formulas, and it will make our life easier later.  Meir and I
%do this in our paper.  I'll point out in a later comment where we
%could use this.
%joe21*: we should still discuss this
formula of the form $X = x$, for  $X \in \V$ and $x \in \R(X)$.  
A {\em causal formula (over $\Scal$)\/} is one of the form
$[Y_1 \gets y_1, \ldots, Y_k \gets y_k] \varphi$,
where
$\varphi$ is a Boolean
combination of primitive events,
%joe25
%$Y_1, \ldots, Y_k$ are distinct variables in $\V$, and
$Y_1, \ldots, Y_k$ are distinct variables in $\U \cup \V$, and
$y_i \in \R(Y_i)$.%
%joe25
%joe32
%\footnote{In earlier work \citet{Hal48,HP01b}, each $Y_i$ was taken to
\footnote{In earlier work \cite{Hal48,HP01b}, each $Y_i$ was taken to
  be an endogenous variable.  For technical reasons (explained in
  Section~\ref{sec:combining}), we also allow $Y$ to be exogenous.} 
Such a formula is abbreviated
as $[\vec{Y} \gets \vec{y}]\varphi$.
The special
case where $k=0$
is abbreviated as
$\varphi$.
Intuitively,
$[Y_1 \gets y_1, \ldots, Y_k \gets y_k] \varphi$ says that
$\varphi$ would hold if
$Y_i$ were set to $y_i$, for $i = 1,\ldots,k$.
%dalal22 Another example here?
%Reviewer 1 says "Here a potential confusion is, when some of Y's are already 
%contradictory to each other, how should the satisfiability of \varphi be determined?

We call a pair $(M,\vec{u})$ consisting of a causal model $M$ and a
context $\vec{u}$ a \emph{(causal) setting}.
A causal formula $\psi$ is true or false in a setting.
We write $(M,\vec{u}) \satt \psi$  if
the causal formula $\psi$ is true in
the setting $(M,\vec{u})$.
The $\satt$ relation is defined inductively.
$(M,\vec{u}) \satt X = x$ if
the variable $X$ has value $x$
in the unique (since we are dealing with acyclic models) solution
to the equations in
$M$ in context $\vec{u}$ (that is, the
unique vector of values for the exogenous variables that simultaneously satisfies all
equations 
in $M$ 
with the variables in $\U$ set to $\vec{u}$).
%joe25
%Finally, 
%$(M,\vec{u}) \satt [\vec{Y} \gets \vec{y}]\varphi$ if 
%$(M_{\vec{Y} = \vec{y}},\vec{u}) \satt \varphi$.
If $k \ge 1$ and $Y_k$ is an endogenous
variable, then
$$\begin{array}{ll}
(M,\vec{u}) \models [Y_1 \gets y_1, \ldots, Y_k \gets
y_k]\varphi \mbox{ iff } \\
(M_{Y_k \gets y_k},\vec{u}) \models
[Y_1 \gets y_1, \ldots, Y_{k-1} \gets y_{k-1}]\varphi.\end{array}$$
If $Y_k$ is an exogenous variable, then
$$\begin{array}{ll}
(M,\vec{u}) \models [Y_1 \gets y_1, \ldots, Y_k \gets
y_k]\varphi \mbox{ iff }\\ (M,\vec{u}[Y_k/y_k]) \models
[Y_1 \gets y_1, \ldots, Y_{k-1} \gets y_{k-1}]\varphi,\end{array}$$
where $\vec{u}[Y_k/y_k]$ is the result of replacing the value of $Y_k$
in $\vec{u}$ by $y_k$.

%dalal22 Another example here? 
% Reviewer 1 says "An example about inference is necessary."

\section{Interventions}\label{sec:intervention}

%joe1
%In this section we present our definitions for causal interventions,
%compare them with the definition of actual cause and argue that for
%policymakers, the concept of interventions is more relevant than 
%the concept of actual causes.
In this section, we define  (causal) interventions, and compare 
the notion of intervention to that of cause.

%joe1
%\dfn\label{def-intervention}[Intervention]
%joe10
%\dfn\label{def-intervention}[Intervention]
\dfn\label{def-intervention}\fullv{[Intervention]}
%joe1: should we perhaps call it an effective intervention?  I don't
%feel strongly about this.
%dalal22 changed for to on to be consistent with  definition 1
%$\vec{X} = \vec{x}$ is an \emph{intervention for $\varphi$ in
%joe32*: ``on'' is inappropriate here.  ``Intervening on a variable'' has a
%standard interpretation in the literature.  It means changing the
%value of that variable.  In this definition, we are intervention on
%X, not on \phi.  We could say
%``intervention leading to \phi'' if we wanted to be more explicit.
%changing its value
%$\vec{X} = \vec{x}$ is an \emph{intervention on $\varphi$ in
$\vec{X} = \vec{x}$ is an \emph{intervention leading to
%hana37 changed \phi to \neg{\phi} and added "for \phi"
%joe41: this reads strangely; I don't think we want to say that
%\vec{X} = \vec{x} is an intervention on \vec{X} for \varphi; I just
%cut it.
  %  $\neg{\varphi}$ (or an `intervention on $\vec{X}$ for $\varphi$')
    $\neg{\varphi}$ 
in $(M, \vec{u})$} if the following
three conditions hold:
\begin{description}
   \item[{\rm I1.}]\label{ac1} $(M,\vec{u}) \satt \varphi$.
\item[{\rm I2.}]\label{ac2} 
$(M,\vec{u}) \satt [\vec{X} \gets \vec{x}]\neg \varphi.$
\item[{\rm I3.}] \label{ac3}
$\vec{X}$ is minimal; there is no strict subset $\vec{X}'$ of $\vec{X}$
%joe3
%  and values $\vec{x}'$ such that $\vec{X}' = \vec{x}'$ satisfies I1 and I2.
  and values $\vec{x}'$ such that $\vec{X}' = \vec{x}'$ satisfies I2.
\end{description}
\end{definition}

%joe1: added
%joe2
%I1 just says that if we intervene on $\varphi$, then $\varphi$ must be true.
%I2 says that performing the intervention must result on $\varphi$ must result in  
%dalal2: removed "just"
I1  says $\varphi$ must be true in the current setting $(M,\vec{u})$, while
I2 says that performing the intervention results in
$\varphi$ no longer being true.  I3 is  a minimality condition.  From
%joe2
%a policymaker's perspective, I2 is the key condition.  It says by
a policymaker's perspective, I2 is the key condition.  It says that by
making the appropriate changes, we can bring about a change in $\varphi$.

%joe3: shortened, and tried to make more accessible
%Our notion of intervention is related to, but slightly different from 
%Pearl introduces interventions as additional variables that
%\emph{disrupt} the causal structure of the
%model~\citet{Pearl:2009}. That definition assumes that 
%the model was analysed before considering interventions, and
%interventions modif 
%joe8
%Our notion
%joe32
%Our definition of intervention
Our definition of an intervention leading to $\neg{\phi}$
slightly generalizes others in the literature.
%joe32
%Pearl \citeyear{Pearl:2009} assumes that the causal model is first
\citet{Pearl:2009} assumes that the causal model is first
analyzed, and then a new intervention variable $I_V$ is added for each
variable $V$ on which we want to intervene.
%joe12: I'd prefer not to say ``disrupt''
%These variables disrupt the causal structure of the model; that is, if
If $I_V = 1$, then the appropriate intervention
on $V$ takes place, independent of the values of the other parents of
%joe8
%$V$;  If $I_V = 0$, then $I_V$ has no effect, and the behavior of $V$
$V$;  if $I_V = 0$, then $I_V$ has no effect, and the behavior of $V$
is determined by its parents, just as it was in the original model.
%the model was analysed before considering interventions, and
%interventions modify the existing 
%joe3
%causal relationship. Woodward takes a similar approach, defining interventions
%causal relationship. Woodward \shortcite{Woodward03} takes a similar
%approach, defining interventions 
%essentially as \emph{switches}: if the intervention is off, then the
%variable on which it intervenes 
%depends on other variables in the model; if the intervention is on,
%then that variable depends solely 
%joe2
%on the intervention~\citet{Woodward:2003}. The definitions
%on the intervention. The definitions
%in~\citet{Lu02,Korb04} are similar:
%joe32
%Lu and Druzdzel \citeyear{lu02}, Korb et al.~\citeyear{Korb04},
%and Woodward \citeyear{Woodward03} take similar approaches.
\citet{lu02}, \citet{Korb04},
and \citet{Woodward03} take similar approaches.
%joe3: overkill
%all consider only single-variable interventions on a single variable
%in the model, and assume 
%%and state that interventions  have a \emph{cut-off} effect on the
%%variables, that is,
%that the interventions make the intervened 
%variable causally independent of other variables in the model and
%dependent only on the intervention.  

%joe3*: rewrote
%This definition can be regarded as a special case of the definition we
%present here. Indeed, it is 
%fairly straightforward to define a ``switch'' variable in the model
%that, if on, cuts off the causal 
%dependencies of a variable it intervenes on. We can view our
%definition as a generalization, where 
%we do not restrict interventions to single
%variables or only to those that replace the existing causal dependencies; 
%rather, we state that the defining characteristic of an intervention
%is the fact that they have a 
%counterfactual effect on the outcome.
%joe12*: Pearl also assumed that the variables to be intervened on
%were in the model.  He just added extra ``intervention variables''.
%I also don't understand what ``overriding the effects of other
%variagbles mean''
%We assume that the variables to be intervened on are already in the
%model, and do not necessarily override the effects of all other
%variables.
We do not require special intervention variables; we just allow
interventions directly on the variables in the model.
%dalal13:
But we can \fullv{certainly} assume as a special case that for
each variable $V$ in the model there is a special intervention
variable $I_V$ that works just like Pearl's intervention variables,
and thus recover the other approaches considered in the literature.
%joe32: added
Our definition also focuses on the outcome of the intervention, not
just the intervention itself.
%dalal13:
\fullv{In any case, it} 
\shortv{It} should be
clear that all these definitions are trying to capture \fullv{exactly} the
same intuitions, and differ only in minor ways.

%joe1: rewrote
%The following examples illustrate
%the concept of interventions and the difference between interventions
%and actual causes.
Although there seems to be relatively little disagreement about how to
capture intervention, the same cannot be said for causality.
%joe2: moved this sentence back
%joe3
%Causality is a much more subtle notion. 
Even among definitions that involve counterfactuals and structural
equations 
%joe14
%\citet{GW07,Hal47,HPearl01a,HP01b,hitchcock:99,Hitchcock07,Woodward03},
%joe32
%\citet{GW07,Hal47,HP01b,hitchcock:99,Hitchcock07,Woodward03},
\cite{GW07,Hal47,HP01b,hitchcock:99,Hitchcock07,Woodward03},
%joe2
%equations, are a number of subtle different definitions.   For
%joe3
%equations, there are a number of subtle different definitions.   For
there are a number of subtle variations.
%joe3
%For example, the initial Halpern-Pearl 
%definition \citet{HPearl01a} was modified \citet{HP01b}, and then
%%joe2
%%modified again by Halpern \shortcite{Hal47}.  THere are a number of
%modified again \citet{Hal47}.  There are a number of
%other definitions of causality that use causal structures (e.g.,
%\citet{GW07,hitchcock:99,Hitchcock07,Woodward03}).
%joe2
%joe11: line shaving
%Fortunately for us, the definition of intervention does not depend on
%joe12: oops!
%\fullv{Fortunately for us,} the definition of intervention does not depend on
\fullv{Fortunately for us,} 
\shortv{Fortunately,} the definition of intervention does not depend on
how causality is defined.
%joe3
%Nevertheless, it is instructive to compare the definitions.
While we do not get into the details of causality here, it is
instructive to compare the definitions  of causality and intervention.

%joe2
%The definition given by Halpern \shortcite{Hal47} has conditions
%joe3
%The definition of causality given by Halpern \shortcite{Hal47} has conditions
% dalal: corrected spelling of analogues -> analogous
For definiteness, we focus on the definition of causality given by
%joe32
%Halpern \citeyear{Hal47}.  It has
\citet{Hal47}.  It has
%joe12
conditions
%dalal22
AC1--3 that are analogous of I1--3. 
Specifically, AC1 says $\vec{X} = \vec{x}$ is a cause of $\varphi$
%joe2
%if $(M,\vec{u}$ if $(M,\vec{u}) \satt (\vec{X} = \vec{x}) \land \varphi$ and
in $(M,\vec{u})$ if $(M,\vec{u}) \satt (\vec{X} = \vec{x}) \land \varphi$ and
%joe12
%AC3 is a minimality condition.  AC2 is more complicated condition; it
AC3 is a minimality condition.  AC2 is a more complicated condition; it
says that there exist  values $\vec{x}'$ for the variables in
$\vec{X}$, a (possibly empty) subset $\vec{W}$ of variables, and
values $\vec{w}$ for the variables in $\vec{W}$ such
that $(M,\vec{u}) \satt \vec{W} = \vec{w}$ and 
$(M,\vec{u}) \satt [\vec{X} \gets \vec{x}, \vec{W} \gets \vec{w}]
  \neg \varphi$.  We do not attempt to explain or motivate AC2 here,
%joe3
%  since our focus is not causality.  However, the following example,
  since our focus is not causality.

%joe32: not appropriate, given the move  
%The following example,
%  %  due to Lewis \citeyear{Lewis00},
%    due to \citet{Lewis00},
%%joe3
%  %  illustrates the need for some nontrivial requirement, as well as
%%    highlighting the differences 
%  illustrates some of the subtleties, and
%  highlights the differences between causality and intervention. 

%dalal22
%Suppose that Suzy and Billy both pick up rocks and throw them at  a bottle.
Consider Example \ref{rock1} again. 
%dalal22 removed the below since we formalise it now in Example 1
%Suzy's rock gets there first, shattering the
%bottle.  Since both throws are perfectly accurate, Billy's would have
%shattered the bottle had Suzy not thrown.  Most people would say that
%Suzy is a cause, and not Billy.  Part of the difficulty in getting a
%good definition of causality is to ensure that the definition gives us
%this result (given an appropriate causal model).  However,
%dalal22
%Suzy's throw by itself is not an intervention for the bottle shattering.
%joe32
%Suzy's throw by itself is not an intervention on the bottle shattering.
%hana37
Changing the value of Suzy's throw by itself is not an intervention
leading to the bottle not being shattered. 
%Suzy's throw by itself is not an intervention leading to the bottle shattering.
Even if we prevent Suzy from throwing, the bottle will still shatter
because of Billy's throw.
That is,
%joe32: no need to repeat this; we said it in the description of the example
%if we have variables $\ST$ and
%$\BT$ for Suzy's throw and Billy's throw, with possible values 0 and 1
%$(\ST=1$ if Suzy throws, $\ST=0$ if she doesn't, and similarly for
%Billy), then
although $\ST=1$ is a cause of the bottle shattering,
%joe2
%$\ST=0$ is not an intervention for the bottle shattering; $\ST=0 \land
%BT=0$ is an intervention for the bottle shattering (but $\ST=1 \land
%joe3
%$\ST=0$ is not an intervention for the bottle shattering.  On the
%dalal22
%$\ST=0$ is not an intervention for the bottle shattering;
%joe32
%$\ST=0$ is not an intervention on the bottle shattering;
$\ST=0$ is not an intervention leading to the bottle 
%hana37
not being shattered;
%shattering;
intervening on $\ST$ alone does not change the outcome.
On the
other hand,  $\ST=0 \land \BT=0$ is an intervention leading to the bottle
%hana37
not being shattered,
%shattering, 
but $\ST=1 \land \BT=1$ is not a cause of the bottle shattering; it violates
%joe10: line shaving 
%the minimality condition AC3.
\fullv{the minimality condition} AC3.

It is almost immediate from the definitions that we have the following
relationship between interventions and causes:
\pro If $\vec{X} = \vec{x}$ is an intervention  
%hana37
%joe41
%for $\varphi$ in
leading to $\neg \varphi$ in
%leading to $\varphi$ in
$(M,\vec{u})$, then there is some subset  $\vec{X}'$ of $\vec{X}$
%joe3
%such $\vec{X}' = \vec{x}'$ is a cause of $\varphi$ in $(M,\vec{u})$,
such that $\vec{X}' = \vec{x}'$ is a cause of $\varphi$ in $(M,\vec{u})$,
%joe2
%where $\vec{x}'$ is such that $(M,\vec{u}) \satt \vec{X}' = \vec{x}'$.
where
%joe10
$\vec{x}'$ is such that
$(M,\vec{u}) \satt \vec{X}' = \vec{x}'$.
Conversely, if $\vec{X} = \vec{x}$ is a cause of $\varphi$ in
$(M,\vec{u})$ then there is a superset $\vec{X}'$ of $\vec{X}$ and
values $\vec{x}'$ such that $\vec{X}' = \vec{x}'$ is an intervention
%hana38
leading to $\neg\varphi$.
%for $\varphi$.
\end{proposition}

%joe32
%Halpern \citeyear{Hal47} proved that (for his latest definition) the
\citet{Hal47} proved that (for his latest definition) the
complexity of determining 
whether $\vec{X} = \vec{x}$ is a cause 
%joe3
%of $\varphi$ in $(M,\vec{u})$ is $DP$-complete
% dalal2: fixed formatting of DP
of $\varphi$ in $(M,\vec{u})$ is \textit{DP}-complete,
where \textit{DP} consists of those languages $L$ for which there exist a
language $L_1$ in NP and a language $L_2$ in co-NP such that $L = L_1
%joe32
%\cap L_2$ \citet{PY}.
\cap L_2$ \cite{PY}.
It is well known that \textit{DP} is at least as hard as NP and co-NP (and
most likely strictly harder).
%joe2
%
%The following theorem proves that the problem of determining whether
%joe10
\fullv{The following theorem shows that the}
\shortv{The} problem of determining whether
$\vec{X} = \vec{x}$ is an intervention is in a lower complexity class.
\begin{theorem}
Given a causal model $M$, a context $\vec{u}$, and a Boolean formula
$\varphi$, the problem of  
determining whether
%joe1
% subset of endogenous variables $\vec{X}$ and a setting $\vec{x}'$
%are an intervention
$\vec{X} = \vec{x}$ is an intervention
%hana37
leading to $\neg{\varphi}$
%for $\varphi$
in $(M,\vec{u})$ is co-NP-complete.
\end{theorem}

%joe10: put proof in full paper
\fullv{
\prf
First, we prove membership in co-NP. It is easy to see that checking
%joe1
%the conditions AC1 and AC2 of Def.~\ref{def-intervention}
%joe2
%conditions I1 and I2 of Def.~\ref{def-intervention}
conditions I1 and I2 of Definition~\ref{def-intervention} 
can be done in polynomial time by simply evaluating $\varphi$ first in $(M,\vec{u})$ and then in the modified context
where the values of $\vec{X}$ are set to $\vec{x}$. 
Checking whether I3 holds is in co-NP, because the complementary condition is
in NP; indeed,
we simply have to guess a subset $\vec{X}'$ of $\vec{X}$
and values $\vec{x}'$ and verify that I1 and I2 hold for $\vec{X}' =
\vec{x}'$ and $\varphi$, which, as we observed, can be done in polynomial time.

%joe1
%For the co-NP-hardness, we describe the reduction from UNSAT, which is
For co-NP-hardness, we provide a reduction from UNSAT, which is
%dalal4: to be consistent in spelling
% the language of all unsatisfiable Boolean formulae,
the language of all unsatisfiable Boolean formulas,
%joe1
to the intervention problem.
%joe1
%Given an input $\psi$ to UNSAT over the set of variables $\vec{V}$, we
%construct the model $M_\psi$ as follows.
Given a formula $\psi$ that mentions the set $\vec{V}$ of variables,
%joe2
%We construct a causal
we construct a causal
model $M_{\psi}$, context 
$\vec{u}$, and formula $\varphi$ such that $\vec{V}=1$ is an intervention
5joe42
%for $\varphi$ in $(M,\vec{u})$ iff $\psi$ is unsatisfiable.
leading to $\neg\varphi$ in $(M,\vec{u})$ iff $\psi$ is unsatisfiable.

%joe1
%The set of variables of $M$ is $\vec{V} \cup Y$ for a fresh variable
The set of endogenous variables in $M$ is $\vec{V} \cup 
\{V',Y\}$, where $V'$ and $Y$ are fresh variables not in $\vec{V}$.
%joe1
%The only structural equation that $M$ has is that
%$Y = \wedge{V \in \vec{V}} V$, that is, $Y$ is a conjunction of all
Let $\vec{W} = \vec{V} \cup \{V'\}$.
There is a single exogenous variable $U$ that determines the value
of the variables in $\vec{W}$: we have the equation $V = U$ for each
variable $V \in \vec{W}$.  
The equation for $Y$ is 
%joe2
%$Y = \lor_{V \in \vec{W}\,} V=0$ (so $Y=1$ if at least one variables in
%joe3
%$Y = \lor_{V \in \vec{W}\,} (V=0)$ (so $Y=1$ if at least one variables in
$Y = \lor_{V \in \vec{W}\,} (V=0)$ (so $Y=1$ if at least one variable in
$\vec{W}$ is 0).
%joe1
%We define  the context $\vec{u}$ that assigns all variables the value
%$0$,
%joe1
%joe1: no longer needed, now that I've defined \psi'
%and the formula   $\varphi$ as $\neg{\psi \vee  Y}$.
Let $\varphi$ be $\neg \psi \land (Y=1)$.  We claim that $\vec{W} = \vec{1}$ is an
%dalal22
%intervention for $\varphi$ in $(M_\psi,0)$ iff $\psi \in$ UNSAT.
%joe32
%intervention on $\varphi$ in $(M_\psi,0)$ iff $\psi \in$ UNSAT.
%hana37 \phi -> \neg{\phi}
intervention leading to $\neg{\varphi}$ in $(M_\psi,0)$ iff $\psi \in$ UNSAT.
%In other words, $\varphi$ is \bff  if either $\psi$ is \bft or all
%variables in $Y$ are assigned $1$. 
%joe1: I didn't see why this was wlog.  It is true for the formula
%\psi' that I defined
%W.l.o.g., we assume that $\psi$ has the value \bff when all variables
%in $Y$ are assigned $1$. 

%joe1
%Now, $\psi \in$ UNSAT iff $\vec{V}$ and the setting $\vec{1}$ that
%assigns all the variables the value $1$ is an intervention for 
%$\varphi$ in $(M,\vec{u})$.

%joe1
%Indeed, assume that $\psi \in$ UNSAT. Then, it is easy to see that
Suppose that $\psi \in$ UNSAT. Then, it is easy to see that
%joe2
%$(M,0) \satt \varphi$ (since $\psi$ is valid)
$(M,0) \satt \varphi$ (since $\neg \psi$ is valid)
and $(M,0) \satt [\vec{W} \gets \vec{1}]\neg \varphi$ (since
%joe2
%$(M,0)\satt [\vec{W} \gets \vec{1}](Y=0)$.
$(M,0)\satt [\vec{W} \gets \vec{1}](Y=0)$).
To see that I3 holds, suppose by way of contradiction that 
$\vec{W}' \gets \vec{w}'$ satisfies I1 and I2 for some strict subset
$\vec{W}'$ of  $\vec{W}$.  In particular, we must have
%joe2: corrected
%$(M,0)\satt [\vec{W} \gets   \vec{1}]\neg \varphi$.  We clearly have
%$(M,0)\satt [\vec{W} \gets \vec{1}](Y=1)$, so we must have $(M,0)
%\satt \varphi$ (since $(M,0) \satt \psi$, contradicting the assumption
$(M,0)\satt [\vec{W}' \gets   \vec{w}']\neg \varphi$.  We clearly have
$(M,0)\satt [\vec{W}' \gets \vec{w}'](Y=1)$, so we must have $(M,0)
\satt [\vec{W}' \gets   \vec{w}']\psi$, contradicting the assumption
that $\psi \in$ UNSAT.  Thus, $\vec{W} \gets \vec{1}$ is an
%hana37 \phi -> \neg{\phi}
intervention leading to $\neg{\varphi}$, as desired.

For the converse, suppose that $\vec{W} \gets \vec{1}$ is an
%joe2
%intervention for $\varphi$, as desired.  Then we must have
%dalal22
%intervention for $\varphi$.  Then we must have
%joe32
%intervention on $\varphi$.  Then we must have
%hana37 \phi -> \neg{\phi}
intervention leading to  $\neg{\varphi}$.  Then we must have
$(M,0) \satt [\vec{W}' \gets \vec{w}'] \neg \psi$ for all strict
subsets $\vec{W}'$ of $\vec{W}$ and all settings $\vec{w}'$ of the
variables in $\vec{W}'$.  Since, in particular, this is true for all
subsets $\vec{W}'$ of $\vec{W}$ that do not involve $V'$, it follows
that $\neg \psi$ is true for all truth assignments, so $\psi \in$
UNSAT.
%dalal23
%joe33: I think we should use \eprf consistently to mark the end of
%proofs and of examples
\eprf
%$\Box$
}
%hana9 since we said this in the intro, no longer needed here as well
%\shortv{The proof of this result and all others mentioned in this
%abstract can be found in the full paper.}

%hana added a discussion about real-life bound on the size of interventions
In practice, however, we rarely expect to face the co-NP complexity.
%joe3
%as reasonable interventions are expected to be quite small.
%In fact, a policymaker would usually bound the size of an intervention 
%by either the number of changes she is able to make or the total cost
%of the changes. While our model 
%does not allow to add costs to interventions, this is a
%straightforward extension and does not change 
%the complexity.
For reasons of cost or practicality, we would expect a policymaker to
consider interventions on at most $k$ variables, for some small $k$.
The straightforward algorithm that, for a given $k$,
checks all sets of 
%joe3
%variables of the model $M$ of size at most $k$ works in time $O(|M|^k)$.
variables of the model $M$ of size at most $k$ runs in time $O(|M|^k)$.
%dalal22 R1 asks for a justification here. I suppose for why we assume
%k to be  small
%joe32: ***
%dalal23 an open issue still
%joe33: are we looking for justification for the calculation?  I
%suspect so.  I don't think we need to say more about why we think that
%k should be small.  (Indeed, I don't see what we can say beyond ``for
%reasons of cost or practicality ...'' as we said above.

%joe8
%\section{Combining Compatible Causal Settings}\label{sec:combining}
%dalal23
%\section{Combining Compatible Causal Models}\label{sec:combining}
\section{Merging Compatible Causal Models}\label{sec:combining}
%joe2: rewrote to get rid of the dependence on contexts

This section  presents our definition for compatibility of expert opinions.
%joe3
%We consider each expert opinion to be represented as a causal model
We consider each expert's opinion to be represented by a causal model
and, for simplicity, that each expert expresses her opinion with certainty. 
%joe3
(We can easily extend our approach to allow the experts to have some
uncertainty about the correct model; see
%joe12
%the discussion at
the end of Section~\ref{sec:experts}.)
%hana8 added some text about gradual weakening of compatibility
% dalal7: change talk about weak and strong to strict and general
%joe10: how about ``strong compatibility'' and ``compatibility'' (and
%later, ``partial compatibility'').  It's shorter ...
%We start with a strict notion of compatibility, allowing us to construct
We start with a strong notion of compatibility,
%joe10
%allowing us to construct
%a unified causal model from two given causal models without contradicting any
%of the structural equations in the input models.
%joe10: shortened
%We then argue that in real world,
%this notion might be too restrictive. We define a generalization  of 
%this that support the construction of causal models from compatible sub-models 
%  and demonstrate their usefulness on real-life examples.
and then consider generalizations of this notion that are more widely
applicable.  

%dalal22
%\subsection{xDomination and Compatibility}
%joe32: it seems strange to have ``Full compatibility'' in the title
%when we never formally define that term
%\subsection{Full Compatibility}
%dalal23 definition 4 is named full compatibility 
%joe33: I guess that I can live adding ``full'' everywhere; it does
%force us to occasionally be more precise
\subsection{Full compatibility}

To specify what it means for a set of models to be compatible, we first  
define what it means for the causal model
$M_1$ to contain at least as much information about variable $C$ as
%dalal11:
%$M_2$, denoted $M_1 \succeq_C M_2$.  
 the causal model $M_2$, denoted $M_1 \succeq_C M_2$.  
Intuitively, $M_1$ contains at least as much information about $C$ as $M_2$ if
%dalal22 to address reviewer 3's comment 
%joe32: what change was made here?
%dalal23 I added the phrase " (i.e., they have the same structural
%equation for $C$)" 
%$M_1$ and $M_2$ say the same things about the causal structure of $C$
$M_1$ and $M_2$ say the same things about the causal structure of $C$ 
%dalal28: removed the sentence below as incorrect
% (i.e., they have the same structural equation for $C$)
as far as the variables that $M_1$ and $M_2$ share, but $M_1$ contains
%dalal22: do we want to introduce a notion of projection here to formally
%capture what is said below
%joe32: I lean towards not doing this; we already have lots of definitions ...
%dalal23 OK
%(possibly) more detailed information about $C$, because, for example, there are
(possibly) more information about $C$, because, for example, there are
additional variables in $M_1$ that affect $C$.
%dalal13:
%
% Dalal5: 
%joe11: added paragraph break and rewrote next lne
%For the remainder of this section, we consider
%that $B$ to be an \emph{immediate $M_2$-ancestor of $Y$ in $M_1$}
%dalal9:
%dalal25
%We capture this property  formally below. Say
We capture this property  formally below. We say
%dalal11: we never explain the term immediate M-ancestor of Y.
that $B$ is an \emph{immediate $M_2$-ancestor of $Y$ in $M_1$}
 if $B \in \U_2 \cup \V_2$, $B$ is an ancestor of $Y$ in $M_1$, and
there is a path from $B$ to $Y$ in $M_1$ that has no nodes in $\U_2
\cup \V_2$ other than $B$ and  $Y$ (if $Y \in \U_2 \cup \V_2$).
%joe12: added intuition
%hana15 - $B$ should be $Y$
%joe13
%That is, $Y$ is the first node in $M_2$ on a path from $B$ to $Y$ in
That is, $Y$ is the first node in $M_2$ after $B$ on a path from $B$ to $Y$ in
$M_1$.

% Dalal6: introduce definition label to distinguish form the later weaker definition
% For the remainder of this section, suppose that 
%joe3
%$M_1 = ((\U_1,\V_1,\R_1)\F_1)$ and $M_2 = ((\U_2,\V_2,\R_2)\F_2)$.
%dalal6: introduce "of variable"
%joe10
%\dfn\label{strongdomination}[Strong Domination of Variables]
%joe32: we're defining domination formodels,
%\dfn\label{strongdomination}\fullv{[Strong Domination of Variables]}
\dfn\label{strongdomination}\fullv{[Strong domination]}
Let $M_1 = ((\U_1,\V_1,\R_1),\F_1)$ and $M_2 =
%joe10
%((\U_2,\V_2,\R_2),\F_2)$ be two causal models.
((\U_2,$ $\V_2,$ $\R_2),\F_2)$.
%joe7*:
% Define 
% dalal6
Let $\Pa_{M}(C)$ denote
the variables that are parents of $C$ in
%hana37
% (the causal graph corresponding to)
$M$.
% dalal6: introduce rye phrase of strong domination in words
%We say  $M_1 \succeq_C M_2$, if the
%joe10
%We say that  $M_1$ strongly dominates $C$ over $M_2$, denoted
$M_1$ \emph{strongly dominates $M_2$ with respect to $C$}, denoted
$M_1 \succeq_C M_2$, if the 
following  conditions hold:
% dalal: updated -- update numbering
\begin{description}
%joe3
%\item[MI1$_{M_1,M_2,C}$.] All the variables in $\Pa_{M_2}(C)$ are in
%$\U_1 \cup \V_1$. 
%joe7*
  %\item[MI1$_{M_1,M_2,C}$.] $\Pa_{M_2}(C) \subseteq (\U_1 \cup \V_1)$.
%dalal26
%\item[MI1$_{M_1,M_2,C}$.] The parents of $C$ in $M_2$ are the
\item[MI1$_{M_1,M_2,C}$.] The parents of $C$ in $M_2$,  $\Pa_{M_2}(C)$, are the
  immediate $M_2$-ancestors of $C$ in $M_1$.
 %hana15 is the following true?
%joe13: no, definitely not.  All of C's parents in M_1 could be nodes
%in M_1 that aren't in M_2.  Each parent B of  C parents in M_2 is
%an ancestor of C in M_1, but it's not necessarily a parent of C in
%M_1.  But there is a path from B to C such that the only nodes in M_2
%on that path are B and C.  Does that make sense?
%That is, the set of parents of $C$ is the same in $M_1$ and $M_2$.
 % 
%  are ancestors   of $C$ in the causal graph
%  corresponding to $M_1$.
% add the following as MI2 and change the rest 
%joe7*: no longer needed!  I haven't changed the numbering yet, so
%that you can confirm the changes.  
%\item[MI2$_{M_1,M_2,C}$.]  If $B \in (\U_1 \cup \V_1) \cap (\U_2
%\cup \V_2)$,  then $B$ is an ancestor of $C$ in $M_1$ iff $B$ is an
%ancestor of $C$ in $M_2$. 
  \item[MI2$_{M_1,M_2,C}$.] Every path from an   exogenous variable to
    $C$ in $M_1$  goes through a variable in $\Pa_{M_2}(C)$.
  \item[MI3$_{M_1,M_2,C}$.]    Let $X = ((\U_1 \cup \V_1) \cap (\U_2
%joe3*: otherwise it's trivial
    %    \cup \V_2)$.  Then for all settings
        \cup \V_2))-\{C\}$.  Then for all settings  
$\vec{x}$ of the variables in   $\vec{X}$,  all values $c$ of $C$, all
    contexts $\vec{u}_1$ for $M_1$, and all contexts $\vec{u}_2$ for
$M_2$,
%joe10: line shaving
\fullv{we have}  
%dalal16
%\vspace{-5pt}
\[\begin{array}{l}
%joe20: we have the space to put this on one line
%\hspace{-1cm}  (M_1,\vec{u}_1) \satt [\vec{X} \gets \vec{x}](C=c)
%\mbox{ iff } \\ 
%\hspace{1cm}  (M_2,\vec{u}_2) \satt [\vec{X} \gets \vec{x}](C=c).
(M_1,\vec{u}_1) \satt [\vec{X} \gets \vec{x}](C=c)
\mbox{ iff } 
(M_2,\vec{u}_2) \satt [\vec{X} \gets \vec{x}](C=c).
\end{array}
\]
%dalal16
%\vspace{-7pt}
%(There is an abuse of notation here; see below.)
\end{description}
\edfn

%dalal22: to address reviewer 2:M strongly dominates M' w.r.t. *all*
%variables C if and only if M = M'?
%joe32*: It's true that if M strongly dominates M' and they have the
%same set of variables, then M=M'.  However, I don't see any benefit
%in saying this.  I view it as a distraaction.  Let's discuss
%In the case $M_1$ and $M_2$ have the same set of endogenous and
%exogenous variables --- 
%dalal23 open issue

%joe3: moved this below and rewrote.  (I think it's out of place here.)
%It is important to note  that the above conditions are with respect to
%individual variables appearing  
%in the model as oppose to being conditions over the models
%themselves. Thus  model $M_1$ may be more informative about one
%variable where as model $M_2$  may be so about another.
%joe3
%Note that if MI2$_{M_1,M_2,C}$ holds and, for example, $B$ is a parent
%joe7
%If MI2$_{M_1,M_2,C}$ holds and, for example, $B$ is a parent
If MI1$_{M_1,M_2,C}$ holds and, for example, $B$ is a parent
%joe3: the u's are a holdover from the previous version
%of $C$ in $(M_2,\vec{u}_2)$,
of $C$ in $M_2$, 
then there may be a variable $B'$ on the path from $B$ to $C$ in
%joe3
%$(M_1,\vec{u}_1)$.  Thus, $M_1$ has in a sense more detailed information than
$M_1$.  Thus, $M_1$ has in a sense more detailed information than
$M_2$ about the causal paths leading to $C$.  
%joe7
%MI1$_{M_1,M_2,C}$ and MI2$_{M_1,M_2,C}$ are not by themselves
MI1$_{M_1,M_2,C}$ is not by itself
%joe3
%enough to say that $(M_1,\vec{u}_1)$ and $(M_2,\vec{u}_2)$ agree on
enough to say that $M_1$ and $M_2$ agree on
the causal relations for $C$.  This is guaranteed by
MI2$_{M_1,M_2,C}$ and MI3$_{M_1,M_2,C}$.  MI2$_{M_1,M_2,C}$ says that the
variables in $\Pa_{M_2}(C)$ screen off 
$C$ from the exogenous variables in $M_1$.  (Clearly the variables 
in $\Pa_{M_2}(C)$ also screen off $C$ from the exogenous variables in $M_2$.)
%joe3
%It follows that if $(M_1,\vec{u}_1) \satt [\Pa_{M_2} \gets
It follows that if $(M_1,\vec{u}_1) \satt [\Pa_{M_2}(C) \gets
  \vec{x}](C=c)$ for some context $\vec{u}_1$, then
%joe3
%$(M_1,\vec{u}) \satt [\Pa_{M_2} \gets \vec{x}](C=c)$ for all contexts
%$\vec{u}$ in $M_1$, and similarly for $M_2$.  Once we have this, then 
$(M_1,\vec{u}) \satt [\Pa_{M_2}(C) \gets \vec{x}](C=c)$ for all contexts
$\vec{u}$ in $M_1$, and similarly for $M_2$.  In light of this
observation, it follows that
MI3$_{M_1,M_2,C}$ assures us that $C$ satisfies the same 
%joe3
%causal relations in both models.  There is an abuse of
causal relations in both models.  
%dalal4: add a note on notation  $M_1 \not \succeq_C M_2$
%joe12
%We write  $M_1 \not \succeq_C M_2$ when any of the conditions
We write  $M_1 \not \succeq_C M_2$ if any of the conditions 
above does not hold.
Two technical comments regarding Definition~\ref{strongdomination}:
First, note that in MI3 we used the fact that we allow the $\vec{X}$ in
formulas of the form $[\vec{X}\gets \vec{x}]\phi$ to include exogenous
variables, since some of the parents of $C$ may be exogenous.
Second, despite the suggestive notation, $\succeq_C$ is not a partial
order.  In particular, it is not hard to construct examples showing
that it is not transitive.  However, $\succeq_C$
%joe8
%is a partial order for compatible models (see the proof of
is a partial order on compatible models (see the proof of
Proposition~\ref{oplusproperties}), which is the only context in which
%joe11
%we are interested in transitivity, so the abuse of notation is
%somewhat justified.
\shortv{we are interested in transitivity.}
\fullv{we are interested in transitivity, so the abuse of notation is
somewhat justified.}

%joe3: moved from above
%joe7
%Note that we have a partial order $\succeq_C$ for each variable $C$
Note that we have a relation $\succeq_C$ for each variable $C$
that appears in both $M_1$ and $M_2$.
Model $M_1$ may be more informative than $M_2$ with respect to $C$
whereas $M_2$ may be more informative than $M_1$ with respect to another
variable $C'$.
%joe10: line shaving
%Roughly speaking, we take $M_1$ and $M_2$ to be \emph{strongly
%dalal20: In contrast with partially compatible
%Roughly speaking, $M_1$ and $M_2$ are \emph{strongly
Roughly speaking, $M_1$ and $M_2$ are \emph{fully
  compatible} if for each variable $C \in \V_1 \cap \V_2$, either $M_1
\succeq_C M_2$ or $M_2 \succeq_C M_1$.
%joe10: it actually doesn't say that
%This says that one of the two
%models has strictly more information about $C$ than the other.
%joe3
%so in the combined model, we can take 
%joe10: line shaving
%This allows us to combine $M_1$ and $M_2$ by taking
%dalal23
%We then combine $M_1$ and $M_2$ by taking
We then merge $M_1$ and $M_2$ by taking
the equations for $C$ to be determined
%joe3
%by the model that has more information.
by the model that has more information about $C$.  
%dalal25
Consider another example demonstrating the notion of strong dominance,
%joe35
%taken from   \cite{BDL14} .
taken from   \cite{BDL14}.

\begin{example}\label{ex:famine}
  %dalal23
  \normalfont
%joe34
  \cite{BDL14} 
An aid  agency consults  two experts about causes of famine in  a region.   
%joe3*: technically, your condfounding what the variable denotes (amount of
%rainfall) with a specific value of the variable  (low rainfall).  I
%rewrote this to disentangle.
%joe3
% Both experts agree  that the shortage of rainfall $R$ leads to poor
 % crop yield  $Y$.
  Both experts agree  that the amount of rainfall ($R$) affects 
 crop yield  ($Y$). Specifically, a shortage of rainfall leads to poor
 crop yield.
%  Expert $1$ says that  poor crop yield  $Y$ and disruptive political
 %  conflict $P$ are, combined, the main causes of famine.
%joe3*: I think these are backwards (at least, they're inconsistent
%with the figure)
 %  Expert $1$ says that  poor crop yield  $Y$ and disruptive political
%  conflict $P$ are, combined, the main causes of famine.  
%  Expert $2$ on the other hand considers   political conflict  as a
%  source for the disruption of agricultural production on crop
 %  yields,
 Expert $2$ says that political conflict ($P$) can also directly
 affect famine.  Expert 1, on the other hand, says that $P$ affects
 $F$ only via $Y$.
%  Expert $2$ on the other hand considers   political conflict  as a
%  source for the disruption of agricultural production on crop yields, 
 %  which in turn causes  famine.
 % There are depicted diagrammatically in Figure \ref{fig4}.
The experts' causal graphs are depicted in Figure \ref{fig4}, where
the graph on the left, $M_1$, describes expert 1's model, while
the graph on the right, $M_2$, describes expert 2's model.
%joe3*: added; is this true?
%joe4
%These graphs already appear in the work of Richard et
%al. \shortcite{Bradley:2014}.
%joe11: line shaving
\fullv{These graphs already appear in
%joe14
  %  the work of Bradley et al. \shortcite{BDL14}.}
the work of BDL.}
%joe3: moved up from below
In these graphs (as in many other causal
graphs in the literature),
%joe11
%the exogenous variables are omitted; all the
\fullv{the} exogenous variables are omitted; all \fullv{the}
     variables are taken to be endogenous. 
%joe3: moved
%\end{example}
 %dalal16
 %\vspace{-10pt}
 \begin{figure}[h]
\begin{center}
%joe18
%\setlength{\unitlength}{.13in}
%dalal16: illegal commands 
%joe20: I enlarged all the pictures.  We may need to make some
%adjustments as a consequence, but I didn't do this.
  %  \fullv{\setlength{\unitlength}{.13in}}
    \fullv{\setlength{\unitlength}{.18in}}
\shortv{\setlength{\unitlength}{.09in}}
%dalal22
%\begin{picture}(7,7)
\begin{picture}(13,7)
%joe3: made this change globally
 %dalal16:
%  \put(2,0){\circle*{.2}}
%  \put(2,0){\circle*{.3}}
%\put(4,6){\circle*{.3}}
%\put(0,3){\circle*{.3}}
%\put(4,3){\circle*{.3}}
%\put(4,6){\vector(0,-4){3}}
%\put(0,3){\vector(3,0){4}}
%\put(4,3){\vector(-2,-3){2}}
%\put(2.7,-.2){\scriptsize{$F$}}
%\put(-1,2.8){\scriptsize{$P$}}
%\put(4.45,2.8){\scriptsize{$Y$}}
%\put(4.45,5.8){\scriptsize{$R$}}
%\end{picture}
%\hspace{.3cm}
%\begin{picture}(8,7)
%\put(2,0){\circle*{.3}}
%\put(4,6){\circle*{.3}}
%\put(0,3){\circle*{.3}}
%\put(4,3){\circle*{.3}}
%\put(4,6){\vector(0,-4){3}}
%\put(0,3){\vector(2,-3){2}}
%\put(4,3){\vector(-2,-3){2}}
%\put(2.7,-.2){\scriptsize{$F$}}
%\put(-1,2.8){\scriptsize{$P$}}
%\put(4.45,2.8){\scriptsize{$Y$}}
%\put(4.45,5.8){\scriptsize{$R$}}
%dalal23 
\thicklines
%\put(4.8,2.5){\circle*{.2}}
\put(5,2.5){\circle*{.4}}
\put(0,5){\circle*{.4}}
%\put(0,0){\circle*{.4}}
\put(0.05,0){\circle*{.4}}
\put(-5,5){\circle*{.4}}
%\put(-5,0){\vector(1,0){5}}
\put(-5,5){\vector(1,0){5}}
%\put(0,0){\vector(2,1){5}}
\put(0,0){\vector(0,1){5}}
\put(0,5){\vector(2,-1){5}}
\put(-6,5.5){\scriptsize{$R$}}
\put(-.2,-1){\scriptsize{$P$}}
\put(-.2,5.5){\scriptsize{$Y$}}
\put(5.2,2.3){\scriptsize{$F$}}
\put(-3.4,-2){\scriptsize{$M_1$}}
\end{picture}
%dalal22
%\hspace{3.5cm}
\hspace{1cm}
%dalal22
%\begin{picture}(-4,7)
\begin{picture}(0,7)
%dalal23 
\thicklines
%\put(4.8,2.5){\circle*{.2}}
\put(5,2.5){\circle*{.4}}
\put(0,5){\circle*{.4}}
\put(0,0){\circle*{.4}}
\put(-5,5){\circle*{.4}}
%\put(-5,0){\vector(1,0){5}}
\put(-5,5){\vector(1,0){5}}
%\put(0,0){\vector(2,1){5}}
%\put(0,0){\vector(0,1){5}}
\put(0,5){\vector(2,-1){5}}
\put(0,0){\vector(2,1){5}}
\put(-6,5.5){\scriptsize{$R$}}
\put(-.2,-1){\scriptsize{$P$}}
\put(-.2,5.5){\scriptsize{$Y$}}
\put(5.2,2.3){\scriptsize{$F$}}
\put(3.4,-2){\scriptsize{$M_2$}}
\end{picture}
\end{center}
%joe8
%\caption{Two expert models on Famine.}\label{fig4}
%joe20*: we need labels M_1 and M_2, since we refer to them in the example.
%joe21*: this still needs to be done
%dalal17: done
%\caption{Two expert models of famine.}\label{fig4}
\caption{Two expert's models of famine.}\label{fig4}
\end{figure}
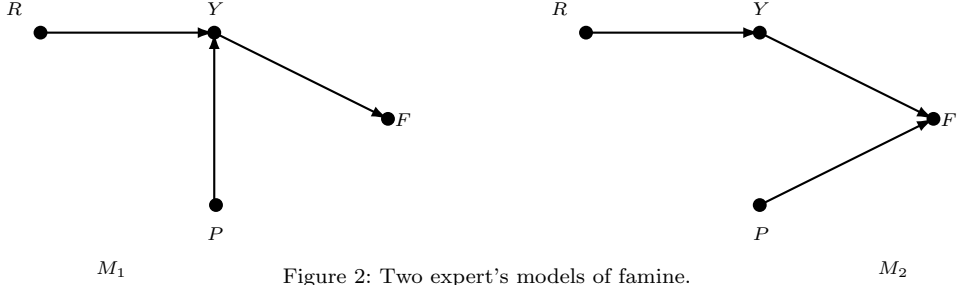
%hana12
%dalal16
%\vspace{-0.2cm}
%joe14*: this was a problem pointed out by a reviewer.
%MI1$_{M_1,M_2,F}$ does not hold, since $P$ is not 
%  an $M_2$-immediate ancestor of $F$ in $M_1$.   However,
%   MI1$_{M_2,M_1,F}$ does hold.
%    MI1$_{M_1,M_2,Y}$ holds, but
%      MI1$_{M_2,M_1,Y}$ does not, since $P$ is not an $M_1$-immediate
%joe20
 % Neither MI1$_{M_1,M_2,F}$ nor MI1$_{M_2,M_1,F}$, since $P$ is not
  Neither MI1$_{M_1,M_2,F}$ nor MI1$_{M_2,M_1,F}$ holds, since $P$ is not 
 an $M_2$-immediate ancestor of $F$ in $M_1$.   Similarly,
   neither MI1$_{M_1,M_2,Y}$ nor MI1$_{M_2,M_1,Y}$ holds, since $P$ is
   not an $M_1$-immediate 
      ancestor of $Y$ in $M_2$ (indeed, it is not an ancestor at all).
%  an $M_2$-immediate ancestors of $F$ in $M_1$.   However,
 %   MI1$_{M_1,M_2,F}$ does hold.
      %    F$ has the same set of ancestors in
%both models.
%dalal: check this holds
%joe3: it does hold
%MI2$_{M_1,M_2,F}$ is satisfied since every path from exogenous
    MI2$_{M_1,M_2,F}$ holds since every path in $M_1$ from an exogenous
   variable to $F$ goes through a variable 
   that is a parent of $F$ in $M_2$ (namely, $Y$);
%joe7: added
%joe12
%   MI2$_{M_2,M_1,F}$ does not holds (there is a path in $M_2$ to $F$
   MI2$_{M_2,M_1,F}$ does not hold (there is a path in $M_2$ to $F$
   via $P$ that does not go through a parent of $F$ in $M_1$).
%joe3: rewrote to sharpen it.  don't need to assume anything.
%   However MI3$_{M_1,M_2,F}$ is not
%   met since there may exist 
%   a setting to variables $\{R, P, Y\}$ such that $F$ holds in $M_1$
%   and not in $M_2$; for instance 
%   by  assuming $\F_1(Y) = P\vee R$ and $\F_2(F) = P\wedge Y$ in $M_2$, we have
%  $(M_1,\vec{u}_1) \satt [P=0,Y=1,R=0](F=0)$ but $(M_2,\vec{u}_2)
   %   \satt [P=0,Y=1,R=0](F=1)$.
%joe7
   %   However MI3$_{M_1,M_2,F}$ does not hold.  Since $P$ is a parent of
%joe7
%    However MI3$_{M_1,M_2,F}$ does not hold.  Since $P$ is a parent of
Although we are not given the equations, we also know that
    MI3$_{M_1,M_2,F}$ does not hold.  Since $P$ is a parent of
    $F$ in $M_2$ according to expert 2, there must be a setting $y$ of $Y$
 such that the value of $F$ changes depending on the value of $P$ if $Y=y$.
This  cannot be the case in $M_1$, since $Y$ screens off $P$ from
$F$.  It easily follows that taking $\vec{X}= (P,Y)$ we get a
  counterexample to MI3$_{M_1,M_2,F}$.
%joe7
  %  Therefore $M_1\not \succeq_C M_2$.
%joe12
  %  Therefore we have neither $M_1 \succeq_F M_2$ or
    Therefore, we have neither $M_1 \succeq_F M_2$ nor
  $M_2 \succeq_F M_1$.
%joe4
 % It is worth noting   failure of the relation
%  $M_1 \succeq_C M_2$ to hold, does not imply that $M_2 \succeq_C M_1$.
%joe7: already said this for F.
%  It is worth noting the fact that
% $M_1 \succeq_C M_2$ does not hold does not imply that $M_2 \succeq_C
%  M_1$ does.
  %joe3
  %joe7
  \eprf
\end{example}

 %
%joe3
  % While the above definition is
   While the definition of dominance given above is
   %dalal23: I am not that happy with saying "we may want " as 
   % a start for the justification . It's more than what we want...    
   %useful, it does not cover all cases where we may want to combine
useful, it does not cover all cases where a policymaker may want to merge
models.  Consider the following example, taken from the work of
%joe32
%Sampson, Winship, and Knight \citeyear{Sampson:2013}.
\citet{Sampson:2013}.

\begin{example}\label{domestic1}  
  %dalal23
  \normalfont
  Two experts have provided causal models regarding the
  causes of domestic violence.  
%joe2: note that I've switched the order of the experts.
%joe3
  %  According to the first expert, an arrest policy (\AP) may lessen
  %both  offenders' belief that their partners would report
%  any abuse to police (\PLS) and a victims' actual calling to
  %  report abuse
%  (\CA), though it may increase  domestic violence (\DV). The  
  According to the first expert, an appropriate arrest policy (\AP)
    may affect 
    both an offender's belief that his partner would report
%joe1    
    %any abuse to police (\PLS),
    any abuse to police (\PLS)
%joe3: not according to the model
%whether a victim would actually cal to report abuse (\CA), and
%joe12
%dalal13: reviewer suggested moving and 3 words forwards but disagree
and the amount of 
domestic violence (\DV).
%joe3: I'm not sure what ``randomized'' arrests are
%The  
% increase in \DV\   would also lead to an increase in victim's
% reporting which in turn  increases likelihood of randomized arresting (\RA).
The amount of domestic violence also affects the likelihood of
a victim calling to report abuse (\CA), which in turn
affects the likelihood of there being 
%hana15 added explanation of random arrest
%a random arrest (\RA).\footnote{Decisions on whether to arrest the
%offender in cases of domestic violence were randomized, leading to a
%drop in rates of re-offending.} 
%dalal13
a random arrest (\RA). (Decisions on whether to arrest the offender in
%joe14: ``leading to'' suggests causality.  Do we know that?  I just
%cut that phrase.
%cases of domestic violence were randomized, leading to a drop in rates
%of re-offending.)
cases of domestic violence were randomized.)

%a random arrest (\RA).
According to the second expert, 
%joe3
%randomized arrests (\RA), in the presence of domestic violence (\DV),
%has a positive influence on the reduction of  
\DV\ affects \RA\ directly, while \RA\
affects the amount  of
repeated violence (\RV) through both formal sanction (\FS) and
informal sanction on socially embedded individuals  (\IS).  
%joe3
%The annotated causal
%joe32
%Sampson et al.~\citeyear{Sampson:2013} use the following causal
\citet{Sampson:2013} use the causal
graphs  shown in Figure \ref{fig6}, which are annotated with the
direction of the influence 
%joe8
%(the only information provided by the experts) to describe the experts
(the only information provided by the experts) to describe the expert's
opinions.  
%joe2*: Dalal/Hana, can we add the +/- to the graph?
%dalal: done

%joe3: we need to move the - in front of DV in M_1, and add an
%anntoation (+) to the edge between DV and A in M_2.
%dalal13:
%dalal16
%\vspace{-5pt}
%joe18*: can we rotate these figures 90 degrees?  This would be more
%consistent with Figure 3 (as well as saving space).
\begin{figure}[h]
%joe18*: can we turn these 90 degrees? That would save space, and make
%it more consistent with Figure 3.
   \begin{center}
%joe20
     %     \setlength{\unitlength}{.09in}
          \setlength{\unitlength}{.18in}
\begin{picture}(8,13)
\thicklines
\put(0,12){\circle*{.4}}
\put(0,8){\circle*{.4}}
\put(0,4){\circle*{.4}}
\put(0,0){\circle*{.4}}
\put(3,10){\circle*{.4}}
\put(0,12){\vector(0,-1){4}}
\put(0,8){\vector(0,-1){4}}
\put(0,4){\vector(0,-1){4}}
\put(0,12){\vector(3,-2){3}}
\put(3,10){\vector(-3,-2){3}}
\put(-1.7,12){\scriptsize{$\AP$}}
\put(-1.5,7.8){\scriptsize{$\DV$}}
\put(0.7,5.8){$+$}
\put(-1,9.6){$-$}
\put(-1.1,3.2){\scriptsize{$\CA$}}
\put(-1.1,-0.2){\scriptsize{$\RA$}}
%dalal8
%joe21: moved up a bit to create some space above the caption
%\put(-.7,-2.0){\scriptsize{$M_1$}}
\put(-.7,-1.5){\scriptsize{$M_1$}}
\put(0.7,2){$+$}
\put(1.7,10.9){$-$}
\put(3.7,9.8){\scriptsize{$\PLS$}}
\put(1.7,8.3){$-$}
\qbezier(0,12)(-6,8)(0,4)
\put(0.07,3.9){\vector(1,-1){0}}
\put(-4,7.8){$-$}
%\put(0,11.8){$\DV$}
%\put(6.7,3.8){$\FS$}
%\put(-1.7,3.8){$\IS$}
%\put(0,7.8){$\RA$}
\end{picture}
\hspace{1.5cm}
\begin{picture}(8,13)
%dalal23
\thicklines
\put(3,12){\circle*{.4}}
\put(3,8){\circle*{.4}}
\put(6,4){\circle*{.4}}
\put(0,4){\circle*{.4}}
\put(3,0){\circle*{.4}}
\put(3,12){\vector(0,-1){4}}
\put(3,8){\vector(-3,-4){3}}
\put(3,8){\vector(3,-4){3}}
\put(6,4){\vector(-3,-4){3}}
\put(0,4){\vector(3,-4){3}}
%\put(0,4){\vector(3,-4){3}}
%\put(6,4){\vector(-3,-4){3}}
\put(3.8,-.2){\scriptsize{$\RV$}}
%joe21
%\put(2.1,-2.0){\scriptsize{$M_2$}}
\put(2.1,-1.5){\scriptsize{$M_2$}}
\put(1.3,12){\scriptsize{$\DV$}}
\put(4.9,5.8){$+$}
\put(6.7,3.8){\scriptsize{$\FS$}}
\put(4.9,1.8){$-$}
\put(-0.2,5.8){$+$}
\put(-1.7,3.8){\scriptsize{$\IS$}}
\put(-0.2,1.8){$-$}
\put(1.8,7.8){\scriptsize{$\RA$}}
\end{picture}
\end{center}
%joe8
%\caption{Experts on Domestic Violence.}\label{fig6}
%joe20: 
%   \caption{Expert's  models of domestic violence.}\label{fig6}
   \caption{Experts'  models of domestic violence.}\label{fig6}
\end{figure}
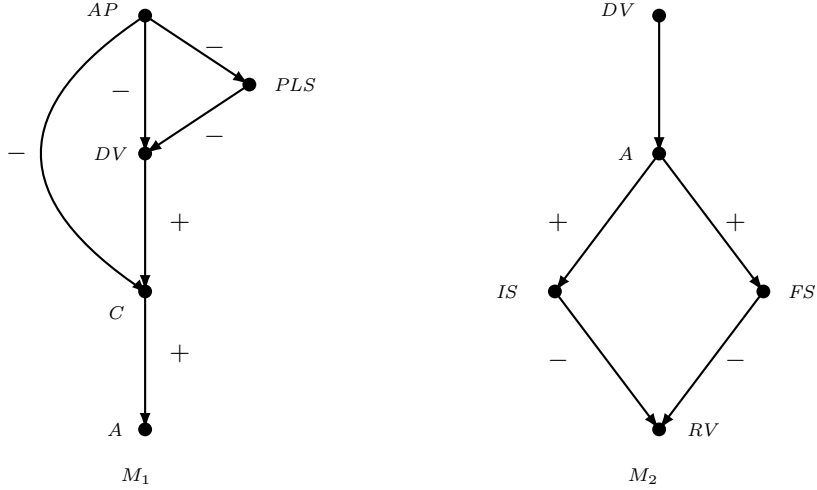
%hana12
%dalal16
%\vspace{-0.2cm}
%dalal:
%dalal8
%If we call the first expert's model $M_1$ and the second expert's
%model $M_2$, then it  is easy to check that, for
For the two common variables $\DV$ and $\RA$,  MI1$_{M_1,M_2,DV}$
and MI1$_{M_1,M_2,A}$ both hold.
%joe7: no longer needed
%Similarly MI2$_{M_1,M_2,DV}$
%%joe3
%%and MI2$_{M_1,M_2,A}$ also hold  since $\Pa_{M_2}(DV) = \emptyset$,
%and MI2$_{M_1,M_2,A}$ also hold, since $\Pa_{M_2}(DV) = \emptyset$,
%  $\Pa_{M_2}(A) = \{\DV\}$, and   
%$\DV$ is an ancestor of $A$ in $M_1$.
%joe3*
%However, neither MI2$_{M_1,M_2,DV}$ nor MI3$_{M_1,M_2,A}$ holds.
If the only variables that have exogenous parents are  $\AP$ in $M_1$ and
$\DV$ in $M_2$, and the set of parents of $\AP$ in $M_1$ is a subset
of the set of parents of $\DV$ in $M_2$, then MI2$_{M_1,M_2,DV}$ holds.
%joe3: cut
%of $\AP$  
%%dalal
%There is path 
%in $M_1$ from some exogenous variable to \RA\ that 
%that directly influences \CA\ without going  through the parent variable 
%\DV\ (the sole parent of   \RA\ in $M_2$).
%There is path 
%in $M_1$ from some exogenous variable that is a parent of $\AP$ both to 
%$\DV$ and to $\CA$, and thus a path to
%$\RA$ in $M_2$ that does not go through $\DV$.  
%Thus, even ignoring
%MI3, neither $M_1 \succeq_{DV} M_2$ or $M_1 \succeq_A M_2$ is true, 
%although there seems to be a sense in which $M_1$ has
%more information about $\DV$ and $A$ than $M_2$.
%dalal22 reviewer1 would like the structural equations for this example. 
% the problem is how to map +/-
%joe32: I don't see how we can give structural equations if Sampson et
%al. don't give us the information needed to do this.  I put in a
%little about this below
Sampson et al.~seem to be implicitly assuming this, and that MI3 holds, so they
%dalal23
% as Sampson also go straight to combining and dropping +/- when experts 
%for instance disagree I would have used combine but because
%we are discussing merging of model I decided to change... please
%revert if needed
%combine $M_1$ and $M_2$ to get the
merge $M_1$ and $M_2$ to get the 
causal graph shown in Figure~\ref{sampsoncom}.
%joe18: moved figure here
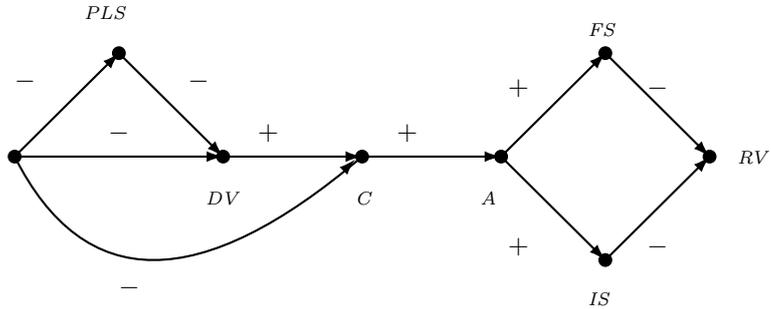
\begin{figure}[htb]
\vspace{0.8cm}
%joe12*: my latex complains about the -14pt here; I've removed
%it. Please check
%dalal9: sorted
%\shortv{\begin{wrapfigure}[13]{r}[-14pt]{0.2\textwidth}}
%dalal16:
%\shortv{\begin{wrapfigure}[14]{R}{0.13\textwidth}}
%\begin{floatingfigure}{0.2\textwidth}
\begin{center}
%\vspace{-\intextsep}
%\hspace*{-1.55\columnsep}
%joe20
  %\setlength{\unitlength}{.09in}
  \setlength{\unitlength}{.18in}
%dalal16:
%\begin{picture}(5,18)
%\put(6,18){\circle*{.2}}
%\put(3,20){\circle*{.2}}
%\put(3,16){\circle*{.2}}
%\put(3,12){\circle*{.2}}
%\put(3,8){\circle*{.2}}
%\put(6,4){\circle*{.2}}
%\put(0,4){\circle*{.2}}
%\put(3,0){\circle*{.2}}
%\put(3,20){\vector(3,-2){3}}
%\put(6,18){\vector(-3,-2){3}}
%\put(3,20){\vector(0,-1){4}}
%\put(3,16){\vector(0,-1){4}}
%\put(3,12){\vector(0,-1){4}}
%\put(3,8){\vector(-3,-4){3}}
%\put(3,8){\vector(3,-4){3}}
%\put(6,4){\vector(-3,-4){3}}
%\put(0,4){\vector(3,-4){3}}
%\put(1.7,17.9){$-$}
%
%\put(4.7,18.8){$-$}
%\put(4.7,16.5){$-$}
%\put(6.3,17.8){\scriptsize{$\PLS$}}
%\put(1.7,17.9){$-$}
%\qbezier(3,20)(-4,16)(3,12)
%\put(2.96,11.92){\vector(1,-1){0}}
%\put(-2.4,16){$-$}
%\put(3.8,-.2){\scriptsize{$\RV$}}
%\put(0.6,19.8){\scriptsize{$\AP$}}
%\put(3.3,13.8){$+$}
%\put(0.6,15.8){\scriptsize{$\DV$}}
%\put(3.3,9.8){$+$}
%\put(1.2,11.3){\scriptsize{$\CA$}}
%\put(4.9,5.8){$+$}
%\put(6.7,3.8){\scriptsize{$\FS$}}
%\put(4.9,1.8){$-$}
%\put(-0.2,5.8){$+$}
%\put(-1.7,3.8){\scriptsize{$\IS$}}
%\put(-0.2,1.8){$-$}
%\put(1,7.8){\scriptsize{$\RA$}}
%\end{picture}
\begin{picture}(7,6)
%dalal23
\thicklines
\put(-5,3){\circle*{.4}}
\put(-2,6){\circle*{.4}}
\put(1,3){\circle*{.4}}
\put(5,3){\circle*{.4}}
\put(9,3){\circle*{.4}}
\put(12,6){\circle*{.4}}
\put(12,0){\circle*{.4}}
\put(15,3){\circle*{.4}}
\put(-5,3){\vector(1,1){3}}
\put(-5,3){\vector(1,0){6}}
\put(-2,6){\vector(1,-1){3}}
\put(1,3){\vector(1,0){4}}
\put(5,3){\vector(1,0){4}}
\put(9,3){\vector(1,-1){3}}
\put(9,3){\vector(1,1){3}}
\put(12,6){\vector(1,-1){3}}
\put(12,0){\vector(1,1){3}}
\put(-5,5){$-$}
\put(-2.3,3.5){$-$}
\put(-3,7){\scriptsize{$\PLS$}}
\put(0,5){$-$}
\put(.5,1.6){\scriptsize{$\DV$}}
\put(2,3.5){$+$}
\qbezier(-5,3)(-2,-3)(5,3)
\put(4.8,2.92){\vector(1,1){0}}
\put(-2,-1){$-$}
\put(6,3.5){$+$}
\put(4.8,1.6){\scriptsize{$\CA$}}
\put(8.4,1.6){\scriptsize{$\RA$}}
\put(11.5,6.5){\scriptsize{$\FS$}}
\put(11.5,-1.3){\scriptsize{$\IS$}}
\put(9.2,4.8){$+$}
\put(9.2,.2){$+$}
\put(13.2,4.8){$-$}
\put(13.2,.2){$-$}
\put(15.8,2.8){\scriptsize{$\RV$}}

%\put(3.3,9.8){$+$}

%\put(4.9,5.8){$+$}

%\put(4.9,1.8){$-$}
%\put(-0.2,5.8){$+$}
%\put(-1.7,3.8){\scriptsize{$\IS$}}
%\put(-0.2,1.8){$-$}

\end{picture}
\end{center}
%\vspace{-7pt}
%joe8
%\caption{Combined  Experts' Model on Domestic Violence.}\label{sampsoncom}
%dalal23
%\caption{Combined  experts' model of domestic violence.}\label{sampsoncom}
%joe33
%\caption{Merged  experts' model of domestic violence.}\label{sampsoncom}
\caption{The result of merging experts' model of domestic violence.}\label{sampsoncom}
%\shortv{\end{wrapfigure}}
%\fullv{\end{figure}}
\end{figure}

%dalal:
%joe18
%Note that in  $M_2$ of Figure~\ref{fig6}, the expert does not state
%joe32
%Note that model $M_2$ in Figure~\ref{fig6} does not state
%how \DV\ influences $A$.
Sampson et al.~do not provide structural equations.  Moreover, for
edges that do not have a $+$ or $-$  annotation, such as the edge from
$\DV$ to $A$ in Figure~\ref{fig6}, we do not even know qualitatively
what the impact of interventions is.  
%joe18
%We can view this as the expert  being uncertain about how
%\DV\ influences $A$.  We can capture this uncertainty by viewing the
%dalal22 fixed typo
%Presumably, this respresents the expert's undertainty. 
%joe32
%Presumably, this represents the expert's uncertainty.
Presumably, the lack of annotation represents the expert's uncertainty. 
We can capture this uncertainty by viewing the
expert as having a 
probability on two models that disagree on the direction of \DV's
influence on $A$
(and thus are incompatible because they disagree
%joe12
%on the equations). We discuss in Section~\re
%f{sec:experts} how
on the equations). We discuss in Section~\ref{sec:experts} how 
%joe7
%this case may be  handled.
such uncertainty can be handled.
%joe3*: add figure here
%joe7: I like marking the end of examples.  I would prefer also not to
%use italics for examples, but we can worry about that later
\eprf
\end{example}
%hana9: I don't see any "gutter" anywhere in the figure in my pdf!
%joe10*: in my pdf, it says ``gutter'' in the upper left-hand corner
%othe figure.
%joe11: for some reason it doesn't seem to work in my environment.
%It's fine as long as you generate the pdf!
%dalal8 reduced gap between figure and text
%dalal9: tried a new environment. It works in mine.
%\begingroup
%dalal16 need to be removed
% \setlength\intextsep{4pt}
%\shortv{\setlength\columnsep{8pt}}
%dalal16:
%\fullv{\begin{figure}}
%\endgroup
%\end{floatingfigure}
%hana12
%dalal13:
%joe18
%\shortv{\vspace{0.2cm}}
%joe3*: not quite
%The fact that there is a path to $\RA$ in $M_1$ that does not go through
%joe7
%But suppose that
Suppose that some parent of $\AP$
%hana15 added ``or AP itself''
(or $\AP$ itself)
in $M_1$ is not a 
parent of $\DV$ in $M_2$.  Then, 
%in $M_1$, it is possible to change the value of $\DV$ by intervening
in $M_1$, it may be possible to change the value of $\DV$ by intervening
on $\AP$, while keeping the values of all the exogenous variables that
%joe7
are
parents of $\DV$ in $M_2$ fixed.  
%joe3*: not true
%$\DV$ suggests that 
%the value of $\RA$ can be changed by intervening on $\AP$ while
%keeping all the 
%variables in $\Pa_{M_1}(\RA)$ (namely $\DV$) fixed. This will seem like an
%inexplicable change in the value of $A$ from the perspective of the
This will seem like an 
inexplicable change in the value of $\DV$ from the perspective of the
%joe12
%expert who is using model $M_2$.
second expert.
%joe3
%Similarly, there can be inexplicable changes to the value of $\DV$.
If the second expert had been aware of such
possible changes, she 
%joe7
%surely would have added additional variables to $M_2$ to model them
%into account.  One way to explain the fact that no changes were
surely would have added additional variables to $M_2$ to capture this 
%joe12
%situation.  One way to explain the fact that no changes were
situation.  One explanation of the fact that no changes were
observed is that the second expert was
working in a setting where the values of all variables that she cannot
%joe12
%affect by an intervention is determined by some default setting of
%exogenous variables that she is not aware of (or is not modeling).  
affect by an intervention are determined by some default setting of
%dalal9:
%exogenous variables that she is not aware of (or not modeling).  
exogenous variables of which  she is not aware (or not modeling).  

 %hana37 added v*
%joe39*: rewrote; we need to explain the role of v^* better.
%Moreover, we actually don't need v^* in M1, since we're assuming
%strongly recursive models (see my earlier joe39).  Sorry about that.
%In light of this discussion, we re-write our definition of strong
% domination,  denoting this default setting by $\vec{v}^*$. 
% First, we introduce $\vec{v}^*$ to the condition {\bf MI1} as follows: 
Given models $M_1$ and $M_2$, 
we now want to define a notion of weak domination relative to a
default setting $\vec{v}^*$ of the exogenous variables in $\U_1 \cup
\U_2$.  We say that contexts    
 $\vec{u}_1$ for $M_1$ and $\vec{u}_2$ for
    $M_2$ are \emph{compatible with $\vec{v}^*$} if $\vec{u}_1$ and
$\vec{u}_2$ agree on the variables 
    in $\U_1 \cap \U_2$, $\vec{u}_1$ agrees with $\vec{v}^*$ on the
    variables in $\U_1 - \U_2$, and $\vec{u}_2$ agrees with
    $\vec{v}^*$ on the variables in $\U_2 - \U_1$.  
\dfn\label{weakdomination}\fullv{[Domination relative to a
    default setting]} 
%\dfn\label{weakdomination}\fullv{[Weak Domination of Variable]}
%joe18
%Let $\vec{v}^*$ be a default setting for the variables in $\V^*$.
Let $\vec{v}^*$ be a default setting for the variables in $M_1$ and $M_2$.
%dalal
% Say that 
%$M_1$ contains at least as much information about $C$ as $M_2$
%dalal6:  introduce notion of weak domination rather than more and less information
%$M_1$ is said to contain at least as much information about $C$ as $M_2$
%joe10
%$M_1$ is said to weakly dominate $C$ over $M_2$
%hana17 replaced weak domination by domination
$M_1$ \emph{dominates} $M_2$ with respect to $C$
%$M_1$ \emph{weakly dominates} $M_2$ with respect to $C$
%joe3
%relative to $\vec{v}^*$ if  MI1$_{M_1,M_2,C}$ holds and, in addition,
%joe7: should have looked at Dalal's comments more carefully.  She
%observed that we never defined the notation. 
%relative to $\vec{v}^*$ if  MI1$_{M_1,M_2,C}$ and MI2$_{M_1,M_2,C}$ hold,
relative to $\vec{v}^*$, denoted $M_1 \succeq_C^{\vec{v}^*} M_2$, if
%hana37 added v*
%joe39:no need; sorry
%MI1$_{M_1,M_2,C,\vec{v}^*}$ holds,
MI1$_{M_1,M_2,C}$ holds,
and, in addition,
the following condition (which can be viewed as a replacement for 
MI2$_{M_1,M_2,C}$ and  MI3$_{M_1,M_2,C}$) holds:
% dalal: was X = (\U_1 \cup \V_2).. changed to X = (\U_1 \cup \V_1)
\begin{description}
%joe7
%\item[MI4$_{M_1,M_2,C,\vec{v}^*}$]  Let $X = (\U_1 \cup \V_1) \cap  (\U_2
  \item[MI4$_{M_1,M_2,C,\vec{v}^*}$]  Let $\vec{X} = (\U_1 \cup \V_1) \cap (\U_2
%joe3
  %  \cup \V_2)$.  Then for all settings
   \cup \V_2) - \{C\}$.  Then for all settings
  $\vec{x}$ of the variables in   $\vec{X}$,  all values $c$ of $C$, and
    all contexts $\vec{u}_1$ for $M_1$ and $\vec{u}_2$ for
    $M_2$
%joe39
%    such that $\vec{u}_1$ and $\vec{u}_2$ agree on the variables
%    in $\U_1 \cap \U_2$, $\vec{u}_1$ agrees with $\vec{v}^*$ on the
%    variables in $\U_1 - \U_2$, and $\vec{u}_2$ agrees with
    %    $\vec{v}^*$ on the variables in $\U_2 - \U_1$, we have
    that are compatible with $\vec{v}^*$, we have
\[\begin{array}{l}
%joe20: put on one line; we have the space.
%\hspace{-1cm}
%(M_1,\vec{u}_1) \satt [\vec{X} \gets \vec{x}](C=c) \mbox{ iff } \\
%\hspace{1cm}  (M_2,\vec{u}_2) \satt [\vec{X} \gets \vec{x}](C=c).
(M_1,\vec{u}_1) \satt [\vec{X} \gets \vec{x}](C=c) \mbox{ iff } 
(M_2,\vec{u}_2) \satt [\vec{X} \gets \vec{x}](C=c).
\end{array} 
 \]
\end{description}
\edfn

It is easy to see that $\succeq_C$ is a special case of
$\succeq_C^{\vec{v}^*}$:

%dalal11: correct formatting from \M_2 to M_2
\lem\label{lem:specialcase}  If $M_1 \succeq_C M_2$, then, for all
%joe18
%default settings $\vec{v}^*$ of the variables in $\V^*$, we have
default settings $\vec{v}^*$ of the variables in $M_1$ and $M_2$, we have 
$M_1 \succeq_C^{\vec{v}^*} M_2$. \elem

%dalal22 we should address here reviewer 1's comment
% "is still unclear how restrictive this requirement is in practice. Intuitively, MI3 and MI4/MI4' are 
%quite restrictive because at a high level it says that despite the
%structure and variables of the  
%two models, they more or less make similar judgements on C=c."
%joe32*: I'm nor sure what we can/should say beyond what we now say.
%We should discuss this.
%dalal23 open issue
%joe3
%\prf Still to come. \eprf
%joe7: added proof
\fullv{\prf Suppose that $M_1 \succeq_C \M_2$.  Fix default values
  $\vec{v}^*$.  Clearly MI4$_{M_1,M_2,C,\vec{v}^*}$ is a special case
  %dalal23
  %joe33
  of MI2.  Thus, $M_1 \succeq_C^{\vec{v}^*} \M_2$. \eprf}
%  of MI2.  Thus, $M_1 \succeq_C^{\vec{v}^*} \M_2$. $\Box$}
  
% dalal: notation $\succeq_c^*{\vec{v}^*} with \succeq_c^* no explained before. Typo?
%dalal2: corrected the above replaced $\succeq_c^{\vec{v}^*}$ ->
%succeq_C^{\vec{v}^*} 
In light of Lemma~\ref{lem:specialcase}, we give all the definitions
in the remainder of the paper using $\succeq_C^{\vec{v}^*}$.  
% dalal9:
%All the
%definitions and the technical results then apply without change if we
%joe12: it seems strange to say that a definition hold
%All definitions and technical results hold if we
All the technical results hold if we
%joe3
%replace $\succeq_C^{\vec{v}^*}$ throughout.
replace $\succeq_C^{\vec{v}^*}$ by $\succeq_C$ throughout.

%dalal22 split as suggested by reviewer
%joe10
%\begin{definition}\label{combat}[Compatibility of Causal Models]
%dalal22 distinguish from later compatibility
%joe32: it seems strange to see ``Full Compatibility'' when the
%definition just talks about ``compatible''. We should either change
%the notion we're defining to ``fully compatible'' or revert back to
%just ``Compatible''.  My preference is for the former, but I could
%live with the latter.
%\begin{definition}\label{combat}\fullv{[Full Compatibility of Causal
%      Models]}
\begin{definition}\label{combat}\fullv{[Full compatibility of causal
      models]} 
If $M_1 = ((\U_1,\V_1,\R_1),\F_1)$ and $M_2 = ((\U_2,\V_2,$ $\R_2),\F_2)$, 
%dalal23
%then $M_1$ and $M_2$ are \emph{compatible}
%joe34: we have to talk about $v^*$
%then $M_1$ and $M_2$ are \emph{fully compatible}
then $M_1$ and $M_2$ are \emph{fully compatible with respect to
 default setting $\vec{v}^*$}
if (1) for all variables $C \in (\U_1 \cup \V_1) \cap
(\U_2 \cup \V_2)$, we have $\R_1(C) = \R_2(C)$ and (2) for all variables
%joe20*: It is important here that we allow C to be exogenous in one
%model, otherwise we don't get acyclicity.  For example, if we have
%models M_1 and M_2 that both contain the variables A, B, C, D, but in
%M_1, A and C are exogenous, A is a parent of B, and C is a parent of
%D, while in M_2, B and D are exogenous, B is a parent of C, and D is
%a parent of A, then under the current definition, M_1 and M_2 are
%compatible, but M_1 \oplus M_2 is cyclic.  I have now corrected this.
%(My student Meir Friedenberg pointed out this problem.)
%$C \in \V_1 \cap \V_2$,
$C \in (\V_1 \cap \V_2) \cup (\V_1 \cap \U_2) \cup (\V_2 \cap \U_1)$,
either $M_1 \succeq_C^{\vec{v}^*} M_2$ or $M_2 \succeq_C^{\vec{v}^*} M_1$.
%joe34
We say that $M_1$ and $M_2$ are \emph{fully compatible} if $\U_1 =
\U_2$ (so we can ignore the default setting).
\end{definition} 
 %
%dalal22 added fully
%joe32: Again, I'm not happy about the inconsistency
%\begin{definition}\label{mergecombat}\fullv{[Merging Compatible Causal Models]}
%\begin{definition}\label{mergecombat}\fullv{[Merging Fully Compatible
%      Causal Models]}
%dalal23
%\begin{definition}\label{mergecombat}\fullv{[Combining fully compatible
\begin{definition}\label{mergecombat}\fullv{[Merging fully compatible
      models]} 
If $M_1 = ((\U_1,\V_1,\R_1),\F_1)$ and $M_2 = ((\U_2,\V_2,\R_2),\F_2)$ are
%joe8
%compatible, we define $M_1 \oplus M_2$ to be the causal model
%dalal22
%compatible, then  $M_1 \oplus M_2$ is the causal model
%joe32
%compatible, then the merged model $M_1 \oplus M_2$ is the causal model
%dalal23
%compatible, then the combined model $M_1 \oplus M_2$ is the causal model
%joe34: again, we need to say something about v^*
%fully compatible, then the merged model $M_1 \oplus M_2$ is the causal
fully compatible with respect to $\vec{v}^*$, then the merged model
$M_1 \oplus^{\vec{v}^*} M_2$ is the causal 
model 
$((\U,\V,\R),\F)$, where
\begin{itemize}
\item $\U = \U_1 \cup \U_2 - (\V_1 \cup \V_2)$;
\item $\V = \V_1 \cup \V_2$;
  \item if $C \in \U_1 \cup \V_1$, then $\R(C) = \R_1(C)$, and if $C
    \in \U_2 \cup \V_2$,
    then $\R(C) = \R_2(C)$;
  \item if $C \in \V_1 - \V_2$ or if both $C \in \V_1 \cap \V_2$ and
    $M_1 \succeq_C^{\vec{v}^*} M_2$,
    then $\F(C) = \F_1(C)$; 
    if $C \in \V_2 - \V_1$ or if both $C \in \V_1 \cap \V_2$ and
    $M_2 \succeq_C^{\vec{v}^*} M_1$, 
    then $\F(C) = \F_2(C)$.%
    %joe8: added footnote
    %dalal4: typo
    %\footnote{There is a slight abuse of notation here.  Technicaly,
    %dalal16: Joe's suggestion
%    \footnote{There is a slight abuse of notation here.  Technically,
%      $\F(C)$ is a function whose domain is $\times_{U \in \U} \R(U)$
%      (i.e., it gets as input values for all the variables in $\U$),
%      while $\F_i(C)$ is a function whose domain is $\times_{U \in
%        \U_i} \R(U)$, for $i = 1, 2$,  so unless $\U = \U_i$, we
%      cannot have $\F(C) = \F_i(C)$.  However, we can view the domain
%      of $\F_i(C)$ as being the values of the variables in
%      $\Pa_i(C)$, since its output is independent of the values of
%      all other variables.
%%joe12
%%      When we write $\F(C) = \F_i(C)$, we are
%%      implicitly taking the parents of $C$ in $M_1 \oplus M_2$ to be
%%      the parents of $C$ in $M_i$, so assuming that the output of
%%      $\F(C)$ is independent  the values of all other variables.}
%Since in this case the parents of $C$ in $M_1 \oplus M_2$ are
%the parents of $C$ in $M_i$, the notation is well defined.}
%joe18
%    \footnote{Here, we are abusing notation and viewing $\F_i(C)$ as a
    \footnote{We are abusing notation here and viewing $\F_i(C)$ as a
    function from the 
values of the parents of $C$ in $M_i$ to the value of $C$, as opposed to a
function from all the values of all variables other than $C$ to the value
of $C$.}
\end{itemize}
%joe34*: we need to check our usage of \oplus throughout hte paper, to
%make sure it's consistent, and perhaps add some \vec{v}^* superscripts
We write  $M_1 \oplus M_2$ if $\U_1 = \U_2$.
\end{definition}   

%dalal22 addressing reviewer 1's comment
Note that we assume that when experts use the same variable, they are
%dalal23
%referring to the same thing within the same domain. Our approach does
referring to the same phenomenon within the same domain. Our approach does
%joe32: I'm not sure what ``renaming'' is
%not deal with the renaming of variables; this is beyond the scope of
%this paper. 
not deal with the possibility of two experts using the same variable
%dalal23
%name to refer to different concepts.
name to refer to different phenomenon.

%dalal22 TO ADD we should add here a discussion on the complexity of f
%checking two  expert models are compatible as requested by reviewer 1
%and argue for feasibility "(Looks like a co-NP complete problem. If
%so, then the applicability of the proposed method is further restricted.)"
%joe32*: Still to do ...
%dalal23 Done by Hana

%dalal22 TO ADD We also need to say something here about what we gain
%from the merged models as reviewer 1 argues "There does not seem to
%be enough justifications of the proposed method, even assuming that
%the models are compatible.  What's the benefit of merging compared to
%letting the experts voting on \varphi?"
%dalal23 already said in the introduction

%joe12: there are some formatting issues here (stretched line), but
%I'm not sure why. 
% dalal9: latex does that when it can't hyphenate some if the words/symbols 
Returning to Example \ref{domestic1}, assume that either
MI2$_{M_1,M_2,DV}$, MI2$_{M_1,M_2,A}$,
%joe10
%MI3$_{M_1,M_2,DV}$  and MI3$_{M_1,M_2,A}$ all hold, or
MI3$_{M_1,M_2,DV}$,  and MI3$_{M_1,M_2,A}$ all hold, or
there is a default setting $\vec{v}^*$ such that 
%MI4$_{M_1,M_2,DV,\vec{v}^*}$  and MI4$_{M_1,M_2,A,\vec{v}^*}$ hold.
Then $M_1 \oplus M_2$ has the causal graph described in
Figure~\ref{sampsoncom}; that is, even though
%joe32
%Sampson et al.~\citeyear{Sampson:2013} do not have a formal theory for
\citet{Sampson:2013} do not have a formal theory for
%joe10
%combining models, they do so in just the way
%dalal23
%combining models, they actually combine models in just the way
%joe33
%merging models, they actually do so models in just the way
merging models, they actually merge models in just the way
that we are suggesting.

% the notion of conflict is not explicitly defined anywhere
Let $M_1 \sim_C^{\vec{v}^*} M_2$ be an abbreviation for
   $M_1 \succeq_C^{\vec{v}^*} M_2$ and
$M_2 \succeq_C^{\vec{v}^*} M_1$.
%joe7: added
We also write $M_1 \succ_C^{\vec{v}^*} M_2$ if
%joe11
%$M_1 \succeq_C^{\vec{v}^*} M_2$ if 
$M_1 \succeq_C^{\vec{v}^*} M_2$ and 
$M_2 \not\succeq_C^{\vec{v}^*} M_1$. 

%joe7
%The next proposition says that  Definition \ref{combat} is reasonable
%hana30
The next theorem
%The next proposition 
provides evidence that  Definition \ref{combat}
is reasonable 
%joe12
%and captures our intuitions.  Part (a) says that it is well defined
%(so that the  clauses in the definition where there might be
and captures our intuitions.  Part (b) says that it is well defined,
so that in the  clauses in the definition where there might be
    potential conflict, such as in the definition of $\F(C)$ when $C
%joe7: Aargh; missed this one
    %    \in \V_1 \cap \V_2$ and $(M_1,\vec{v}_1) \sim_C (M_2,\vec{u}_2)$,
%joe8
    %    \in \V_1 \cap \V_2$ and $M_1 \sim_C M_2$,
        \in \V_1 \cap \V_2$ and $M_1 \sim_C^{\vec{v}^*} M_2$,
        there is in fact no conflict;
        part (a) is a technical result
  needed to prove part (b).
  %dalal23
  %Part (c) states that the combined model 
%joe35
  %  Part (c) states that the merged model
    Part (c) says that the merged model 
%dalal23
%  in the compatible case is guaranteed to be acyclic. 
%joe35: there is no other case yet
    % fully compatible case is guaranteed to be acyclic.
     is guaranteed to be acyclic. 
%joe8
%    Part (d) says that at least as far
%    as formulas involving the variables in $M_1$ go, $M_1 \oplus M_2$
%    and $M_1$ agree, provided that the default values are used for the
%    exogenous variables not in $\U_1 \cap \U_2$.
%    (In the language of \citet{Hal44}, $M_1 \oplus M_2$ is essentially
%    a conservative extension of $M_1$.)  Finally, parts (e) and (f) say
%joe35: ^{\vec{v}^*}
     Part (d) says that causal paths in $M_1$ are preserved in $M_1
     \oplus^{\vec{v}^*} 
M_2$, while 
    part (e) says that at least as far
%joe35: again
    as formulas involving the variables in $M_1$ go, $M_1 \oplus^{\vec{v}^*} M_2$
    and $M_1$ agree, provided that the default values are used for the
    exogenous variables not in $\U_1 \cap \U_2$.
    Parts (d) and (e) can be viewed as saying that the essential
%joe32
    %    causal structure of $M_1$ is preserved in $M_1 \oplus M_2$.
        causal structure of $M_1$ and $M_2$ is preserved in $M_1
%joe35
        \oplus^{\vec{v}^*} M_2$.  All conclusions that can be drawn in $M_1$ and
        $M_2$ individually can be drawn in $M_1 \oplus^{\vec{v}^*} M_2$.
        %joe14
    \fullv{
    (In the language of \citet{Hal44}, part (e) says that $M_1 \oplus^{\vec{v}^*}
      M_2$ is essentially a conservative extension of $M_1$.)
      %joe14
      }
    %joe32: added
But it is important to note that $M_1 \oplus^{\vec{v}^*} M_2$ lets us go beyond
%dalal23 typo
%$M_1$ and $M_2$, since we can, for example, consider intereventions
$M_1$ and $M_2$, since we can, for example, consider interventions
%dalal23 typo
%that simulatenously affect in $M_1$ that are not in $M_2$ and
%joe33: there was actually a bigger typo here :-)
%that simultaneously affect in $M_1$ that are not in $M_2$ and
that simultaneously affect variables in $M_1$ that are not in $M_2$ and
variables in $M_2$ that are not in $M_1$.     
    Finally, parts (f) and (g) say
    that $\oplus^{\vec{v}^*}$ is commutative and associative over its domain.
%joe32: promoted this to a theorem
    %    \pro\label{oplusproperties}
        \thm\label{oplusproperties}
%joe34*: we need to add the \vec{v}^* superscript here (and talk about
%compatible with respect to vec{v}^*, if you agree with my earlier change.
        Suppose that $M_1$
%dalal23
%$M_2$, and $M_3$ are  pairwise compatible.  Then
$M_2$, and $M_3$ are  pairwise fully 
%hana30
compatible with respect to
 default setting $\vec{v}^*$.
%compatible. 
Then
%joe6
the following conditions hold.
\begin{itemize}
%joe6: simplified
%joe7: missed this too!
%\item[(a)] If $(M_1,\vec{v}_1) \sim_C^{\vec{v}^*} (M_2,\vec{u}_2)$ then
  \item[(a)] If $M_1 \sim_C^{\vec{v}^*} M_2$ then
    (i) $\Pa_{M_1}(C) = \Pa_{M_2}(C)$ and (ii) $\F_1(C) = \F_2(C)$;
      %        agree (when viewed as
%joe8: subsumed by the footnote I added above)
%    (when $\F_1(C) = \F_2(C)$ are viewed as
%  functions whose domains are just the values of the variables in
%$\Pa_{M_1}(C)$ (= $\Pa_{M_2}(C)$).
%joe7: didn't get commented out
%\item[(a)] If $(M_1,\vec{v}_1) \sim_C^{\vec{v}^*} (M_2,\vec{u}_2)$ then
%dalal: \oplus is used before between models not models and settings pairs
%joe7: unfortunately, there were a few other occurrences
  %\item[(b)] $(M_1,\vec{u}_1) \oplus (M_2,\vec{u}_2)$ is well defined.
%joe35: adding ^{\vec{v}^*} here and in a bunch of other places.
%Please check.
%  \item[(b)] $M_1  \oplus M_2$ is well defined.
    %\item[(c)]  $M_1 \oplus M_2$ is acyclic.
      \item[(b)] $M_1  \oplus^{\vec{v}^*} M_2$ is well defined.
\item[(c)]  $M_1 \oplus^{\vec{v}^*} M_2$ is acyclic. 
%joe8: added, then renumbered
\item[(d)] If $A$ and $B$ are variables in $M_1$, then $A$ is an
  ancestor of $B$ in $M_1$ iff $A$ is an ancestor of $B$ in $M_1
%joe35
  %  \oplus M_2$.
    \oplus^{\vec{v}^*} M_2$.
\item[(e)]  If $\varphi$ is a formula that mentions only the endogenous
%joe7*
%  variables in $M_1$,   $\vec{u}$ and $\vec{u}'$ agree on the
  %  variables in $\U_1 - \V_2$, and $\vec{u}'$ agrees with $\vec{v}^*$ on
  %  the variables in $\U_2 - (\U_1 \cup \V_1)$, then
%joe35
  %  variables in $M_1$, $\vec{u}$ is a context for $M_1 \oplus M_2$,
    variables in $M_1$, $\vec{u}$ is a context for $M_1 \oplus^{\vec{v}^*} M_2$,
    $\vec{u}_1$ is a context for $M_1$,
%joe39
%    $\vec{u}$ and $\vec{u}_1$ agree on the
%  variables in $\U_1  \cap \U_2$, $\vec{u}$ agrees with $\vec{v}^*$ on
%the variables in $\U - (\U_1 \cap \U_2)$, and $\vec{u}_1$ agrees with
    %$\vec{v}^*$ on the variables in $\U_1 - \U_2$,
and $\vec{u}$ and $\vec{u}_1$ are compatible with $\vec{v}^*$, 
    then
%joe7
%$(M_1, \vec{u}) \satt \varphi$ iff $(M_1 \oplus M_2, \vec{u}') \satt \varphi$.
$(M_1, \vec{u}_1) \satt \varphi$ iff $(M_1 \oplus M_2, \vec{u}) \satt \varphi$.
%joe35
%\item[(f)] $M_1 \oplus M_2 = M_2 \oplus M_1$.
\item[(f)] $M_1 \oplus^{\vec{v}^*} M_2 = M_2 \oplus^{\vec{v}^*} M_1$.
%joe7*
%\item[(f)] $M_3$ is compatible with $M_1 \oplus  M_2$, $M_1$ is compatible 
  %    with $M_2 \oplus M_3$, and
  %dalal23
  %  \item[(g)] If $M_3$ is compatible with $M_1 \oplus  M_2$ and $M_1$
%joe35
%\item[(g)] If $M_3$ is fully compatible with $M_1 \oplus^{\vec{v}^*}
  %M_2$ and $M_1$
  \item[(g)] If $M_3$ is fully compatible with $M_1 \oplus^{\vec{v}^*}
M_2$ with respect to $\vec{v}^*$ and $M_1$
  %dalal23
  %  is compatible     with $M_2 \oplus M_3$, then
%joe35
%is fully compatible     with $M_2 \oplus M_3$, then
%    $M_1 \oplus (M_2 \oplus M_3) =     (M_1 \oplus M_2) \oplus M_3$.
is fully compatible     with $M_2 \oplus^{\vec{v}^*} M_3$ with respect to $\vec{v}^*$, then
    $M_1 \oplus^{\vec{v}^*} (M_2 \oplus^{\vec{v}^*} M_3) =     (M_1
\oplus^{\vec{v}^*} M_2) \oplus^{\vec{v}^*} M_3$.   
\end{itemize}
%joe32
%\epro
\ethm

    %joe3
%    \prf Still to come (but it should be straightforward). \eprf
%joe7: added proof
%hana8: move the proof to appendix
%joe10
%joe32
%\fullv{The proof is rather involved, and appears in full in
\fullv{The proof of Theorem~\ref{oplusproperties} 
    is rather involved; the details can be found in~\ref{proof-oplusproperties}.} 
    
%joe6
    %    In light of Lemma~\ref{oplusproperties}, if $M_1, \ldots,
%joe7*
%dalal23 we talk about mutual compatibility here without enclosing
%this in a definition. Before this point we dont use the term mutual
% I didnt insert "fully" here but suspect we should
%joe33: I think ``mutual'' is reasonable.  There's an analogous notion
%of mutual independence
We define what it means for a collection $\M = \{M_1, \ldots, M_n\}$ of causal
%joe35: again, we need the $\vec{v}^*$.  I added it a bunch of times
%below without marking it.
%models to be \emph{mutually compatible} by induction on the
models to be \emph{mutually compatible with respect to default setting
 $\vec{v}^*$ (for all the exogenous variables that are not common to
  $M_1, \ldots, M_n$)} by induction on the
cardinality of $\M$. 
If $|\M| = 1$, then mutual compatibility trivially
holds.  If $|\M| = 2$, then the models in $\M$ are mutually compatible
with respect to $\vec{v}^*$ if
%dalal23
%they are compatible according to Definition~\ref{combat}.  If $|\M| = n$,
they are fully compatible with respect to $\vec{v}^*$ according to
Definition~\ref{combat}.  If $|\M| = n$, 
then the models in $\M$ are mutually compatible with respect to
$\vec{v}^*$ if the models in every
subset of $\M$ of cardinality $n-1$ are mutually compatible with
respect to $\vec{v}^*$, and for
%dalal23
%each model $M \in \M$, $M$ is compatible with $\oplus_{M' \ne M} M'$.
%joe35
%each model $M \in \M$, $M$ is fully compatible with $\oplus_{M' \ne M} M'$.
each model $M \in \M$, $M$ is fully compatible with
$\oplus^{\vec{v}^*}_{M' \ne M} M'$ with respect to $\vec{v}^*$.
%joe32
%By Proposition~\ref{oplusproperties}, if $M_1, \ldots,
By Theorem~\ref{oplusproperties}, if $M_1, \ldots,
%joe35
M_n$ are mutually compatible with respect to $\vec{v}^*$, then the causal model
%$M_1 \oplus \cdots \oplus M_n$ is well defined;
$M_1 \oplus^{\vec{v}^*} \cdots \oplus^{\vec{v}^*} M_n$ is well defined; 
%joe8
we do not have to worry about parenthesization, nor the order in which
%joe35
%the settings are combined.  Thus, the model $\oplus_{M' \ne M} M'$
the settings are combined.  Thus, the model $\oplus^{\vec{v}^*}_{M' \ne M} M'$
considered in the definition is also well defined.
%joe6
%joe8
%Proposition~\ref{oplusproperties}(d) also tells us that $M_1 \oplus
%joe32
%Proposition~\ref{oplusproperties}(e) also tells us that $M_1 \oplus
%joe35: adding ^{\vec{v}^*} again
Theorem~\ref{oplusproperties}(e) also tells us that $M_1 \oplus^{\vec{v}^*}
\cdots \oplus^{\vec{v}^*} M_n$
%joe7
%contains, in some sense, at least as much information as $M_1$.
contains, in  a precise sense, at least as much information as each
model $M_i$ individually.
%joe7: added
%dalal23 
%Thus, by combining mutually compatible models, we are maximizing our
Thus, by merging mutually compatible models, we are maximizing our
use of information. 

%joe22: moved from below
%dalal26
%This approach to aggregating models is our main contribution.
This approach to merging models is our main contribution.
Using it, we 
show in Section~\ref{sec:experts} how
experts' models can be combined to reason about interventions.

%joe14*: added discussion of Bradley et al.
%dalal23
%We now discuss the extent to which our approach to combining models
We now discuss the extent to which our approach to merging models
$M_1$ and $M_2$ satisfies BDL's desiderata.  
Recall that BDL considered
only causal networks, not 
causal models in our sense; they also assume that all models mention
the same set of variables.  They consider four desiderata.  We
briefly describe them and their status in our setting.
%joe35: 
Since BDL do not consider contexts explicity, we assume for simplicity
in this discussion that the context is the same for all models, and
talk only about $\oplus$ rather than $\oplus^{\vec{v}^*}$.
\begin{itemize}
%dalal23 again this is about to decide whether we say BDL talk about
%merging models or combing experts... since we talk about their
%in the context of merging models, i think we should refer to their
% as providing conditions for when to merge models and say that 
% for them it is the same as combining expert opinions
\item \emph{Universal Domain}: the rule for combining models accepts
  all possible inputs and can output any logically possible model.
  This clearly holds for us.
  %dalal23
  %\item \emph{Acyclicity}: the result of combining $M_1$ and $M_2$ is
\item \emph{Acyclicity}: the result of merging $M_1$ and $M_2$ is
%joe32
  %  acyclic.  This follows from Proposition~\ref{oplusproperties}(c),
    acyclic.  This follows from Theorem~\ref{oplusproperties}(c),
    provided that $M_1 \oplus M_2$ is defined.
\item \emph{Unbiasedness}: if $M_1 \oplus M_2$ is defined, and $M_1$
  and $M_2$ mention the same variables, then
  whether $B$ is a parent of $C$ in $M_1 
  \oplus M_2$ depends only on whether $B$ is a parent of $C$ in $M_1$
  %dalal22 Is something missing from the below. "and    This is"
%joe32
%  and in $M_2$, and    This is trivial for us, since if $B$ and $C$ are in
  and in $M_2$.   This property holds trivially for us, since if $B$
  and $C$ are in 
  both $M_1$ and $M_2$ and $M_1 \oplus M_2$ is defined, then $B$ is a
  parent of $C$ in $M_1 \oplus M_2$ iff $B$ is a parent of $C$ in both
  $M_1$ and $M_2$.
  (The version of this requirement given by BDL does not say  ``if $M_1 \oplus M_2$ is defined'',
  %dalal23 I did not change this as I assume they go straight into combining models
  %but for reasons set out earlier maybe we should explain for them these are
  %the same thing
%joe33: 
  %  since they assume that arbitrary models can be combined.)  BDL also
    since they assume that arbitrary models can be merged.)  BDL also
  have a \emph{neutrality} requirement as part of
  unbiasedness.  Unfortunately, an aggregation rule that says that $B$ is
  a parent of $C$ in $M_1 \oplus M_2$ iff $B$ is a parent of $C$ in
  both $M_1$ and $M_2$
%joe16: added
  (which seems quite reasonable to us)
  is not neutral in their sense, so we do not
  satisfy neutrality 
  %hana16 commented out
  %(nor do we find it natural in our setting).
  %joe43
    (nor, in light of the observation above, do we consider it a
  reasonble requirement to satisfy).

%joe35
  %\item \emph{Non-Dictatorship:} no single expert determines the
  \item \emph{Non-dictatorship:} no single expert determines the
  aggregation.  This clearly holds for us.
  \end{itemize}
  
%hana29 changed to a subsubsection and added the lemma about complexity.
%hana27
%joe34: I don't like having a single subsubsection
%\subsubsection{Complexity of full compatibility}\label{sec:complexity-full}
%\subsubsection{The complexity of full
%  compatibility}\label{sec:complexity-full} 
%\begin{remark}\label{remark:complexity-full}
%Checking whether two given causal models $M_1$ and $M_2$ are fully
We conclude this section with a characterization of the complexity of
determining whether two causal models are fully compatible.
Determining whether $M_1$ and $M_2$ are fully
%joe34
%compatible means checking whether the conditions of
compatible requires checking whether the conditions of
%joe33
%Def.~\ref{combat} hold. If the
Definition~\ref{combat} hold. 
%joe39
%This amounts to checking the conditions MI1$_{M_1,M_2,C,\vec{v}^*}$ and
This amounts to checking the conditions MI1$_{M_1,M_2,C}$  and
MI4$_{M_1,M_2,C,\vec{v}^*}$ for all variables 
$C \in (\U_1 \cup \V_1) \cap (\U_2 \cup \V_2)$.
%joe35*: slowing down.  Unfortunately, after I wrote this, I realized
%that this was a problem.  While assuming an explicit presentation
%does let us prove that the problem is in co-NP, and breaks our
%argument that the problem is co=NP hard, since there we just use the
%equation C=\phi, and don't give an explict presentation of C as a
%table.  If we did present C as a table, the problem would be in ptime.

%hana34 rewrote following our recent discussions
How hard this is to do depends in part on how the models are presented.
If the models are presented \emph{explicitly}, which means that, for
each variable $C$, the 
equation for $C$ is described as a (huge) table, giving the value of
$C$ for each possible setting of all the other variables, the problem
is \emph{polynomial} in the sizes of the input models.
%joe36: added
However, the size of the model will be exponential in the number of
variables.  

%joe36*: I think we should also allow a default values v^* here (so
%some rewriting needs to be done)
%Indeed, {\bf MI1}$_{M_1,M_2}$ amounts to checking whether
%hana36
In this case, checking whether
%In this case, checking if
%joe39
%MI1$_{M_1,M_2,C,\vec{v}^*}$ holds
MI1$_{M_1,M_2,C}$ holds 
%hana36
for all $C$
amounts to
checking whether
%joe36: we're just checking for a specific C; there is no
%MI1_{M_2,M_2} condition
%for all variables $C \in (\V_1 \cap \V_2) \cup (\V_1 \cap
% \U_2) \cup (\V_2 \cap \U_1)$,
%hana37 added v*
the parents of $C$ in $M_2$ are the immediate $M_2$-ancestors of $C$
%joe39
%in $M_1$, relative to the default setting $\vec{v}^*$.
in $M_1$.
%joe36*: this seems a little too complicated.  Moreover, we need to
%solve the dependence problem for all variables, not just for C
%For a given $C$, this can be reduced to the following problem that we
%denote by $Dep(F,\vec{X})$: {\it `given a function $F$ and a subset
%of its variables $\vec{X} = \{ X_1, \ldots, X_k \}$, does the value of
%$F$ depend on all variables in $\vec{X}$? As we argue below, it is
%actually useful to know the 
%exact subset of $\vec{X}$ that $F$ depends on, so the answer `no'
%should be accompanied by the subset of $\vec{X}$ which affect the
%value of $F$.
To solve this, we need to determine, for each pair of endogenous
variables $X$ and $Y$ in $M_i$ for $i=1,2$, whether $X$ depends on $Y$.
With this information, we can construct the causal graphs
for $M_1$ and $M_2$, and then quickly determine whether
%joe39
%MI1$_{M_1,M_2,C,\vec{v}^*}$ holds.
MI1$_{M_1,M_2,C}$ holds.  

%joe36: there is no \phi
%If $\varphi$ is given as an explicit table of values, then
%$Dep(F,\vec{X})$ amounts to finding, for each $X_i \in \vec{X}$, two
%rows in the table of values of $F$ that differ only in the value of $X_i$ and
If the model is given explicitly, then
determining wheter $X$ depends on $Y$ amounts to finding two
%joe37
%rows in the table of values of $F_XX$ that differ only in the value of
rows in the table of values of $F_X$ that differ only in the value of
$Y$ and 
in the outcome. As the number of pairs of rows is quadratic in the
size of the table, this is polynomial in the size of the input.
%joe39
%Thus, we can determine if MI1$_{M_1,M_2,C,\vec{v}^*}$ holds in
Thus, we can determine if MI1$_{M_1,M_2,C}$ holds in
polynomial time.   

%joe36
%The condition {\bf MI4}$_{M_1,M_2,C}$ for each variable $C \in (\V_1
%\cap \V_2) \cup (\V_1 \cap  \U_2) \cup (\V_2 \cap \U_1)$
Checking if MI4$_{M_1,M_2,C,\vec{v}^*}$ holds
 amounts to checking whether 
 \[ (M_1,\vec{u}_1) \satt [\vec{X} \gets \vec{x}](C=c), \]
iff 
\[ (M_2,\vec{u}_2) \satt [\vec{X} \gets \vec{x}](C=c). \]
%joe36: The real issue here is that there are lots of settings
%$\vec{X} \gets \vec{x}$.  
For a specific setting $\vec{u}$ and choice of $\vec{X}$ and
$\vec{x}$, we can easily compute the value of $C$ in a context
$\vec{u}$ if $\vec{X}$ is set to $\vec{x}$ (even if the model is not
given explicitly).  Since the number of possible settings is smaller
than the size of an explicitly presented model, we can also determine 
%hana36
whether MI4$_{M_1,M_2,C,\vec{v}^*}$ holds
%if MI4$_{M_1,M_2,C,\vec{v}^*}$
in polynomial time if the model is presented explicitly.
%joe36: not quite true, since \vec{X} might not be parents.
%which, if the models are presented explicitly, amounts to a
%comparison between the rows in the table of values for $C$ in $M_1$
%and $M_2$, which is again polynomial in the size of the table.

On the other hand, if the models are presented in a more compact way,
%joe36
%with structural equations being of polynomial size in the number of
%joe37
%with the structural equations being of polynomial size in the number of
with the structural equations, and hence the (descriptions of the)
models, being of size polynomial in the number of 
variables in the 
model, checking full compatibility is in a higher complexity class, as
%joe37
%we argue in 
%%hana35
%Proposition~\ref{lemma:complexity-combat} below.
%Propositions~\ref{lemma:complexity-MI4}
%and~\ref{lemma:complexity-MI1}  below. 
we now show.

%hana35
\begin{proposition}\label{lemma:complexity-combat}
%\begin{proposition}\label{lemma:complexity-MI4}
%joe36
% Given two causal models $M_1$ and $M_2$ of polynomial size in the
 % number of their variables, checking the condition {\bf
% number of variables, checking the condition {\bf    MI4}$_{M_1,M_2}$ 
 % of full compatibility according to Definition~\ref{combat} is
 %  number of variables, checking {\bf    MI4}$_{M_1,M_2}$
 Given two causal models $M_1$ and $M_2$ of  size polynomial in the
  number of variables, 
  %hana35
  determining whether they are fully compatible
  %joe37
  with respect to a default setting $\vec{v}^*$ 
  is in \ppar and is
  co-NP-hard 
  in the sizes of $M_1$ and $M_2$. 
  %checking {\bf    MI4}$_{M_1,M_2,C}$ 
% holds is 
 %co-NP-complete in the sizes of $M_1$ and $M_2$. 
 \end{proposition}
 %hana37 in the proof, added v* to MI1 and to the definition of the NP oracle
%dalal27
%\begin{proof}
\prf
%joe36
  %  Membership of {\bf MI4} in co-NP is proven by showing that the
  %hana35
  We prove a slightly stronger claim: that checking 
%joe37: we need the C, because we haven't defined it without the C. We
%also need the v^*.  Finally, we don't need boldface for MI4.
%{\bf MI4_{M_1,M_2}. Changed this globally
    %    MI4}$_{M_1,M_2}$ is co-NP-complete in
        MI4$_{M_1,M_2,C,\vec{v}^*}$ is co-NP-complete in 
 the sizes of $M_1$ and $M_2$, and that checking  
%joe39
 % MI1$_{M_1,M_2,C,\vec{v}^*}$ is in \ppar.
  MI1$_{M_1,M_2,C}$ is in \ppar. 
The complexity class
\ppar consists of all decision problems that can
be solved in polynomial time with parallel (i.e., non-adaptive)
queries to an NP oracle (see~\cite{Wag90,JT95,Joh90}).

%joe4
% We start with showing that checking that MI4$_{M_1,M_2}$ holds is
We start by showing that checking that MI4$_{M_1,M_2,C,\vec{v}^*}$ holds is
%dalal26
%joe40: what was the change here
% in co-NP by showing that the 
 in co-NP by showing that the 
complementary problem,
namely demonstrating that MI4$_{M_1,M_2,C,\vec{v}^*}$ does not hold, 
is in NP. 
 %hana35 I think we don't need a particular C - this holds in general
 %for MI4_M1,M2 
%We show  that checking that {\bf MI4}$_{M_1,M_2,C}$ holds is
%in co-NP by showing that the 
%complementary problem, 
%namely demonstrating that {\bf MI4}$_{M_1,M_2,C}$ does not hold 
%is in NP. 
To do this, we guess a witness:
%hana35 added guessing C to the witness
%joe37: removed again
%a variable $C \in (\V_1 \cap \V_2) \cup (\V_1 \cap \U_2) \cup (\V_2
%\cap \U_1)$,  
%joe36
%which is a variable $C \in (\V_1 \cap \V_2) \cup (\V_1 \cap 
% \U_2) \cup (\V_2 \cap \U_1)$, 
a setting $\vec{x}$ for the 
%joe37
%common variables $\vec{X}$ of $M_1$ and $M_2$ other than $C$ and a value
common variables $\vec{X}$ of $M_1$ and $M_2$ other than $C$, a value
%joe36
%of $C$, and contexts $\vec{u}_1$ and $\vec{u}_2$ for $M_1$ and $M_2$
%respectively such that  
$c$ of $C$, and contexts $\vec{u}_1$ and $\vec{u}_2$ for $M_1$ and $M_2$,
respectively, such that  
%hana35 added v*
%joe37
%for the default context $\vec{v}^*$,
%joe39
%$\vec{u}_1$ and $\vec{u}_2$ agree on the variables
%    in $\U_1 \cap \U_2$, $\vec{u}_1$ agrees with $\vec{v}^*$ on the
%    variables in $\U_1 - \U_2$, and $\vec{u}_2$ agrees with
%    $\vec{v}^*$ on the variables in $\U_2 - \U_1$,
$\vec{u}_1$ and $\vec{u}_2$ are compatible with $\vec{v}^*$, 
 \[ (M_1,\vec{u}_1) \satt [\vec{X} \gets \vec{x}](C=c), \]
but 
%joe36
%\[ (M_2,\vec{u}_2) \not\satt [\vec{X} \gets \vec{x}](C=c) \]
\[ (M_2,\vec{u}_2) \not\satt [\vec{X} \gets \vec{x}](C \ne c) \]
(or vice versa). A witness can be verified in polynomial time in the
size of the model, as it amounts to assigning 
values to all variables in the models and checking the value of $C$.

%hana35 added a discussion of C
%joe37: unnecessary
%We prove that that MI4$_{M_1,M_2}$ is co-NP-hard, which implies that
% MI4$_{M_1,M_2}^{v^*}$ is co-NP-hard as well (as {\bf MI4}$_{M_1,M_2}$ holds
% iff {\bf MI4}$_{M_1,M_2,C}$ holds for all $C \in (\V_1 \cap \V_2)
% \cup (\V_1 \cap \U_2) \cup (\V_2 \cap \U_1)$). 
%hana36
%joe38
The proof that the problem is co-NP hard is
%%joe36
%%We prove hardness of {\bf MI4} in co-NP by reduction from the known
%%hana35
%The proof is
%
%We prove that {\bf MI4}$_{M_1,M_2,C}$ is co-NP-hard 
by reduction from the known co-NP-complete 
%joe37: I think that we can assume that everyone knows what a
%tautology is
%problem \textit{Tautology} that determines, given a Boolean formula
%$\varphi$, whether $\varphi$ is a tautology, that is, evaluates to
%\bft under all assignments to its variables. 
problem \textit{Tautology}: determining whether a Boolean formula 
$\varphi$ is a tautology.
Let $\varphi$ be a Boolean formula over the variables $\{ Y_1, \ldots,
Y_n \}$. We construct a causal model $M_1$ as follows:
\begin{enumerate}
\item $\U_1 = \{U_1, \ldots, U_n\}$;
  \item $\V_1 = \{  Y_1, \ldots, Y_n , C\}$;
\item $\R_1(X) = \{ 0, 1 \}$ for all $X \in \V_1$;
\item the equations are $Y_i = U_i$ for $i = 1, \ldots, n$ and $C = \phi$.
\end{enumerate}
We note that since the set of exogenous variables is the same in $M_1$
and in $M_2$, we don't need to 
%joe37
%define the default context $\vec{v}^*$.
define the default setting $\vec{v}^*$.
In other words, the variables $\{ Y_1, \ldots, Y_n \}$ are binary
variables in $M_1$, and the value of $C$ is determined by $\varphi$. 

The second causal model $M_2$ is constructed as follows:
\begin{enumerate}
\item $\U_2 = \{U_1, \ldots,U_n\}$;
\item $\V_2 = \{  Y_1, \ldots, Y_n, C \}$;
\item $\R_2(X) = \{ 0, 1 \}$ for all $X \in \V_2$;
\item  the equations are $Y_i = U_i$ for $i = 1, \ldots, n$, and $C = 1$.
\end{enumerate}

%joe36
%{\bf MI4} holds for $M_1$ and $M_2$ are iff $\varphi$ is a tautology.
MI4$_{M_1,M_2,C,\vec{v}^*}$ holds iff $\varphi$ is a tautology.
%joe36
%Indeed, if $\phi$ is a tautology, then {\bf MI4} holds trivially. On
Indeed, if $\phi$ is a tautology, then MI4$_{M_1,M_2,C,\vec{v}^*}$ holds
trivially. On 
the other hand, if $\phi$ is not 
a tautology, then it is easy to see that MI4$_{M_1,M_2,C,\vec{v}^*}$ does
not hold, since there is some setting of the variables $Y_1, \ldots,
Y_n$ that makes $C = 0$.
%joe36*: I would cut this, since we haven't talked about general
%vs. binary models up to now
%We note that co-NP-completeness result holds for general (not only
%binary) causal models, as the proof of membership does not require the
%model to be binary. 

%hana35
%joe37*: again, we need the C.  We actually need the v^* too for MI1,
%but we ddin't include it, so I left it out here.  We may want to add
%it (but then we need to do so everywhere.
%To prove membership of {\bf MI1}$_{M_1, M_2}$ in \ppar, we describe a
%joe39
%To prove membership of MI1$_{M_1, M_2,C,\vec{v}^*}$ in \ppar, we describe a
To prove membership of MI1$_{M_1, M_2,C}$ in \ppar, we describe a
polynomial-time algorithm 
%joe39
%for deciding MI1$_{M_1, M_2,C,\vec{v}^*}$ that makes parallel queries
for deciding MI1$_{M_1, M_2,C}$ that makes parallel queries
to an NP oracle. 
%joe39
%We define an oracle $O^{Dep}(M,X,Y,\vec{v}^*)$ as follows: for a causal model
We define an oracle $O^{Dep}(M,X,Y)$ as follows: for a causal model
%joe37: note that here is where we use the v^*
%$M$ and two variables $X$ and $Y$ of $M$,  is gives an answer
%``yes'' if for the default context $\vec{v}^*$ $F_X$ depends on
$M$ and two variables $X$ and $Y$ of $M$,  it answers
%joe39: we're drawing the graph independent of the context ...
%``yes'' if, for the default context $\vec{v}^*$, $F_X$ depends on 
``yes'' if $F_X$ depends on 
the variable $Y$ in $M$ and ``no'' otherwise. It is easy to see that
%joe39
%$O^{Dep}(M,X,Y,\vec{v}^*)$  is in NP, since a witness for the
$O^{Dep}(M,X,Y)$  is in NP, since a witness for the
positive answer is 
a pair of 
assignments to the variables of $F_X$ that differ only in the value of
%joe37
%$Y$ and in the result. A witness is verifiable in polynomial time as
%it amounts to instantiating $F_X$ on these two assignments and
%verifying that the results are different.
$Y$ and in the result. A witness is clearly verifiable in polynomial
time; we simply instantiate $F_X$ on these two assignments and
verify that the results are different.
%joe36*: added; this is important, and probably should be said earlier
(We have implicitly assumed here that the equation $F_X$ can be
computed in polynomial time, as it is a part of $M$.)
%joe37*: this is not quite right.  We don't know the parents (that's
%what we're tring to determine) and the parents don't have to be
%exactly the same in any case.
%MI1$_{M_1, M_2,C}$ is decided by making parallel queries to
%$O^{Dep}(M,X,Y)$  with $M=M_1$ or $M=M_2$,  $X=C$ for
%each variable $C \in (\V_1 \cap \V_2) \cup (\V_1 \cap \U_2) \cup (\V_2
%\cap \U_1)$, and each parent $Y$ of $C$ in $M_1$ and in $M_2$.
%If the answers of the oracle for each $C$ are the
%same for $M_1$ and for $M_2$, we deduce that 
%{\bf MI1}$_{M_1, M_2}$ holds, otherwise it does not hold.
By querying the oracle
%joe39
%$O^{Dep}(M_i,X,Y,\vec{v}^*)$  for all endogenous variables $X$ and
$O^{Dep}(M_i,X,Y)$  for all endogenous variables $X$ and
$Y$ in $M_i$, 
for $i = 1,2$, we can determine the causal graphs of $M_1$ and $M_2$,
%joe39
%and thus whether MI1$_{M_1, M_2,C,\vec{v}^*}$ holds.
and thus whether MI1$_{M_1, M_2,C}$ holds.
The number of queries is at most quadratic in the sizes of $M_1$ and
 $M_2$, hence the algorithm terminates in polynomial time.
 %dalal27 for consistency in presentation 
% \end{proof}
\eprf
\subsection{Partial compatibility}\label{sec-part-compat}
%dalal6:
%\subsection{Decomposition and Partial Compatibility}

%joe10: larger than what? what can ony be determined?
%While the weaker definition of domination allows us to combine a
%larger class of models, this  it does not cover  cases in which it
%can only be determined for  a subset of  variables. 
% dalal8:
% Answer: larger than those that meeting string compatibility since these are a special case of the later
 While the notion of dominance used in Definition~\ref{combat} is
 useful, it still does not cover many cases of interest. 
 %hana9
%joe20
% \shortv{
%joe11
%   In the full paper, we discuss a number of real-world examples. Here
   % we briefly describe an example
%    that studies the emergence of radicalization in US prisons. 
We briefly describe an example here on causal models for the emergence 
 of radicalization in US prisons. The
 material is taken from 
%joe32
 % Useem and Clayton \citeyear{Useem:2009}.
  \citet{Useem:2009}.  
 %}
% Consider the example below about  emergence of radicalization  in two  US prison settings discussed by
% Useem and Clayton \shortcite{Useem:2009}. 
Although Useem and Clayton do not provide causal models,  
we construct these based on the description provided.
%dalal8:
We do not provide a detailed explanation of all the variables
%joe11
%and their dependencies owing to space limitations.  (Details  are
%provided in the full version. )For the purpose of illustrating a
%limitation of  Definition~\ref{combat}, it is sufficient to focus  on
and their dependencies here
%joe11: line shaving
%owing to space limitations
(details  are
provided in the full paper); for our purposes, it suffices to focus  on
the structure of these models.  

%&3

%e would like to say something about those variables for which such domination
%can be expressed. Consider the example below.

%joe10*: I think that we should significantly shorten this example for
%the abstract.  We can say something like ``we don't go into details
%about what the %variable mean (we do so in the full paper).  But we
%get the following %two models from experts''  That will save us more than half
%a column.
\begin{example} 
%dalal18
\label{prisonexample1}
  %dalal23
  \normalfont
%hana9
%The following two prison settings are taken from \citet{Useem:2009}: 
%dalal8:
%dalal18:
%Consider the two causal models in Figure \ref{fig:radical}. The
Consider the two causal models in Figure \ref{fig:radical}. 
%joe11
%\SCICH\ model represents  expert one's opinion over emerging
%dalal18
%\SCICH\ model represents  expert 1's opinion about emerging
$M_1$  represents  expert 1's opinion about emerging
radicalization ($R$) 
in 
%There are two prison settings: 
%(1) the State Correctional
%dalal8:
 the State Correctional
 %dalal18
%Institution Camp Hill  in Pennsylvania. The \TDCR\ model represents
Institution Camp Hill  in Pennsylvania. $M_2$ represents
%joe11
%expert two's opinion  in the causes of emergence in  Texas Department
%of Corrections and Rehabilitation (\TDCR). Both experts agree on three
%joe12
%expert 2's opinion about the causes of emergence in  Texas Department
expert 2's opinion about the causes of emergence in  the Texas Department
%dalal18:
%of Corrections and Rehabilitation (\TDCR). Both experts agree on 
of Corrections and Rehabilitation. Both experts agree on 
the structural equations for $R$. However they differ 
%joe11
%in the structural equations for \DO, \CB\ and \AM.
on the structural equations for \DO, \CB\ and \AM.
%dalal8:
%Consider two experts provide the  two  causal models    shown in
%Figure \citet{}, respectively.  
%dalal6:
%Although the two models assume quite different variables, they both have
%hana9
%dalal 8:
%\shortv{
%The outcome in both prisons, which we aim to analyze and change using interventions, 
%is the emerging radicalization ($R$). While the prisons share the main factors leading to the emergence of radicalization \textemdash ``order in prisons'' (\DO),
%``a boundary between the prison and potentially radicalizing
%communities'' (\CB), and ``having missionary leadership within the prison 
%organizations" (\AM) \textemdash they differ in the variables that lead to these factors; see the full paper for the full description of other variables in the example. 
%}
\fullv{
The authors point to 
three main factors upon which  the  emergence of radicalization
settings in both prison settings is dependent: ``order in prisons'' (\DO),
``a boundary between the prison and potentially radicalizing
communities'' (\CB), and ``having missionary leadership within the prison 
organizations" (\AM). The also both share
the same outcome \textemdash emerging radicalization ($R$). 
As can be observed from the descriptions provided,   some  variables
 and their dependency relations  are specific to    
 a prison.
%dalal18
%In the SCICH case,   \DO\ was attributed to 
%joe26: present tense makes more sense (at least to me) when talking
%about models.  We're making statements about the models
 % In $M_1$,   \DO\ was attributed to
  In $M_1$,   \DO\ is attributed to 
corruption (\GC) and lax management (\LM) in the prison's staff
together with prisons being allowed to roam freely (\FM). 
%joe26
%\CB\ was seen to be a result of  religious leaders within the
\CB\ is viewed as a result of  religious leaders within the
 facilities being permitted to  provide religious services freely
 (\IL)  and by prisoners showing membership within a prison community
%joe26
 % (\CM) which in turn was signalled by prisoners allowed to wear
%joe28
 % (\CM) which in turn is signalled by prisoners allowed to wear
  (\CM) which in turn is signalled by prisoners being allowed to wear
 distinguished street clothing (\SC). 
Prison authorities' exercising of internal punishments, such as
administrative  segregation (\AS), away from external oversight, and
%joe26
%\IL\ were considered to directly contribute to \AM.
\IL\ are considered to directly contribute to \AM.  
%dalal18
%TDCR instead considered \DO\ to be linked to the rapid growth in
%inmates numbers (\RG), inmates being allowed to assist prison   
%joe26
%$M_2$ instead considered \DO\ to be linked to the rapid growth in
$M_2$ instead considers \DO\ to be linked to the rapid growth in
%joe26: the next line doesn't make sense ``inmates being allowed to
%authorities''?  The letter T doesn't seem right either.
%dalal19
%inmates numbers (\RG), inmates being allowed to authorities in
%joe27
%inmates numbers (\RG), inmates being allowed to assist authorities in
inmate numbers (\RG), inmates being allowed to assist authorities in
%joe26
%maintaining order (\TT) and inmates feeling
%joe28: using the name T to represent ``inmates being allowed to
%assist authorities ...'' seems strange.  I redeined \TT as AA
maintaining order (\TT), and inmates feeling 
%joe11
%significantly  deprived (\DP) within the prisons--- the latter as a
significantly  deprived (\DP) within the prisons---the latter as a
result of being forced to engage in unpaid work (\UW) and having
limited contact with visitors (\MC). 
}
For the common variables \CB\ and \AM,  assuming default values for  $\GC$, $\LM$
%dalal18
%and $\FM$ in  \SCICH\ and  for $\TT$, $\DE$ and $\RG$ in \TDCR, 
%we can show that $\SCICH \succeq_{\sCB}^{\vec{v}^*} \TDCR$ and
%$\SCICH \succeq_{\sAM}^{\vec{v}^*} \TDCR$.
and $\FM$ in  \SCICH\ and  for $\TT$, $\DE$ and $\RG$ in $M_2$, 
we can show that $M_1 \succeq_{\sCB}^{\vec{v}^*} M_2$ and
$M_1 \succeq_{\sAM}^{\vec{v}^*} M_2$.
%joe11
%However no conclusion could be derived  about \PD\ from the two models,
%since neither MI1$_{\SCICH,\TDCR,\sPD}$ nor MI1$_{\TDCR,\SCICH,\sPD}$
However,  neither model dominates the other with respect to $\DO$;
neither MI1$_{M_1,M_2,\sPD}$ nor MI1$_{M_2,M_1,\sPD}$
%joe11
%holds. Therefore the models are incompatible by Definition
%dalal23 %holds. Therefore the models are incompatible according to Definition
holds. Therefore the models are not fully compatible according to Definition
\ref{combat}. 
%dalal23
\eprf
%hana11
\fullv{
\vspace{0.6cm}
}
\begin{figure}[h]
%joe18{\setlength{\unitlength}{.14in}}
%\shortv{\setlength{\unitlength}{.09in}}
  \shortv{\setlength{\unitlength}{.1in}}
%joe20
  \setlength{\unitlength}{.18in}
\begin{center}
\begin{picture}(6,15)(0,-6)
%dalal23
\thicklines
%\put(3,6){\circle*{.3}}
\put(3,4){\circle*{.4}}
\put(0,3.8){\circle*{.4}}
%\put(-3,9.5){\circle*{.3}}
\put(-3,8){\circle*{.4}}
%\put(0,8){\circle*{.3}}
\put(0,6){\circle*{.4}}
\put(-2,4){\circle*{.4}}
\put(-4,4){\circle*{.4}}
\put(5,4){\circle*{.4}}
\put(3,0){\circle*{.4}}
\put(0,-0.3){\circle*{.4}}
\put(-3,0){\circle*{.4}}
\put(0,-4){\circle*{.4}}
\put(-3,0){\vector(3,-4){3}}
\put(0,0){\vector(0,-4){4}}
\put(3,0){\vector(-3,-4){3}}
\put(-2,4){\vector(-1,-4){1}}
\put(-4,4){\vector(1,-4){1}}
\put(-3,8){\vector(2,-3){3}}
\put(-3,8){\vector(0,-4){8}}
\put(0,6){\vector(0,-4){3}}
\put(0,4){\vector(0,-4){4}}
\put(5,4){\vector(-1,-2){2}}
\put(3,4){\vector(0,-4){4}}
\put(3,4){\vector(-2,-3){3}}
\put(5,4){\circle*{.3}}
%\put(5.3,6){\scriptsize{$I_4$}}
\put(5.4,4){$\sAS$}
%\put(0.3,8){\scriptsize{$I_2$}}
\put(0.3,5.4){$\sSC$}
\put(0.3,3){$\sCM$}
\put(3.3,3.3){\sIL}
%\put(3.3,6){\scriptsize{$I_3$}}
%\put(-4.5,9.3){\scriptsize{$I_1$}}
\put(-4.8,7.3){$\sFM$}
\put(-5.5,3.3){$\sGC$}
\put(-2.1,3){$\sLM$}
\put(-4.5,-0.5){$\sPD$}
\put(0.3,-1){$\sCB$}
\put(3.3,-0.8){$\sAM$}
%joe3
%\put(-1.7,-5){\scriptsize{$\varphi = \sRS$}}
\put(-1.4,-5){\scriptsize{$R$}}
%joe21
%\put(-6.4,-5){\scriptsize{(1)~\SCICH}}
%dalal18
%\put(-6.4,-6){\scriptsize{\SCICH\ model}}
\put(-6.4,-6){$M_1$}
\end{picture}
%joe20
%\hspace{2.8cm}
\hspace{2cm}
\begin{picture}(6,12)(-6,-6)
%dalal23
\thicklines
\put(3,8){\circle*{.4}}
%\put(0,10){\circle*{.3}}
%\put(0,8){\circle*{.3}}
\put(0,3){\circle*{.4}}
\put(-5,8){\circle*{.4}}
\put(-3,8){\circle*{.4}}
%\put(-2,8){\circle*{.3}}
\put(-6,4){\circle*{.4}}
\put(-4,4){\circle*{.4}}
\put(-2,4){\circle*{.4}}
\put(3,0){\circle*{.4}}
\put(0,0){\circle*{.4}}
\put(-3,0){\circle*{.4}}
\put(0,-4){\circle*{.4}}
\put(-3,0){\vector(3,-4){3}}
\put(0,0){\vector(0,-4){4}}
\put(3,0){\vector(-3,-4){3}}
\put(-6,4){\vector(3,-4){3}}
\put(-4,4){\vector(1,-4){1}}
\put(-2,4){\vector(-1,-4){1}}
\put(-5,8){\vector(1,-4){1}}
\put(-3,8){\vector(-1,-4){1}}
%\put(0,6){\vector(0,-4){3}}
\put(0,3){\vector(0,-4){3}}
\put(3,8){\vector(0,-4){8}}
\put(3,8){\vector(-1,0){6}}
\qbezier(3,8)(0,12)(-5,8)
\put(-5.07,7.9){\vector(-1,-1){0}}
%joe3*(0.35,6){\sRC}(0.35,6){\sRS}
\put(0.3,3){$\sCM$}
%joe3: perhaps raise AS, so the edges leading out from it are going down?
\put(3.3,7.8){\sAS}
\put(-4,8){\sMC}
\put(-6,8){\sUW}
%\put(-1.7,7){\scriptsize{$I_5$}}
%joe28: raised slightly and shifted right
%\put(-1.7,3.2){$\sTT$}
%\put(-5,3.2){$\sDE$}
%\put(-7.5,3.2){$\sRG$}
\put(-1.7,3.4){$\sTT$}
\put(-4.8,3.4){$\sDE$}
\put(-7.3,3.4){$\sRG$}
\put(-4.5,-0.8){$\sPD$}
\put(0.2,-0.5){$\sCB$}
\put(3.3,-0.5){$\sAM$}
%joe3*
%\put(-1.8,-5){\scriptsize{$\varphi = \sRS$}}
\put(-1.4,-5){\scriptsize{$R$}}
%joe21
%\put(-6.4,-5){\scriptsize{(2)~\TDCR}}
%dalal18
%\put(-6.4,-6){\scriptsize{\TDCR\ model}}
\put(-6.4,-6){$M_2$}
\end{picture}
\caption{Schematic representation  of the two prison
  models.}\label{fig:radical} 
%dalal16
%\shortv{
%\vspace{-5pt}
%}
%joe20
\end{center}
\end{figure}
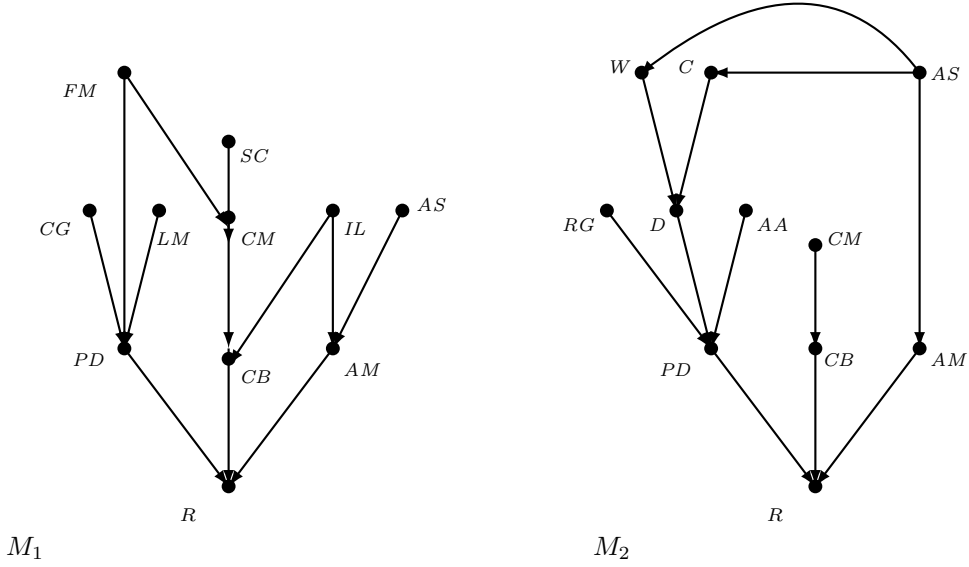
\end{example}

%dalal6:
%joe11: rewrote glue here
%In spite of the conclusion above,  there is still a sense in which
%\emph{parts of the models} are compatible.
%dalal23
%Although the models are not compatible according to our definition, 
Although the models are not fully compatible according to our definition, 
%joe24: added forward pointer
%the incompatibility is ``localized'' to the variable $\PD$.
the incompatibility is ``localized'' to the variable $\PD$, a point we
make precise in Section~\ref{sec:decomposition}. 
Moreover,
%joe12
%it is not even clear that there is disagreement with regard to $\DO$
it is not even clear that there is disagreement with regard to $\DO$;
the experts could just be focusing on different variables.
%joe22: added
(The issue of what the experts are focusing on and how this might
%joe24
%affect the issue of combining experts' models is dsicussed by
affect the issue of combining experts' models is discussed by 
%joe32
%Friedenberg and Halpern \citeyear{FH18}.)
\citet{FH18}.)
In a richer model, $\DO$ might have
%joe32
%six parents.  The trouble is, we have no idea from the two models what
six parents.  The trouble is, knowing the two models does not give
us any insight into what 
%joe12
%the equations for $\DO$ would be in the richer model, so we have no
%idea how to combine the models.
the equations for $\DO$ would be in the richer model.

%hana12 full paper should not refer to full paper
\shortv{
%joe11
In the full paper, we define an approach to 
%dalal23
%combining models even when they are not completely
merging models even when they are not fully
compatible.  We consider a notion of \emph{partial compatibility} 
  % dalal9:
  % and combine partially compatible causal models to
of causal models and construct a partial causal model.
%joe12: moved below
%(where, for
%example, we give $\DO$ six parents, but do not completely specify the
%equation for the value of $\DO$ as a function of its parents' values).  
% dalal9: missing words
%Roughly speaking, the causal model does not completely
%joe12
%Roughly speaking, the causal model does not define
Roughly speaking, the causal model does not completely define
the equations for variables $X$ that are common to two models, where
neither dominates the other with respect to $X$.
%joe12: moved here
In the example above, $\DO$ has six parents in the partial model, but
we do not completely specify the
equation for the value of $\DO$ as a function of its parents' values.
}

%joe14: made this a new paragraph and rewrote slightly (so as to shave
%a line)
%dalal15:
%This concludes the core contribution of this paper. Based on this, we
%illustrate in the next section how compatible
%experts' models can be combined to reason about interventions' efficacy.   
%dalal16
%joe22: moved earlier
%This approach to aggregating models is our main contribution.
%Using it, we 
%show in the next section how
%experts' models can be combined to reason about interventions.

%joe11
%This suggests the need for a yet more weaker notion of compatibility.
%dalal8
%\shortv{We discuss such a notion in the full paper.
%  We also show how
%  our definitions work on a number of real-world examples.}
%We now sketch how such a notion could be defined. 

%joe10*: At this point, we have worked out how to use
%decomposability.  It's not clear that we'll work it out by Monday.  I
%would suggest that we insert my notion of partial compatibility
%here, since at least that's worked out and formal.  Of course, it
%will only go in the full paper.

%dalal8: I have inserted what Joe wrote for partial compatibility from
%the radfull file 

%dalal22 put in a definition environement
%joe11: deferred to full paper
\fullv{
%joe32
  %  \begin{definition}[Weak Domination]
\begin{definition}[Weak domination] 
Let $\vec{v}^*$ be a default setting for the variables in $M_1$ and $M_2$.
We say that 
%dalal22 we used before the term relative
%\emph{$M_1$ weakly dominates $M_2$ with respect to a node $C$ and default
\emph{$M_1$ weakly dominates $M_2$ with respect to a node $C$ relative to 
%dalal8:
%setting $v^*$}, written $M_1 \ge_C^{v^*}$ if
%hana17 - have to invent a new notation
%dalal22 shouldnt the below be a vector?
  %setting $v^*$}, written $M_1 \succeq_{C}^{v^*}$ if
  %joe32: yes
  % $\vec{v}^*$}, written $M_1 \succeq_{w,C}^{v^*}$ if
   $\vec{v}^*$}, written $M_1 \succeq_{w,C}^{\vec{v}^*}$ if
MI4$_{M_1,M_2,C,\vec{v}^*}$ and the following weakening of
MI1$_{M_1,M_2,C}$ hold:
\begin{description}
\item[MI1$'_{M_1,M_2,C}$] If $A$ is a node in both $M_1$ and $M_2$ then $A$
is an immediate $M_1$-ancestor of $C$ in $M_2$ iff $A$ is a parent of
$C$ in $M_1$.
\end{description}
\end{definition}

Note that in Example~\ref{ex:famine}, neither $M_1$ nor $M_2$ weakly
dominates the other with respect to $F$: $P$ is a parent of $F$ in $M_2$ and is
not an immediate $M_1$ ancestor of $F$ in $M_1$, so $M_1$ does not
weakly dominate $M_2$ with respect to $F$, while $P$ is an immediate
$M_1$ ancestor of $F$ in 
$M_2$ and is not a parent of $F$ in $M_1$, so $M_2$ does not weakly
dominate $M_1$ either.
Also note that MI1 implies MI1$'$; MI1$'$ is a strictly weaker
condition than MI1, since it allows $M_1$ to weakly dominate $M_2$
with respect to $C$ if $C$ has parents in $M_1$ that are not in $M_2$
at all.  

%Dalal17: added definition environment
%dalal18
%\begin{definition}\label{weakcombat}[Weak Compatibility of Causal Models]
%joe32
%\begin{definition}\label{partialcombat}[Partial Compatibility of
%    Causal Models]
\begin{definition}\label{partialcombat}[Partial compatibility]
If $M_1 = ((\U_1,\V_1,\R_1),\F_1)$ and $M_2 = ((\U_2,\V_2,\R_2),\F_2)$, 
then 
%hana17 added w to the notation of weak compatibility everywhere.
%dalal18
%$M_1$ and $M_2$ are \emph{weakly compatible} iff for all nodes
%joe35
%$M_1$ and $M_2$ are \emph{partially compatible} iff
$M_1$ and $M_2$ are \emph{partially compatible with respect to default
setting $\vec{v}^*$} iff 
%dalal8: \succeq_
%in $C$ in both $M_1$ and $M_2$, either $M_1 \ge_C^{v^*} M_2$ or 
%joe32
%in $C$ in both $M_1$ and $M_2$, either $M_1  \succeq_{w,C}^{v^*} M_2$
%or
%joe33*: added, just like in full compatibility.  Is there a reason
%that we wouldn't want this.
%for all nodes in $C$ in both $M_1$ and $M_2$,
(1) for all variables $C \in (\U_1 \cup \V_1) \cap
(\U_2 \cup \V_2)$, we have $\R_1(C) = \R_2(C)$ and (2) for all variables
%joe20*: It is important here that we allow C to be exogenous in one
%model, otherwise we don't get acyclicity.  For example, if we have
%models M_1 and M_2 that both contain the variables A, B, C, D, but in
%M_1, A and C are exogenous, A is a parent of B, and C is a parent of
%D, while in M_2, B and D are exogenous, B is a parent of C, and D is
%a parent of A, then under the current definition, M_1 and M_2 are
%compatible, but M_1 \oplus M_2 is cyclic.  I have now corrected this.
%(My student Meir Friedenberg pointed out this problem.)
%$C \in \V_1 \cap \V_2$,
$C \in (\V_1 \cap \V_2) \cup (\V_1 \cap \U_2) \cup (\V_2 \cap \U_1)$,
either $M_1  \succeq_{w,C}^{\vec{v}^*} M_2$ or 
%dalal18
%$M_2 \succeq_{w,C}^{v^*} M_1$. If $M_1$ and $M_2$ are weakly compatible, then  $M_1 \oplus M_2 = ((\U,\V,\R),\F)$  
$M_2 \succeq_{w,C}^{\vec{v}^*} M_1$. If $M_1$ and $M_2$ are partially
%joe35
%compatible, then $M_1 \oplus M_2 = ((\U,\V,\R),\F)$
compatible with respect to $\vec{v}^*$, then $M_1 \oplus^{\vec{v}^*} M_2 = ((\U,\V,\R),\F)$
$M_1 \oplus M_2 = ((\U,\V,\R),\F)$  
is defined as follows:
\begin{itemize}
\item $\U$, $\V$, $\R$ are defined just as in Definition~\ref{combat}.
\item For $\F$, 
if $C \in V_1 - V_2$ then $\F(C) = \F_1(C)$.
and if $C \in V_2 - V_1$ then $\F(C) = \F_2(C)$.
If $C \in V_1 \cap V_2$ and
%dalal8:
%$M_1 \ge_C^{v^*}$, then let
%joe11
%$M_1 \succeq_C^{v^*}$, then let
%joe32
%$M_1 \succeq_C^{v^*} M_2$, then let
$M_1 \succeq_C^{\vec{v}^*} M_2$, then let
$\vec{P}_1$ consist of the parents of $C$ in $M_1$ and let 
$\vec{P}_2$ consist of the parents of $C$ in $M_2$ that are not in
$M_1$.  Then the parents of $C$ in $M_1 \oplus M_2$ are the nodes
$\vec{P}_1 \cup \vec{P}_2$.  Let $\vec{v}_2$ consist be the values of
the variables in $\vec{P}_2$ when the exogenous variables in $M_2$
have their default value in $\vec{v}^*$.
Given an arbitrary setting $\vec{x}$ of the variables in $\vec{P}_1$,
we define $\F(C)(\vec{x},\vec{v}_2) = \F_2(C)(\vec{x})$. 
Symmetrically, if $C \in V_2 - V_1$ or both $C \in V_1 \cap V_2$ and
%dalal8:
%$M_1 \ge_C^{v^*}$, then let
%joe11
%$M_1 \succeq_C^{v^*}$, then let
%joe32
%$M_1 \succeq_{w,C}^{v^*} M_2$, then let
$M_1 \succeq_{w,C}^{\vec{v}^*} M_2$, then let
$\vec{P}_1$ consist of the parents of $C$ in $M_1$ and let 
$\vec{P}_2$ consist of the parents of $C$ in $M_2$ that are not in
$M_1$.  Then the parents of $C$ in $M_1 \oplus M_2$ are the nodes
$\vec{P}_1 \cup \vec{P}_2$.  Let $\vec{v}_2$ consist be the values of
the variables in $\vec{P}_2$ when the exogenous variables in $M_2$
have their default value in $\vec{v}^*$.
Given an arbitrary setting $\vec{x}$ of the variables in $\vec{P}_1$,
we define $\F(C)(\vec{x},\vec{v}_2) = \F_2(C)(\vec{x})$. 
If $C \in V_1 \cap V_2$ and
%dalal8:
%$M_1 \ge_C^{v^*} M_2$, then let
%joe32
%$M_1 \succeq_{w,C}^{v^*} M_2$, then let
$M_1 \succeq_{w,C}^{\vec{v}^*} M_2$, then let
$\vec{P}_1$ consist of the parents of $C$ in $M_1$ and let 
$\vec{P}_2$ consist of the parents of $C$ in $M_2$ that are not in
$M_1$.  Then the parents of $C$ in $M_1 \oplus M_2$ are the nodes
$\vec{P}_1 \cup \vec{P}_2$.  Let $\vec{v}_2$ consist be the values of
the variables in $\vec{P}_2$ when the exogenous variables in $M_2$
have their default value in $\vec{v}^*$.
Given an arbitrary setting $\vec{x}$ of the variables in $\vec{P}_1$,
we define $\F(C)(\vec{x},\vec{v}_2) = \F_2(C)(\vec{x})$. 
We have a symmetric definition of $\F(C)$ if
%dalal8:
%$C \in V_1 \cap V_2$ and $M_2 \ge_C^{v^*} M_1$.
%joe32
%$C \in V_1 \cap V_2$ and $M_2 \succeq_{w,C}^{v^*} M_1$.
$C \in V_1 \cap V_2$ and $M_2 \succeq_{w,C}^{\vec{v}^*} M_1$.
Again, we write $M_1 \oplus M_2$ if $\U_1= \U_2$.
\end{itemize}
\end{definition}

This definition does not define $\F(C)$ for all possible values of the
parents of $C$. 
%dalal8: maybe we can include the following in the full version
 \fullv{For example,
%dalal8:
%For example, if $C  \in V_1 \cap V_2$ and $M_1 \ge_C^{v^*} M_2$, we
%joe32
   %if $C  \in V_1 \cap V_2$ and $M_1 \succeq_C^{v^*} M_2$, we
   if $C  \in V_1 \cap V_2$ and $M_1 \succeq_C^{\vec{v}^*} M_2$, we
have not defined $\F(C)(\vec{x},\vec{y})$ if $\vec{y}$ is a setting of
the variables in $\vec{P}_2$ other than $\vec{v}_2$.  Intuitively,
this is because the experts have not given us the information to
determine $\F(C)$ in these cases.  }
% dalal8
%Thus, we can think if $M_1 \oplus
We can think if $M_1 \oplus
M_2$ as a \emph{partial} causal model.  
Intuitively, we cannot define
$\satt$ in $M_1 \oplus M_2$ since we will not be able to define 
value of $(M_1 \oplus M_2, \vec{u}) \satt C=c$ for some setting $\vec{u}$.
Say that causal model $M^* = ((\U^*,\V^*,\R^*), \F^*)$ 
\emph{extends} $M_1 \oplus M_2$ if $(\U^*,\V^*,\R^*) = (\U,\V,\R)$ and
$\F^*(C) = \F(C)$ whenever $\F(C)$ is defined.  We now define a
3-valued version of $\satt$ in $M_1 \oplus M_2$ by taking 
$(M_1 \oplus M_2,\vec{u}) \satt \varphi$ iff $(M^*,\vec{u}) \satt \varphi$ for
all (complete) causal models $M^*$ extending $M_1 \oplus M_2$ and
taking 
%joe24: corrected typos
%$(M_1 \oplus M_2 \satt \varphi$ to be \emph{undefined} if neither 
%$(M_1 \oplus M_2 \satt \varphi$ nor
%$(M_1 \oplus M_2 \satt \neg \varphi$ holds.
$(M_1 \oplus M_2,\vec{u}) \satt \varphi$ to be \emph{undefined} if neither 
$(M_1 \oplus M_2,\vec{u}) \satt \varphi$ nor
$(M_1 \oplus M_2,\vec{u}) \satt \neg \varphi$ holds.

%dalal18
%Note that, according to the above definition, the two models in Figure~\ref{fig:radical} are weakly compatible.
%joe26
%Note that, according to the above definition, the two models in
Note that, according to the definition above, the two models in
Figure~\ref{fig:radical} 
are partially compatible.
%
%joe32
%We can now prove a generalization of Proposition~\ref{oplusproperties}.
We can now prove a generalization of Theorem~\ref{oplusproperties}.

%joe32
%\pro\label{oplusproperties1}
\thm\label{oplusproperties1}
%dalal8:
%Suppose that $M_1$
Suppose that $M_1$,
%dalal18
%$M_2$, and $M_3$ are  pairwise weakly compatible.  Then
%joe35
%$M_2$, and $M_3$ are  pairwise partially compatible.  Then
$M_2$, and $M_3$ are  pairwise partially compatible with respect to
$\vec{v}^*$.  Then 
the following conditions hold.
\begin{itemize}
%hana17 added w to the notation
  \item[(a)] If $M_1 \sim_{w,C}^{\vec{v}^*} M_2$ then
%    \item[(a)] If $M_1 \sim_C^{\vec{v}^*} M_2$ then
    (i) $\Pa_{M_1}(C) = \Pa_{M_2}(C)$ and (ii) $\F_1(C) = \F_2(C)$.
%joe35
%  \item[(b)] $M_1  \oplus M_2$ is well defined.
  %\item[(c)]  $M_1 \oplus M_2$ is acyclic.
  \item[(b)] $M_1  \oplus^{\vec{v}^*} M_2$ is well defined.
\item[(c)]  $M_1 \oplus^{\vec{v}^*} M_2$ is acyclic.
%hana17: added, then renumbered
\item[(d)] If $A$ and $B$ are variables in $M_1$, then $A$ is an
  ancestor of $B$ in $M_1$ iff $A$ is an ancestor of $B$ in $M_1
%joe35
  \oplus^{\vec{v}^*} M_2$.
\item[(e)]  If $\varphi$ is a formula that mentions only the endogenous
%joe7*
%  variables in $M_1$,   $\vec{u}$ and $\vec{u}'$ agree on the
  %  variables in $\U_1 - \V_2$, and $\vec{u}'$ agrees with $\vec{v}^*$ on
  %  the variables in $\U_2 - (\U_1 \cup \V_1)$, then
%joe35
  variables in $M_1$, $\vec{u}$ is a context for $M_1 \oplus^{\vec{v}^*} M_2$,
  $\vec{u}_1$ is a context for $M_1$,
%joe39
%  $\vec{u}$ and $\vec{u}_1$ agree on the
%  variables in $\U_1  \cap \U_2$, $\vec{u}$ agrees with $\vec{v}^*$ on
%the variables in $\U - (\U_1 \cap \U_2)$, and $\vec{u}_1$ agrees with
  %$\vec{v}^*$ on the variables in $\U_1 - \U_2$, then
  and $\vec{u}$ and $\vec{u}_1$ are compatible with $\vec{v}^*$, then
%joe35
$(M_1, \vec{u}_1) \satt \varphi$ iff $(M_1 \oplus^{\vec{v}^*} M_2, \vec{u}) \satt
\varphi$. 
\item[(f)] $M_1 \oplus^{\vec{v}^*} M_2 = M_2 \oplus^{\vec{v}^*} M_1$.
  %dalal18
%  \item[(g)] If $M_3$ is weakly compatible with $M_1 \oplus  M_2$ and $M_1$
   \item[(g)] If $M_3$ is partially compatible with $M_1 \oplus^{\vec{v}^*}  M_2$ and $M_1$
%dalal18
%   is weakly compatible     with $M_2 \oplus^{\vec{v}^*} M_3$, then
      is partially compatible     with $M_2 \oplus^{\vec{v}^*} M_3$, then
 $M_1 \oplus^{\vec{v}^*} (M_2 \oplus^{\vec{v}^*} M_3) =     (M_1
      \oplus^{\vec{v}^*} M_2) \oplus^{\vec{v}^*} M_3$.   
\end{itemize}
%joe32
%\epro
\ethm
} 

%hana17
%joe20
%The proof is similar to the proof of %Proposition~\ref{oplusproperties}.
The proof is almost identical to that of
Theorem~\ref{oplusproperties}, so we omit the details here.

%hana27
%joe34
%\begin{remark}
%Checking whether two given causal models $M_1$ and $M_2$ are partially
%compatible entails checking whether the conditions of
%%dalal24
%%Def.~\ref{partialcombat} hold.
%Definition~\ref{partialcombat} hold.
%
%hana34
It is easy to see that the problem of determining whether $M_1$ and $M_2$
are partially compatible is of the same complexity as the problem of determining
full compatibility
%It is easy to see that the problem of determining whether $M_1$ and $M_2$
%are partially compatible is co-NP complete 
%
%hana30 added - don't like the colon at the end of the sentence
by the following argument.
%hana29
%joe34
%It is easy to see that if $M_1$ and $M_2$ are fully compatible, they
Clearly, if $M_1$ and $M_2$ are fully compatible, they
are also 
%joe34
%partially compatible, and hence the problem of determining partial
partially compatible, so the problem of determining partial
compatibility is at least as hard 
as the problem of determining full compatibility. On the other hand,
by Definition~\ref{partialcombat}, 
the scope of compatibility is defined by the set of the common
variables, hence given 
two models $M_1$ and $M_2$, it is easy to determine the subset of
variables with respect to 
which we need to check compatibility. 
%hana34
Therefore, checking partial compatibility is of the same complexity as checking full compatibility.
%Therefore, checking partial compatibility is in co-NP,
%by the same argument as that in Proposition~\ref{lemma:complexity-full}.  

%joe24: added label
%\subsection{Decomposition of causal models}
\subsection{Decomposition of causal models}\label{sec:decomposition}

%dalal22 added introduction to justify this notion
%joe32: unommented the next six lines, but made some minor changes
Even with the generalised notions of compatibility introduced above, it may
%joe32
%be the case that although two expert models are considered still
%incompatible in the general sense,
%yet each comprises submodels which, if isolated from the rest, could
%be useful  in determining common interventions.
still be the case that two expert's models are
%dalal23 not clear to me whether we want to say are not "partially"compatible
%incompatible.  But we would expect that these models have submodels 
not compatible.  But we would expect that these models have submodels 
%dalal23 I would add below "partially/fully" depending on the intended meaning
%joe33: I would say that we're better off not using the terms
%``fully'' or ``partially'' here.  Rrather, we should be deliberately
%ambiguous, allowing ``compatible'' to stand for both ``fully
%compatible'' and ``partially compatible''
that are compatible.
 %We are now ready to introduce the notion of decomposition of causal
%joe32: 
%This brings us to the notion of  decomposition of causal  models.
% There are several advantages to being able to decompose a
%given model 
%%joe20
%%to a set of smaller submodels. First, consider the situation where the
%into a set of smaller submodels.
Finding such submodels has several advantages.
First, consider the situation where the
%joe20
%policymaker is given several different causal models, which are not
policymaker is given several different causal models that are not
%joe20
%compatible according to Definition~\ref{combat}. However, the models
%have similar parts that can be combined and indicate an effective intervention.
%dalal23
%compatible according to Definition~\ref{combat}. If we could decompose
fully compatible according to Definition~\ref{combat}. If we could decompose
the models, we might be able to ``localize'' the incompatibility, and
%joe32
%combine parts of the models so as to indicate an effective intervention.
%dalal23
%combine the parts of the models that are compatible.  Doing so may suggest
merge the parts of the models that are fully compatible.  Doing so may suggest
effective interventions. 
Another advantage of decomposing a model is that it allows the policymaker to reason about
each submodel in isolation. Since the problem of computing causes is DP-complete and the problem of computing interventions is co-NP-complete,
%joe20
%having a smaller model to reason about would potentially have a very
%significant effect on the complexity of the problem. 
having a smaller model to reason about could have a 
significant impact on the complexity of the problem. 

In order to define the notion of decomposition,
%joe21
%we first introduce the
%notion of a \emph{partial order} between endogenous variables and then 
%between submodels.
we need some preliminary definitions.
%dalal21
%\dfn\label{def:porder}
\dfn\label{def:porder}\fullv{[Order-preserving partition]}
A sequence $\< \V_1, \ldots, \V_k \ra$ of subsets of variables in $\V$
variables in a causal model $M$ is an
%joe21
%\emph{order-preserving} if, for all $i, j$ with $i < j$,
%dalal22
%joe32: what change was made here?
%dalal23 changed capital _J to small _j
%\emph{order-preserving partition} if $V_i \cap V_J = \emptyset$ for $i
 \emph{order-preserving partition} if $V_i \cap V_j = \emptyset$ for $i
\ne j$, $\cup_{i=1}^k \V_i = \V$ (so $\{V_1,\ldots, \V_k\}$ is a
partition of $\V$), and
for all $i, j$ with $i < j$, 
%joe21*: here's my simpler definition.  What am I missing
%we have $V_i \ll V_j$ or $\vec{V}_i \bowtie \vec{V}_j$.
%dalal22 reviewer 3
%no variable in $\V_j$ is the ancestor of a variable in $\V_i$.
no variable in $\V_j$ is an ancestor of a variable in $\V_i$.
\edfn

%\dfn\label{def:decomposable}[Decomposable causal models]
\dfn\label{def:decomposable}\fullv{[Decomposable causal models]}
%joe20
%If $M = ((\U,\V,\R),\F)$, then $M$ is \emph{decomposable} if there
$M = ((\U,\V,\R),\F)$ is \emph{decomposable} if there
%joe20*: we haven't defined what it means to partition a causal model
%into submodels.  But I actually don't think we want a partition.  I
%think we want to allow overlap between exogenous variables in one
%model and endogenous variables in another model.  Specifically, I
%think we need to allow a variable X to appear in two submodels, if X
%is exogenous in one, endogenous in the other, and the model where X
%is endogenous includes all of X's parents in M.  We should not allow
%X to appear in more than two models.  I do think that we want U_1,
%\ldots, \U_k to be a partition of \U.  See further
%comments below on the baby P example
%exists a partition of $M$ into $k>1$ pairwise compatible causal models
%$\{M_{i} = ((\U_{i},\V_{i},\R_i),\F_{i})\}$, where $1\le i \le k$, such that
%the set $\{ \U_{i} \cup \V_{i} : 1\le i \le k \}$ is
%a partial-order-preserving partition of the set of variables $\U \cup \V$.
%hana21 replaced > with \geq
%dalal23 in what sense fully/partially
%joe33
%if there exist $k \geq 1$ compatible causal models
exist $k \geq 1$ fully compatible causal models 
%if there exist $k > 1$ compatible causal models 
%
$\{M_{i} = ((\U_{i},\V_{i},\R_i),\F_{i}): 1 \le i \le k\}$,
%hana18
%joe21
%such that $\{  \V_{i} : 1\le i \le k \}$ is an order-preserving
such that $\<  \V_1, \ldots, \V_k\ra$ is an order-preserving
partition of $\V$, $\F_{i} \subseteq \F$ is the set of structural equations
that assign the values to the variables in $\V_{i}$,
and for each model $M_i$, the set $\U_{i}$ consists of
%joe21*: I don't think that this is correct.  \U_i might also
%include endogenous variables in M.  This happens in our examples
all the endogenous and exogenous variables in $M$ not in $\V_i$ 
that participate in the structural equations for the variables of
%joe21
%$M_i$.
$\V_i$.
%such that $\{ \U_{i} \cup \V_{i} : 1\le i \le k \}$ is an order-preserving
%partition of $\U \cup \V$, $\U_i \subseteq \U$,
%$\{\U_1, \ldots, \U_k\}$ is a partition of $\U$,  and $\F_i$ agrees with $\F$
%on the variables in $\U_i$, 
%
%joe20*: we shouldn't need to say this.  By definition of causal model,
%the variables that are on the left-hand side of the equations in \F_i
%are precisely those in \U_i, and the remaining variables are in \V_i.
%The partition of variables to exogenous and endogenous is as follows:
%for each submodel $M_i$, the variables that are assigned in $M_i$ 
%are endogenous, and the variables that are not assigned in $M_i$ but
%participate in the structural equations for the variables of $M_i$
%are exogenous.   
%joe20: added
$M_1, \ldots, M_k$ is called a \emph{decomposition} of $M$.
\edfn
%joe10: I don't understand the note.  We should this definition an
%iff.  It is true by definition.
%\note{I suspect the above holds for iff.}

%joe20*: We need some discussion here of why we require $V_i
%\ll V_j$ or $\vec{V}_i \bowtie \vec{V}_j$.  This will appear quite
%mysterious to the reader.  I think we should also give a simple
%example of decomposition here, where one model is decomposed into
%two.  I think that the models is Figure 4, for example, are
%decomposable (at least, according to how I think of decomposition)

\lem\label{unique}
%joe20
%Given a decomposable causal model $M$, for any partition of $M$ into
%$k>1$ pairwise compatible causal models 
%$\{M_{i} = ((\U_{i},\V_{i},\R_i),\F_{i})\}$ as in
%Definition~\ref{def:decomposable}, the combined model  
%$M_{1} \oplus ... \oplus M_{k}$ is unique and is equal to $M$.
If $M_1, \ldots, M_k$ is a decomposition of $M$, then
$M_{1} \oplus \cdots \oplus M_{k} = M$.
\elem

%joe20
%\pf 
%The proof is straightforward by observing that we do not change any %of
\prf The proof is immediate given the observation that we do not
change any of 
%joe21
%the structural equations of $M$ when decomposing it to submodels.
the structural equations of $M$ when decomposing it into submodels.
%joe20
%dalal23
%joe33
\eprf
%$\Box$

%hana20
%joe26
%It is easy to see that for a given model, there can be many ways to
It is easy to see that, for a given model, there can be many ways to
%joe23
%decompose it to a set of sub-models according to
%joe26
%decompose it to a set of submodels according to
decompose it into a set of submodels according to
%dalal18
%Def.~\ref{def:decomposable}. Moreover, 
Definition \ref{def:decomposable}. Moreover, 
%dalal18
%all models are decomposable by Def.~\ref{def:decomposable}. Indeed,
all models are decomposable by Definition \ref{def:decomposable}. Indeed,
any model $M$ can be trivially decomposed to 
%joe23
%$|\V|$ sub-models, each of which consists of exactly one endogenous
$|\V|$ submodels, each of which consists of exactly one endogenous
%joe26
%variable. Of course, such a decomposition is useless for any practical 
%purposes, and the decompositions we consider are those that help in
variable. Of course, such a decomposition is useless for practical 
purposes; the decompositions we consider are those that help in
either analysing the model or reducing the complexity of computing 
causes.  In Example~\ref{ex-decomposable} below, we demonstrate a
%joe26
%non-trivial decomposition.
nontrivial decomposition. 

%hana18
%moved the baby P decomposable example and the discussion that was here to the section on baby P
%hana19
\begin{example}\label{ex-decomposable}
  %dalal23
  \normalfont
%dalal17
%In the prison example in Figure~\ref{fig:radical}, we can decompose $M_1$ according to Definition~
Consider the causal models in Figure~\ref{fig:radical} from the prison
%joe26
%example \ref{prisonexample1}. We can
example (Example~\ref{prisonexample1}). We can 
%joe21: there is no model M_1 in the figure
%decompose $M_1$ according to Definition~%
%dalal17: added the correct label
%joe22: 
%decompose the  the model in \SCICH\ model (we label here as $M_1$)
%h:
%decompose the \SCICH\ model (which we call $M_1$ here)
decompose  $M_1$
according to Definition  
%dalal17:
%\ref{def:decomposable}. For example, $FM, CG, LM,$ and $PD$ can be endogenous variables of one submodel 
%joe21*: I think we should give a complete decomposition, not just mention two
%submodels; we should also explicitly mention the exogenous variables,
%which we currently don't.   I rewrote and expanded the example.
%\ref{def:decomposable}.
%A possible decomposition of this model could
%be one in which the variables $\FM, \GC, \LM,$ and $\PD$ are
%endogenous variables of one submodel $M^1_1$, and $\SC$ and $\CM$
%endogenous variable of another submodel $M^2_1$. 
%We note that, for example, $\CB$ cannot be an endogenous variable of
\ref{def:decomposable} into $M_{11}$, $M_{12}$, $M_{13}$, and
%dalal20
%$M_{14}$, where $M_{ij} = ((\U_{ij}, \V_{ij}, \R_{ij}), \F_{ij})$,
$M_{14}$, where $M_{ij} = ((\U_{ij}, \V_{ij},$ $\R_{ij}),$ $\F_{ij})$,
$\V_{11} = \{\FM, \GC, \LM,\PD\}$, $\U_{11}$ consists of all
%dalal17:
%the exogenoous variables in $M$ in the \SCICH\ model (which are not
%joe22 
%the exogenoous variables in $M_1$ in the \SCICH\ model (which are not
%dala18
%the exogenoous variables in $M_1$ (which are not
the exogenous variables in $M_1$ (which are not
explicitly given in Figure~\ref{fig:radical}) that are ancestors of
the variables in $\V_{11}$, $\V_{12} = \{\SC, \CM, \CB, \IL\}$,
$\U_{12}$ consists of all
%dalal18
%the exogenoous variables in $M$ in the \SCICH\ model 
the exogenous variables in $M_1$
that are ancestors of the variables in $\V_{12}$ together $\FM$,
$\V_{13} = \{\AS,\AM\}$, $\U_{13}$ consists of all
%dalal18:
%the exogenoous variables in $M$ in the \SCICH\ model 
the exogenous variables in $M_1$
that are ancestors of the variables in $\V_{13}$ together $\IL$,
$\V_{14} = \{R\}$, and $\U_{14} = \{\PD,\CB,\AM\}$.  
%dalal18
Similarly we can decompose $M_2$ into four submodels $M_{21}$,
%joe30
%$M_{22}$, $M_{23}$, and
$M_{22}$, $M_{23}$, where
$M_{24}$, where $M_{ij} = ((\U_{ij}, \V_{ij}, \R_{ij}), \F_{ij})$,
%dalal21
%$\V_{21} = \{\MC,\sUW, \sTT, \sDE, \sRG, \sPD\}$, $\U_{21}$ consists of all
$\V_{21} = \{\MC,\UW, \TT, \DE, \RG, \PD\}$, $\U_{21}$ consists of all
the exogenous variables in $M_2$
%dalal21
%that are ancestors of the variables in $\V_{21}$. $\V_{22} = \{\sCM, \sCB\}$,
 that are ancestors of the variables in $\V_{21}$. $\V_{22} = \{\CM, \CB\}$,
$\U_{22}$ consists of all the exogenous variables in $M_2$
that are ancestors of the variables in $\V_{22}$, 
$\V_{23} = \{\AS,\AM\}$, $\U_{23}$ consists of all the exogenous variables in $M_2$
that are ancestors of the variables in $\V_{23}$, 
$\V_{24} = \{R\}$, and $\U_{24} = \{\PD,\CB,\AM\}$. 
%dalal17
%Figure \ref{fig:radical_decom}
%dalal18
Figures \ref{fig:radical_decom1} and Figures \ref{fig:radical_decom2}
%joe22
show
%dalal18
%the four submodels resulting from this decomposition.
%joe26: ``respsectively'' doesn't seem to me needed here.
%the four submodels resulting from these decompositions respectively.
the four submodels resulting from these decompositions.
%joe21*: added
There is some flexibility in how we do the decomposition.  For
example, we could move $\IL$ from $\V_{12}$ and $\U_{13}$ to $\V_{13}$
and $\U_{12}$.  
However, we cannot, for example,  move  $\CB$ into $\V_{i1}$, for
then $\<\V_{i1}, \V_{i2},\V_{i3},\V_{i4}\ra$ would not be an
order-preserving partition 
%joe21
%$M_1$ because of the partial order between variables ( since $\FM \ll
%\CM \ll  \CB$).
(since $\FM$ is an ancestor of $\CM$, which is an ancestor of $\CB$).
%dalal23
\eprf
%joe21: no need to say this here; we say it below
%Such a 
%decomposition can be beneficial if we can analyse each submodel
%separately and then use these conclusions to derive conclusions about
%the whole model $M_1$. 
%dalal17
%In Section~\ref{sec-babyp} we discuss decomposition in the case of
%Baby P., where different sets of variables correspond to different
%aspects of the case. 
\end{example}
%dalal17:
  \setlength{\unitlength}{.14in}
\begin{figure}[h]
%\begin{center}
\begin{picture}(6,15)(-6,-6)
%dalal23
\thicklines
\put(-3,8){\circle*{.4}}
\put(-2,4){\circle*{.4}}
\put(-4,4){\circle*{.4}}
\put(-3,0){\circle*{.4}}
%\put(0,-4){\circle*{.3}}
%\put(-3,0){\vector(3,-4){3}}
\put(-2,4){\vector(-1,-4){1}}
\put(-4,4){\vector(1,-4){1}}
\put(-3,8){\vector(0,-4){8}}
\put(-4.8,7.3){$\sFM$}
\put(-5.5,3.3){$\sGC$}
\put(-2.1,3){$\sLM$}
\put(-4.5,-0.5){$\sPD$}
%\put(-1.4,-5){\scriptsize{$R$}}
\put(-2.4,-2){\scriptsize{$M_{11}$}}
\end{picture}
\hspace{0.5cm}
\begin{picture}(6,15)(-4,-6)
%dalal23
\thicklines
\put(3,4){\circle*{.4}} % IL 
\put(0,3.8){\circle*{.4}} %CM
\put(-3,8){\circle*{.4}} %FM
\put(0,6){\circle*{.4}} %SC
\put(0,-0.3){\circle*{.4}} % CB
%\put(0,-4){\circle*{.3}} % R
%\put(0,-0.3){\vector(0,-4){3.55}} %CB to R
\put(0,6){\vector(0,-4){2}} %SC to CM
\put(0,4){\vector(0,-4){4.2}} %CM to CB
\put(3,4){\vector(-2,-3){2.8}} % IL to CB
\put(-3,8){\vector(2,-3){2.8}} %FM to CM
\put(0.3,5.4){$\sSC$}
\put(0.3,3){$\sCM$}
\put(3.3,3.3){\sIL}
\put(-4.8,7.3){$\sFM$}
\put(0.3,-1){$\sCB$}
%\put(-1.4,-5){\scriptsize{$R$}}
\put(-2.4,-2){\scriptsize{$M_{12}$}}
\end{picture}
\hspace{0.5cm}
\begin{picture}(6,15)(-4,-6)
%dalal23
\thicklines
\put(5,4){\circle*{.4}} % AS
\put(3,0){\circle*{.4}} %AM
\put(3,4){\circle*{.4}} % IL 
%\put(0,-4){\circle*{.3}} %  R
%\put(3,0){\vector(-3,-4){2.9}} % AM to R
\put(5,4){\vector(-1,-2){1.95}} %AS to AM
\put(3,4){\vector(0,-4){3.9}} %IL to AM
\put(5.4,4){$\sAS$}
\put(3.3,3.3){\sIL}
\put(3.3,-0.8){$\sAM$}
%\put(-1.4,-5){\scriptsize{$R$}}
\put(1,-2){\scriptsize{$M_{13}$}}
\end{picture}
\hspace{3cm}
\begin{picture}(6,15)(-4,-6)
%dalal23
\thicklines
\put(-3,4){\circle*{.4}} % PD
\put(0,4){\circle*{.4}} %CB
\put(3,4){\circle*{.4}} %AM
%\put(0,3.8){\circle*{.3}} %CB
\put(0,0){\circle*{.4}} %R
\put(0,4){\vector(0,-4){4}} %CB to R
\put(3,4){\vector(-3,-4){2.9}} % AM to R
\put(-3,4){\vector(3,-4){2.9}} %PD to R
\put(3.3,3.8){$\sAM$}
\put(-1.4,-1){\scriptsize{$R$}}
\put(-2.4,-2){\scriptsize{$M_{14}$}}
\put(-4.5,3.5){$\sPD$}
\put(0.3,4){$\sCB$}
\end{picture}
%\end{center}
\caption{Decomposition of  the model $M_1$ from Example \ref{prisonexample1}.}\label{fig:radical_decom1} 
\end{figure}
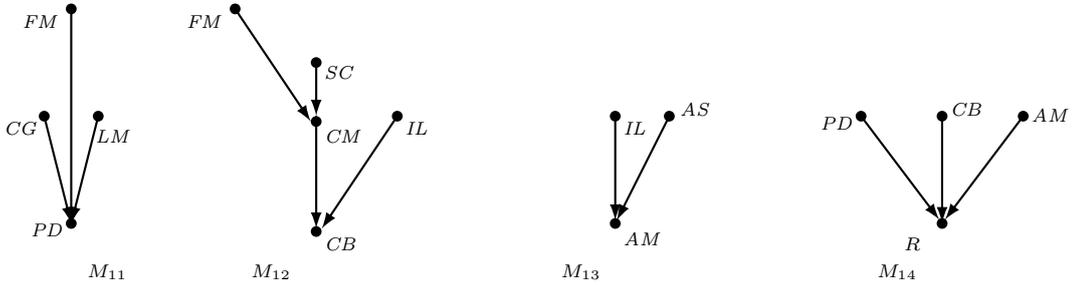
%hana19

%Dalal17
\begin{figure}[h]
\begin{center}
\begin{picture}(5,14)(-5,0)
%dalal23
\thicklines
\put(3,13){\circle*{.4}} %AS
%\put(0,8){\circle*{.3}} %CM
\put(-5,13){\circle*{.4}} %W
\put(-3,13){\circle*{.4}} %C
\put(-6,9){\circle*{.4}} %RG
\put(-4,9){\circle*{.4}} %D
\put(-2,9){\circle*{.4}} %T
%\put(3,5){\circle*{.3}} %AM
%\put(0,5){\circle*{.3}} %CB
\put(-3,5){\circle*{.4}} %PD
%\put(0,1){\circle*{.3}} %R
%\put(-3,5){\vector(3,-4){3}}
%\put(0,5){\vector(0,-4){4}}
%\put(3,5){\vector(-3,-4){3}} %AM to R
\put(-6,9){\vector(3,-4){3}}
\put(-4,9){\vector(1,-4){1}}
\put(-2,9){\vector(-1,-4){1}}
\put(-5,13){\vector(1,-4){1}}
\put(-3,13){\vector(-1,-4){1}}
%\put(0,8){\vector(0,-4){3}}
%\put(3,13){\vector(0,-4){8}}
\put(3,13){\vector(-1,0){6}}
\qbezier(3,13)(0,17)(-5,13)
\put(-5.07,12.9){\vector(-1,-1){0}}
%\put(0.3,8){$\sCM$}
\put(3.3,12.8){\sAS}
\put(-4,13){\sMC}
\put(-6.5,13){\sUW}
\put(-1.7,8.2){$\sTT$}
\put(-5,8.2){$\sDE$}
\put(-7.5,8.2){$\sRG$}
\put(-5,4.8){$\sPD$}
%\put(0.2,4.5){$\sCB$}
%\put(3.3,4.5){$\sAM$}
%\put(-1.4,0){\scriptsize{$R$}}
\put(-2.4,3){\scriptsize{$M_{21}$}}
\end{picture}
\hspace{1.8cm}
\begin{picture}(5,14)%(-6,0)
%dalal23
\thicklines
%\put(3,13){\circle*{.3}} %AS
\put(0,8){\circle*{.4}} %CM
%\put(-5,13){\circle*{.3}} %W
%\put(-3,13){\circle*{.3}} %C
%\put(-6,9){\circle*{.3}} %RG
%\put(-4,9){\circle*{.3}} %D
%\put(-2,9){\circle*{.3}} %T
%\put(3,5){\circle*{.3}} %AM
\put(0,5){\circle*{.4}} %CB
%\put(-3,5){\circle*{.3}} %PD
%\put(0,1){\circle*{.3}} %R
%\put(-3,5){\vector(3,-4){3}} %PD to R
%\put(0,5){\vector(0,-4){4}} %CB to R
%\put(3,5){\vector(-3,-4){3}} %AM to R
%\put(-6,9){\vector(3,-4){3}}
%\put(-4,9){\vector(1,-4){1}}
%\put(-2,9){\vector(-1,-4){1}}
%\put(-5,13){\vector(1,-4){1}}
%\put(-3,13){\vector(-1,-4){1}}
\put(0,8){\vector(0,-4){3}}
%\put(3,13){\vector(0,-4){8}}
%\put(3,13){\vector(-1,0){6}}
%\qbezier(3,13)(0,17)(-5,13)
%\put(-5.07,12.9){\vector(-1,-1){0}}
\put(0.3,8){$\sCM$}
%\put(3.3,12.8){\sAS}
%\put(-4,13){\sMC}
%\put(-6,13){\sUW}
%\put(-1.7,8.2){$\sTT$}
%\put(-5,8.2){$\sDE$}
%\put(-7.5,8.2){$\sRG$}
%\put(-4.5,4.8){$\sPD$}
\put(0.2,4.5){$\sCB$}
%\put(3.3,4.5){$\sAM$}
%\put(-1.4,0){\scriptsize{$R$}}
\put(-2.4,3){\scriptsize{$M_{22}$}}
\end{picture}
\hspace{.7cm}
\begin{picture}(10,14)%(-6,0)
%dalal23
\thicklines
\put(3,13){\circle*{.4}} %AS
%\put(0,8){\circle*{.3}} %CM
%\put(-5,13){\circle*{.3}} %W
%\put(-3,13){\circle*{.3}} %C
%\put(-6,9){\circle*{.3}} %RG
%\put(-4,9){\circle*{.3}} %D
%\put(-2,9){\circle*{.3}} %T
\put(3,5){\circle*{.4}} %AM
%\put(0,5){\circle*{.3}} %CB
%\put(-3,5){\circle*{.3}} %PD
%\put(0,1){\circle*{.3}} %R
%\put(-3,5){\vector(3,-4){3}} %PD to R
%\put(0,5){\vector(0,-4){4}} %CB to R
%\put(3,5){\vector(-3,-4){3}} %AM to R
%\put(-6,9){\vector(3,-4){3}}
%\put(-4,9){\vector(1,-4){1}}
%\put(-2,9){\vector(-1,-4){1}}
%\put(-5,13){\vector(1,-4){1}}
%\put(-3,13){\vector(-1,-4){1}}
%\put(0,8){\vector(0,-4){3}}
\put(3,13){\vector(0,-4){8}}
%\put(3,13){\vector(-1,0){6}}
%\qbezier(3,13)(0,17)(-5,13)
%\put(-5.07,12.9){\vector(-1,-1){0}}
%\put(0.3,8){$\sCM$}
\put(3.3,12.8){\sAS}
%\put(-4,13){\sMC}
%\put(-6,13){\sUW}
%\put(-1.7,8.2){$\sTT$}
%\put(-5,8.2){$\sDE$}
%\put(-7.5,8.2){$\sRG$}
%\put(-4.5,4.8){$\sPD$}
%\put(0.2,4.5){$\sCB$}
\put(3.3,4.5){$\sAM$}
%\put(-1.4,0){\scriptsize{$R$}}
\put(0.4,3){\scriptsize{$M_{23}$}}
\end{picture}
\hspace{1cm}
\begin{picture}(4,14)%(-8,-6)
%dalal23
\thicklines
\put(-3,9){\circle*{.4}} % PD
\put(0,9){\circle*{.4}} %CB
\put(3,9){\circle*{.4}} %AM
%\put(0,3.8){\circle*{.3}} %CB
\put(0,5){\circle*{.4}} %R
\put(0,9){\vector(0,-4){4}} %CB to R
\put(3,9){\vector(-3,-4){2.9}} % AM to R
\put(-3,9){\vector(3,-4){2.9}} %PD to R
\put(3.3,8.8){$\sAM$}
\put(-4.5,8.8){$\sPD$}
\put(0.3,8.8){$\sCB$}
\put(-1.4,4){\scriptsize{$R$}}
\put(-2.4,3){\scriptsize{$M_{24}$}}
\end{picture}
\end{center}
%dalal18
\caption{Decomposition of the model $M_2$ from Example \ref{prisonexample1}.}
\label{fig:radical_decom2} 
%\caption{Decomposition of  model in the
%%joe24
%  %  \TDCR\ model=.}\label{fig:radical_decom2}
%    \TDCR\ model.}\label{fig:radical_decom2}  
\end{figure}
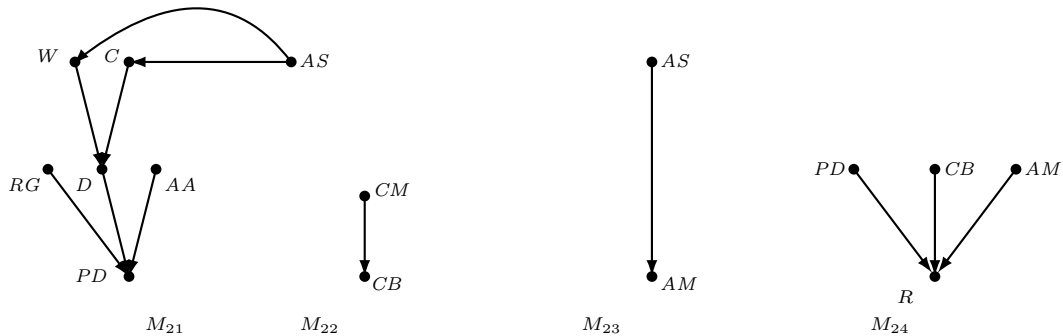

%dalal26: brought forward
We observe that decomposing  incompatible models into smaller
submodels can in some cases help determine common  interventions over
shared outcomes in the original models in spite of their
incompatibility.  Consider, for example, the two incompatible models
$M_1$ and $M_2$ in the prison example. Although the two are
incompatible (as observed in Example \ref{prisonexample1}), the
submodels $M_{12}$ and $M_{22}$ in Figures \ref{fig:radical_decom1}
and \ref{fig:radical_decom2}, respectively, obtained from their
decomposition, are fully compatible according to Definition \ref{combat}. We 
have $M_{12} \succeq_{\sCB}^{\vec{v}^*} M_{22}$ and $M_{12}
\succeq_{\sCM}^{\vec{v}^*} M_{22}$. The composition of the two
submodels  yields a merged model similar to $M_{12}$. Given this, it
may be concluded that interventions  over $\SC$ or $\IL$ make it
possible to change the value of $\CB$ in the two models $M_1$ and
$M_2$ and ultimately $R$ (assuming the structural equation for $R$ in
both models $M_{1}$ and $M_{2}$ to be the conjunction of $\PD$, $\CB$
and $\AM$)---a similar conclusion reached by considering partial
compatibility between the models  $M_1$ and $M_2$ in
Section~\ref{sec-part-compat}. We conjecture that reasoning using model
decomposition and full compatibility of submodels is an alternative to
reasoning using partial compatibility.

%dalal26
%The advantage of decomposing a causal model $M$ into a set of smaller submodels is that we can reason about each submodel 
Another advantage of decomposing a causal model $M$ into a set of
smaller submodels is that we can reason about each submodel
separately. In particular, we can compute the set of causes and
possible interventions for a given outcome. However, in order to use
these results to reason about the whole model, we need to perform 
additional calculations. Informally, when decomposing $M$ into a set of smaller submodels, we can view each submodel as a black box, with inputs and outputs
%joe22*: I don't think that this is quite right.  I rewrote 
%being the variables of the submodels. Then, this set induces an
%abstract causal graph based on the $\ll$ relationship
%between the submodels. This graph is an abstract version of the
%causal network induced by $M$. 
being the exogenous variables of the submodel and the leaves in the
causal graph of the submodel, respectively. We can then put connect
these variables into an abstract causal graph for the original model,
essentially ignoring the internal variables.
If the submodels are fairly large, the graph of 
submodels will be significantly smaller than the causal network of
$M$. We can then apply causal reasoning to the abstract graph, which
will result in a set of submodels 
being causes for the outcome. For these submodels, we can calculate the causes of their outcomes for each submodel separately. As causality is \textit{DP}-complete,
and computing interventions is co-\textit{NP}-complete, solving a set of smaller problems instead of a large problem is cheaper.

%hana22
%joe26*: I think that we need to say something about the connections
%between them, and why these are interesting
%We note that in fact, interesting decompositions (that is,
%decompositions of a large model to a set of submodels of a reasonable
%size) are only possible in models that are somewhat loosely
We note that, in fact, interesting decompositions (that is,
decompositions of a large model into a set of submodels of a reasonable
%hana23
size
with relatively few interconnections between them, which means that we
can analyse causality both within a submodel and between submodels
relatively easily) are possible only
in models that are somewhat loosely connected.
%hana23
%joe32
%Such models tend to be derived from real-life cases, where several
%factors are in play, possibly in different 
%%hana26 case study moves here from Section 6
%areas of expertise (see Example~\ref{ex:babyP} below for two such cases).
Such a decomposition can often be done for real-life cases; see 
Example~\ref{ex:babyP}.
%areas of expertise (see Section~\ref{sec:casestudies} for two such cases). 
%
%oje26*: this doesn't really follow from the previous sentence, but
%perhaps the previous sentence is enough.
%In other words, the decomposition depends on the number of
%paths in the model (or its degree of connectivity).  
%joe26: we haven't proved this
%On the other hand, the complexity analysis examines the worst case,
%which in causality occurs when a model has a large degree of connectivity.
We believe that, in practice, analyzing the effect of interventions
in a model will be difficult precisely when a model is highly
connected, so that there are many causal paths.
%joe26
%In practice, we 
%expect the causal models to be much more loosely connected, and hence
%amenable to decomposition. Hence, the computation of causes and
We expect the causal models that arise in practice to be much more
loosely connected, and thus
amenable to useful decompositions. Hence, the computation of causes and
interventions 
%joe26
%in practice is expected to have a much lower complexity both in the
%whole model and in the decomposed model.
in practice should not be as bad as what is suggested by our
worst-case analysis.
%The exact savings depend on the size
%of the submodels.
%
%joe22*
%trivial decomposition of an arbitrary model into a lot of submodels,
%where the set of endogenous variables for each submodel is a
%singleton.  (Thus, there is one submodel for each endogenous variable
%in the original model.)  This is not a very interesting decomposition.
%hana 18 removed this illustration of the reduction in complexity - to be replaced by a similar discussion on a smaller example
%In Example~\ref{decompose-example}, we demonstrate one example of such decomposition. Depending on the structural equations, we can determine the set of submodels that are causes for $PD$. If $PD = CS \wedge \neg{RFH} \wedge CA$ (that is,
%the child was declared safe in the family home by the social services, he was not removed from the family home by the court, and he was abused), and in the
%current context $CS = CA = 1$ and $RFH = 0$, 
%then each of the submodels for the Social services $2$, for the court and the police, and for the family is a cause of $PD$. We can then compute the set of
%causes for $CS$, for $CA$, and for $RFH$ separately, hence reducing the overall complexity of the computation.
%

%dalal23 I think we should a sentence about the purpose of this example
% what is meant to demonstrate? decomposition for reusability?
% It would be good to highlight what is special about this compared to
% the prisons
%joe33: It does show decomposition for resuability, but I think it
%also shows that what we're doing is someting that can be used in practice.
Below, we briefly discuss the relevant aspects of two cases of child
abuse that resulted in the death of a child: the ``Baby P'' case and 
%dalal23 Later we say this is to demonstrate that decompositions 
% supports reuse of models... if this is the case
%then I am not sure talking about partial compatibility of two
%"existing" models makes sense. 
% we should also add this as an advatnage as the start of this
% section.
%joe33: Are they in fact only partially compatible?  Do we actually
%show this?
the Victoria Climbi{\`e} case. In these cases, experts' opinions were in fact only partially
compatible, and there were natural ways to decompose the causal
model. 

% dalal23
%The example is very lengthy (spanning over three pages)... maybe that
%is ok but I would prefer if we could shorten the description of the
%cases if we can 
% and dive sooner rather than later into what the example is intended to show 
% in the spirit of previous examples if possible.
%joe33: I don't feel strongly like this, but I think we should perhaps
%also sell it as a more worked-out real-world example, thus justifying
%its length (at least partially).  Do you want to take a pass at a
%rewrite, Dalal?
%dalal18

%hana26 the babyP examples goes here
%\input{babyPexample}
%dalal24
%\begin{example}[The cases of ``Baby P'' and Victoria Climbi{\`e}]\label{ex:babyP}
\begin{example}[The cases of Baby P and Victoria Climbi{\`e}]\label{ex:babyP}
  %dalal23
  \normalfont
Baby P (Peter Connelly) died in 2007 
%``Baby P'' (Peter Connelly) was born in March 2006 and died in August 2007 
%in the London borough of Haringey 
%joe3
%after suffering physical abuse over a sustained period of time.
after suffering physical abuse over an extended period of time
%His death led to inquiry into
%the current practices of several governing bodies, because it occurred despite 
%him being listed in
%the Child Protection Register due to ``physical abuse and neglect'' and him and his carers being
%actively monitored by a system of ``joined-up governance'' involving determinedly collaborating 
%professionals and organisations 
%joe32
%\citet{Marinetto:2011}.
\cite{Marinetto:2011}.
%dalal24
%with baby Peter
%hana28: changed Baby P to baby Peter
The court ultimately found the three adults living in a home with baby Peter guilty of
%joe32
%``causing or allowing [Peter's] death'' \citet{SentencingRemarks:BabyP}.
``causing or allowing [Peter's] death'' \cite{SentencingRemarks:BabyP}. 
%He was seen regularly by medical professionals, social workers
% and other professionals involved in his case.  On a number of occasions, these professionals and their associated organisations had opportunities to remove Peter from the care of the people that caused or allowed the abuse to take place. 
%hana28 changed back
%dalal24
After baby Peter's death, there was an extensive inquiry into
%After Baby P's death, there was an extensive inquiry into
practices, training, and governance in each of the involved professionals and organizations separately.
%dalal23 added footnote
\footnote{\citet{CFKL15} provide a more detailed discussion
%hana28 changed back
%Dalal24
of the case of ``Baby P'', including a construction of the causal model.} 
  As shown by \citet{CFKL15}, the complete
  causal model for the 
%joe3
%  Baby P case is very complex 
  %  and involved many variables and interactions between them. It is,
    Baby P case is  complex,
involving many variables and interactions between them. 
%joe28*: later we say that we can't quite decompose this way.  So this
%is not correct.  I rewrote it, but please check that this is what you
%intended.  
%It is, however, decomposable into 
%%joe3: sub-model -> submodel globally
% % compatible sub-models as defined in
%  compatible submodels in the sense of
% Section~\ref{sec:combining}. Specifically, we identify 
%%joe3*: can we replace ``Family'' by ``Family life'' in the figure?
% % the sub-models of ``Family'', ``Social services and police'',
%the submodels of ``family life'',
%joe28: doesn't social services also deal with family life, or is there
%a service correspdonding to ``family life'' that's distinct from
%``social services''?  Is it called ``family life''?  ``Family
%services'' sounds better to me.  
%
%hana25
There were several authorities involved in the legal proceedings,
specifically social services, the police, the medical system, and the
court. In addition, the legal proceedings considered the family
situation of Baby P. 
Roughly speaking, the causal model
can be viewed as having 
the schematic breakdown presented in Figure~\ref{fig:babyP}. 
\begin{figure}[htb]
\centering
%joe20
%\scalebox{0.55}{\input{babyP.pdf_t}}
%joe28: the figure says '\phi = Baby P's death'', but it should be D =
%Baby P's death''.  I would also replace ``medical care'' by ``medical
%system''; it seems more like an ``authority'' that way.  Similarly, I
%would relabel ``family'' with ``family life'' or ``family services''
%(see the comment below)
\scalebox{0.7}{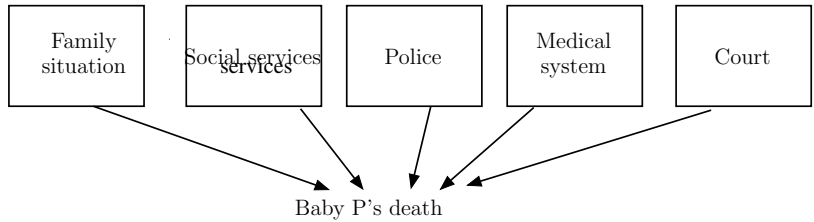}
%joe27: sub-model -> submodel globally
%\caption{Schematic representation of causal sub-models in the baby P
%dalal24
%\caption{Schematic representation of causal submodels in the baby P
\caption{Schematic representation of causal submodels in the Baby P
case.}\label{fig:babyP} 
\end{figure}

%joe3
%We  note that in this case, 
%joe28: I'm confused.  Is it the case that, in fact, there was one
%expert for each submodel.  If so, the next two sentence should be
%rewritten as ``In the actual case, there was exactly one expert for
%each of these authorities, who was assumed to be right (and thus can 
%be viewed as being assigned degree of confidence 1 in our
%framework).'', or something like that 
%hana25
%joe29
%There experts involved in the legal inquest and enquiry had expertise
%that fitted inside one of the boxes in Figure~\ref{fig:babyP} (that is,
%there were no experts with expertise in more than one box).
%The figure also suggests that we might divide the causal model into
Each of the experts involved in the legal inquest and enquiry had expertise
that corresponded to one of the boxes in Figure~\ref{fig:babyP} (i.e.,
there were no experts with expertise that covered more than one box).
%joe30: removed paragraph break
%
The figure suggests that we might divide the causal model into
submodels corresponding to each box. 
%end of commented out text
%joe29: removed paragraph break (and one before previous sentence)
%
%hana20 rewrote
%hana18
%joe28*: the flow here doesn't quite make sense.  We talked
%about a decomposition into submodels in the previous paragraph.  Are
%we talking about a different decomposition here?  I think what we're
%trying to say is that the decomposition of the previous paragraph is
%too naive.  I have no rewritten it to say that.  But if that's not
%what you intended, please rewrite it further.
%As evident from the case, there are several authorities involved in
%the legal proceedings, hence it makes sense to decompose
%%joe27
%%the causal model representing the case to several sub-models. One such
%the causal model representing the case to several submodels. One such
%decomposition is schematically presented in Fig.~\ref{fig:babyP}. 
%However, it does not take into account the interaction between
%dalal26
%Unfortunately, the schematic representation in 
The schematic representation in 
Figure~\ref{fig:babyP} does not take into account the interaction between 
%dalal24
%submodels. In reality, there were numerous interaction between, for
submodels. In reality, there were numerous interactions between, for
example, 
the social services and the court submodels, leading to court
hearings, which in turn determined the course of action taken by the
social services 
and the police after the court decision.
%joe28
Once we model these interactions more carefully, we need a somewhat
more refined decomposition.

%joe28
%In Example~\ref{decompose-example},
%consistent with Definition~\ref{def:decomposable} for a part of the
%case -- namely, the part that concerns the social services, the court
%and the %police, and the family life.
%joe29
%We give such a decomposition in Figures~\ref{fig:babyPcompose}
%that take into account the interactions for part of the case, 
We give a decomposition in Figure~\ref{fig:babyPcompose}
that takes into account the interactions for part of the case, 
namely, the part that concerns the social services, the court, the
police, and family life.  To make the decomposition consistent with
Definition~\ref{def:decomposable}, we break up social services
into two submodels, for reasons explained below.

\begin{figure}[htb]
\centering
%joe20
%\scalebox{0.5}{\input{babyPcompose.pdf_t}}
%joe28: could we relabel ``Family'' as ``family life'' or ``Family
%services''.  Perhaps ``CourtPolice'' can be ``Court+Police'',
%``SocialServices1'' could be ``Social Services #1'' and
%``SocialServices2'' could be ``Social Services #2''.
%joe29: I still think that we should relabel the boxes
%dalal23 @Hana could you please format the pictures so they have the same
%point size and line thickness as the previous diagrams?
\scalebox{0.7}{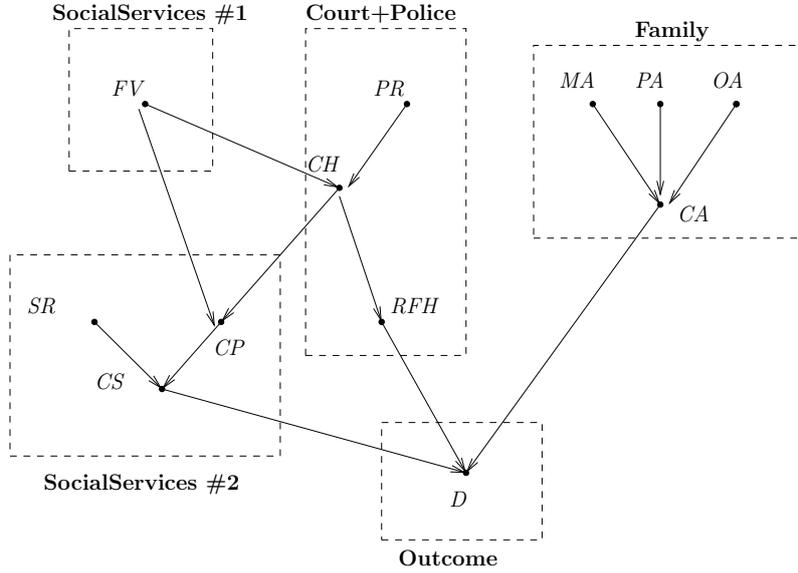}
\caption{Simplified causal model $M$ of a part of the Baby P case.}\label{fig:babyPcompose}
\end{figure}

%joe29: rewrote, but I'm not sure that I got it right.  Please check
%The variables in the figure are: $\textit{FV}$ for a family visit from
%the social services, $\textit{PR}$ for the police report,
%$\textit{CH}$ for the court hearing, 
%$\textit{RFH}$ for removal from home (of the child), $\textit{CP}$
%for whether the child was put on the Child Protection Register,
%$\textit{SR}$ for the social services report,
%$\textit{CS}$ for whether the child is declared safe in the family home,
%and the variables $\textit{MA}$, $\textit{PA}$, and $\textit{OA}$ are
%for the abuse of Baby P by his mother, the mother's partner, 
%or another adult in the house, respectively. The variable
%$\textit{CA}$ indicates whether the child was abused.
%The final outcome is $\textit{D}$, which stands for the death of baby
%P from abuse.
The variables in the figure are: $\textit{FV}$ for whether there was a
family visit from the social services; $\textit{PR}$ for whether there was
a police report;
$\textit{CH}$ for whether there was a court hearing;
$\textit{RFH}$ for whether the child was removed from home; $\textit{CP}$
for whether the child was put on the Child Protection Register;
$\textit{SR}$ for whether there was a social services report; 
$\textit{CS}$ for whether the child was declared safe in the family home;
$\textit{MA}$, $\textit{PA}$, and $\textit{OA}$ for whether the child was
abused by his mother, the mother's partner, or another adult in the
house, respectively; 
$\textit{CA}$ for whether the child was abused; and, finally, 
$\textit{D}$ for whether the final outcome was death (of
%dalal24
%baby P) due to abuse.
Baby P) due to abuse.
%joe28: added next few sentences explaining the rectangles
%dalal24 rephrased as we do 
%Note that, as usual, we have omitted exogenous variables in the
%figure; it shows only the endogenous variables.  For example, we do
Note that, as usual, we have omitted exogenous variables of the full
model in the 
figure; it shows only the endogenous variables.  Thus, we do
%joe29
%not have the the exogenous variables that determine $\textit{FV}$
%(family visits) or $\mathit{PR}$ (a police report).  The dotted
not have the exogenous variables that determine $\textit{FV}$,
%dalal24
%or $\mathit{PR}$.  The dotted
$\mathit{PR}$, $\mathit{MA}$, $\mathit{PA}$ or $\mathit{OA}$.  The dotted
%joe29
%rectangles in the figure determine a decomposition.  
rectangles in  Figure~\ref{fig:babyPcompose} determine a decomposition.
Each rectangle 
%dalal24
%consists of the engodenous variables in one submodel.  The exogenous
consists of the endogenous variables of one submodel.  The exogenous
%dalal24
%variables are the parents of the variables in the rectangle that are
%joe34
%variables for the  SocialServices\#2 and Outcome submodels are
variables of the  SocialServices\#2 and Outcome submodels are 
 those parent variables appearing in the other submodels.
%dalal24
%themslves not in the rectangle.  Thus, for example, in the submodel
%themselves not in the rectangle.  Thus, for example, in the submodel
Thus, for example, in the Outcome submodel,
%dalal24
%corresponding to $D$, the parents of $D$ are $\mathit{CS}$,
the exogenous variables are $\mathit{CS}$,
%dalal 24
%$\mathit{RFH}$, $\mathit{CA}$.  
%joe34
%$\mathit{RFH}$ and $\mathit{CA}$.
$\mathit{RFH}$, and $\mathit{CA}$.  
The submodels are described 
in Figure~\ref{fig:babyPsubmodels}.
The dotted rectangles in Figure~\ref{fig:babyPcompose} can 
%dalal24
%be viewed as providing a compact representation of the submodels in 
be viewed as  compact representations of the submodels in 
Figure~\ref{fig:babyPsubmodels}.

%hana24
%joe28: cut; essentially said above
%We show one possible decomposition of $M$ according to
%Definition~\ref{def:decomposable} in Figure~\ref{fig:babyPsubmodels}
%(note that the model 
%also contains exogenous variables that are not presented in the
%figure; these are inherited by the submodels depending on the
%structural equations for each 
%submodel's endogenous variables).
%joe28
%We note that there are two submodels for the
%social services. This is because 
%the variable $\textit{CH}$ depends on $\textit{FV}$, and the variable
%$\textit{CP}$ in turn depends on $\textit{CH}$, hence 
%$\textit{FV}$ and $\textit{CP}$  cannot be in the same submodel. 
Of course, there is more than one way to decompose the model of 
Figure~\ref{fig:babyPcompose}.  
%It is easy to see that $M$ can be decomposed in more than one way. For
For example, the submodel currently standing for the court and the police
%joe28
%can be in turn decomposed into two smaller submodels,
can be decomposed into two smaller submodels,
%joe28
%for the court and for the police.
one for the court and one for the police.
However, it is critical that we have decomposed social services into
two submodels.  
The variable $\textit{CH}$ depends on $\textit{FV}$, and the variable
$\textit{CP}$ in turn depends on $\textit{CH}$, hence 
$\textit{FV}$ and $\textit{CP}$  cannot be in the same submodel (or
else we would violate the requirement of
Definition~\ref{def:decomposable} that the sets of endogenous
variables of each submodel form an order-preserving partition of the
endogenous variables of the original model).
%We note that there are two submodels for the
%social services. This is because 
%hana20
%%the variable $\textit{CH}$ depends on $\textit{FV}$, and the
%%variable $\textit{CP}$ in turn depends on $\textit{CH}$, hence 
%$\textit{FV}$ and $\textit{CP}$  cannot be in the same submodel. 
%$FH \ll CH \ll FUV$, hence either all three of these endogenous variables should be in the same submodel, or, if $CH$ is placed in a different submodel, then the %variables $FH$ and $FUV$ should also be in different submodels according to Definition~\ref{def:decomposable}.
%joe28
%\begin{figure}
\begin{figure}[htb]
\centering
%joe20
%\scalebox{0.5}{\input{babyPcompose.pdf_t}}
%joe28: can the variable SR be moved a bit to the right?  Also, the
%separate submodels should be relabled as in the previous figure.
%joe29: I still think that we should relabel the boxes
\scalebox{0.7}{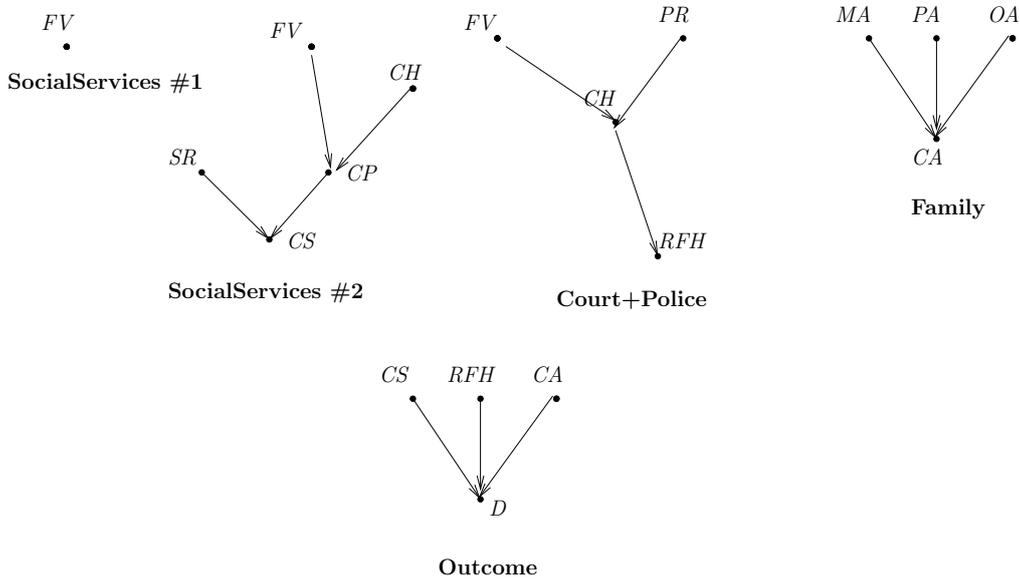}
%dalal21
%\caption{One possible decomposition of the model $M$.}\label{fig:babyPsubmodels}
\caption{One possible decomposition of the Baby P model $M$.}\label{fig:babyPsubmodels}
\end{figure}
%joe28
%\exam

%hana26 added the description of Victoria Climbie's case here, as a
%part of the example 
%dalal23 I am inclined to shorten the description of the case itself
% and focus the introduction of this case on how Baby P model's 
% decomposition can be used to construct Victoria's models
% so the comparison might be in terms of the description o f
%the case but not in terms of the models ...
%dalal24
%We compare the ``Baby P'' case with another case of child abuse that
%We compare the Baby P case with another case of child abuse that
We consider another case of child abuse that
resulted in child's death: Victoria Climbi{\`e}~\cite{Marinetto:2011}.
%joe31
%Victoria Climbi{\`e} died in 2000, 18 months after arriving in the UK
Victoria died in her house from hypothermia, 18 months after arriving in the UK 
from the Ivory Coast to live with 
her great-aunt.
%sustaining a series of injuries over 
%the 18 months that she was in custody of her great-aunt.
Her great-aunt and the great-aunt's boyfriend were
found guilty of Victoria's murder 
%dalal24
%(in contrast with ``Baby P'' case, where the adults in the house were
%joe34
%(in contrast with Baby P case, where the adults in the house were
(in contrast with the Baby P case, where the adults in the house were
found guilty of causing or allowing 
his death).

The inquiry into the circumstances of Victoria's death
placed the blame on social workers, 
who failed to notice Victoria's injuries, paediatricians, who accepted
the explanation of  
Victoria's great-aunt that Victoria's injuries were self-inflicted,
and the metropolitan police. In 
addition, the inquiry noted that the pastors in the church to which
Victoria's great-aunt belonged, 
had concerns about the child's well-being but failed to contact any
child protection services.
%joe31: moved from above
%dalal24 I am missing which one happened first. the suggestion below
%that baby p happened afterwards as the interventions on Victoria didnt
%prevent baby P? But I thought it was the other way around and hence
%the reuse argument
%joe34: Baby P died in 2007 and Victoria Climbie died in 2000, so the
%next sentence seems OK, although we may want to check that we don't
%suggest Baby P happened earlier somewher in the text.
%joe42
%The inquiry suggested several interventions into
The inquiry suggested several interventions on 
the procedures of social workers and paediatricians. These
interventions turned out to be inadequate, 
as the death of Baby P occurred under somewhat similar circumstances
and his abuse also went unnoticed until his death. 

%Victoria Climbi{\`e} died in her house from hypothermia, after
%sustaining a series of injuries over 
%the 18 months that she was in custody of her great-aunt.
%joe34
%As the
%joe31
%circumstances of her death were different from that of Baby P (he died
%dalal24: the last paragraph ended with a statement saying they had 
%similar circumstances and now we say they are different. Maybe 
%we can rephrase to specify what  about the circumstances was the same 
% and what was different from a causal model view: the variables themselves?
%joe34
%circumstances of Victoria Climbi{\'e}'s death were different from
%that of Baby P (he died
%in the hospital), the causal model differs in the set of dependencies
Although there were some similarities between the Baby P case and the
Victoria Clibmbi{'e} cases, there were also some differences.  For
example, while Vicitoria Climbie{\'e} died at home, 
Baby P died in the hospital.  Thus, the causal models for these two
cases differ somewhat.
%joe32
%it captures. Victoria Climbi{\`e}'s case is also complex with many
%joe34
%it captures. Victoria Climbi{\`e}'s case also involves many
%variables and interactions between them, and, similarly to the Baby
%joe32
%P's case, decomposable into compatible submodels in the sense of 
%dalal23 fully/partially
%joe33
%P's case, is decomposable into compatible submodels in the sense of
%P's case,
However, the causal model for the Victoria Climbi{\'e} case 
is also decomposable into fully compatible submodels in the sense of 
%dalal24 
%Section~\ref{sec:combining}. 
Section~\ref{sec:decomposition}. 
 %hana25
 %joe30
% The submodels are, however, a bit different and reflect the
 %joe31:
%joe34
%While the causal model for Victoria Climbie{\`e}'s case is different
 %dalal24
 % from that of baby P, some of the submodels in the decomposition are
 %dalal24: I assume what is meant is the names of the submodels, i.e.,
 %what they represent rather than causal models in the specific sense
% from that of Baby P, some of the submodels in the decomposition are
% identical.
Moreover, some of the submodels in the decomposition are identical to
those in the causal model for Baby P.
Specifically, there are submodels for the police, the
 medical system, the family system, and the courts, just as in the
 %dalal24
 % case of baby P, as well as a submodel for the church.  
 case of Baby P, as well as a submodel for the church.  
% The submodels are, however, a bit different, and reflect the
% authorities involved in the legal proceedings of Victoria
% Climbi{\`e}'s case: 
%the social services, the police, the medical system, and the
%church. In addition, as in the Baby P case, the legal proceedings
%considered the family situation of 
%Victoria Climbi{\`e}. 
 %joe28: removed paragraph break
 %
 The schematic breakdown is presented in Figure~\ref{fig:victoria}. 
\begin{figure}[htb]
\centering
%joe20
%\scalebox{0.55}{\input{victoriaclimbie.pdf_t}}
\scalebox{0.7}{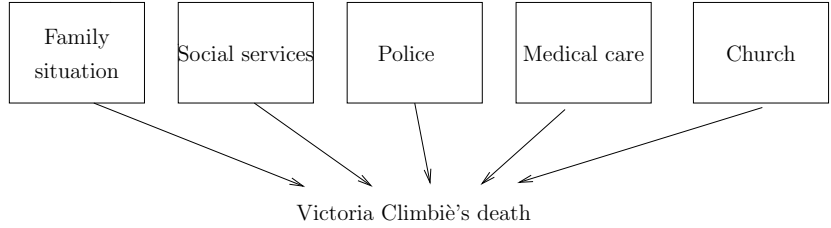}
%joe33
%\caption{Schematic representation of causal sub-models in the Victoria
\caption{Schematic representation of causal submodels in the Victoria
  Climbi{\`e}'s case.}\label{fig:victoria} 
\end{figure}
%joe20*: Again, we should point out that the model is decomposable,
%just as in the baby P case.
%joe31*: added; this is the important punchline (at least, to me).
%dalal23
%Although we don't provide the causal model in detail here, this
Although we do not provide the causal model in detail here, this
discussion already illustrates a major advantage of decomposition: it
%dalal23 This is the point that I find confusing a bit 
%joe33: I'm not sure what the problem is here?  Could you clarify, Dalal?
%dalal24: reuse suggests to me that the new model doesnt fully exist and 
% you use another model to build the new one...
%joe34: it seems OK to me as written, but feel free to rewrite it ...
allows us to reuse causal models that were developed in one case and
apply them to another, thus saving a lot of effort.  
%hana27
%dalal23 Then there is another advantage here... Once we clarify the purpose
% of the examples I think we should state them explicitly before the example
% description
%joe33: again, sub-model -> submodel; also, we have no notion of
%similarity, so I think we shold downplay the similarity part
%Moreover, if the same (or similar) sub-model appears in several
%different cases, such as the social services sub-model in these
Moreover, if the same  submodel appears in several
different cases, such as the social services submodel in these
%joe33
%examples, this can guide 
%the policymaker to prefer interventions that address the problems
%%demonstrated by this sub-models, as they are likely to affect %several
examples, this suggests that 
the policymaker should prefer interventions that address the problems
%demonstrated by this sub-models, as they are likely to affect several
demonstrated by this submodel, as they are likely to affect several
cases. In fact, 
the cases of child abuse that remains undetected due to problems in
the social services sadly continue to occur (see, for example, the
recently 
%joe33: was it one or several cases?
%published cases in~\cite{Ind19}).
published cases discussed in~\cite{Ind19}).
Even though the
causal models for different cases will undoubtedly be different, we
can still take advantage of the common submodels.  We expect that this
will be the case in many other situations as well.
%dalal23
\eprf
\end{example}

From a practical perspective, Example~\ref{ex:babyP} demonstrates one
%dalal27
%benefit of decomposition: the decomposition allow us
benefit of decomposition: the decomposition allows us
to capture different aspects of the case, each requiring different expertise.
%In Example~\ref{ex:babyP} we considered the causal models for two legal cases. 
%In these cases, decomposition can have additional benefits.
%We provided one decomposition that takes into
%account the different aspects of the case, requiring different expertise.
%dalal26
%This allows different experts two work on each of the submodels
%dalal27
%This allows different experts to work on each of the submodels independently. 
This facilitates different experts working on each of the submodels independently. 
%hana21
%joe26
%The process can also work in the other direction: a policymaker
%usually has a crude idea about how the causal model should look like,
The process also works in the other direction: a policymaker
often has a crude idea of the general structure of the causal model,
and what components 
are involved in the decision-making process. She can then decompose her initial causal model into submodels and, guided by these submodels, decide which
%joe26
%experts are needed to analyse each of the submodels.
areas of expertise are critical.

%joe32 shortened
%hana21
%Finally, the decomposition can be helpful if we have several cases
%that differ from each other for the most part, but are similar in a
%specific submodel. Then,
%interventions computed in this submodels can be relevant for both
%Finally, a decomposition can be helpful if we have several cases
%that have both similarities and differences, and the similarities can
%be captured by a submodel.
%dalal26
%A second benefit of decomposition illustrated by these examples is
A further benefit of decomposition illustrated by these examples is
that, although different, the causal models had some common submodels.
%dalal26
%Thus, decomposition allows from modularity in the analysis, and allows
%dalal27
%Thus, decomposition allows for a form of modularity in the analysis, and allows
Thus, decomposition supports a form of modularity in the analysis, and enables
results of earlier analyses to be reused.
\section{Combining Experts' Opinions}\label{sec:experts}
%\section{Combining Experts' Opinions for Interventions}\label{sec:experts}

%dalal23 added  introduction
In this section, we show how we can combine experts' causal options.
%joe33: I think we discuss only full compatibility, so I cut this
%We discuss cases where (some of) the models are either fully compatible,
%partially compatible or  neither, 
%by referring to the examples presented in the previous section.
%
%joe10
%We assume that we have a collection of pairs $(M_1,p_1),
Suppose that we have a collection of pairs $(M_1,p_1)$,
\ldots, $(M_n, p_n)$, with $p_i \in (0,1]$; we can think of
  $M_i$ as the model that expert $i$
  thinks is the right one and 
%joe32
  %  $p_i$ as the policymaker's degree of confidence
    $p_i$ as the policymaker's prior degree of confidence
  that expert $i$ is correct.
  %joe32: slowing down
(The reason we say ``prior'' here will be clear shortly.)  
Our goal is to combine the expert' models.  We present one way of
doing so, that uses relatively standard techniques.  The idea is to
%dalal23 typo
%treat the probabilities $p_1, \ldots, p_n$ as mutually inependent.
treat the probabilities $p_1, \ldots, p_n$ as mutually independent.
Thus, if $I$ is a subset of $\{1,\ldots,n\}$, the prior
probability that 
exactly the experts in $I$ are right, which we denote $p_I$, is
$p_I= \prod_{i \in I} (p_i)*\prod_{j \not\in I} (1-p_j)$.   
Now we have some information regarding whether all the experts in $I$
are right.  Specifically, if the models in $\{M_i: i \in I\}$ are not
mutually compatible, then it is impossible that all the experts in $I$
are right.  Intuitively, we want to condition on this information.  
We proceed as follows.
  
%
%joe12: to avoid bad line streteching
  %  Let $\Compat = \{I \subseteq \{1, \ldots, n\}: \mbox{the models in
%    $\{M_i: i \in I\}$ are mutually compatible}\}$.  For $I \in
%joe35: adding default setting
Fix a default setting $\vec{v}^*$ of the exogneous variables that are
not common to $M_1, \ldots, M_n$
Let $\Compat^{\vec{v}^*} = \{I \subseteq \{1, \ldots, n\}:$ the models in 
$\{M_i: i \in I\}$ are mutually compatible
%joe35
with respect to $\vec{v}^*\}$.  For $I \in 
    \Compat^{\vec{v}^*}$, define $M_I = \oplus_{i \in I}^{\vec{v}^*}
  M_i$.
  %joe3
  By Proposition~\ref{oplusproperties}, $M_I$ is well defined.
  %joe32: slowing down again; also correcting problem noted by reviewer
  %  The policymaker considers the models in
%    $\M_{\Compat} = \{M_I: I \in \Compat\}$, placing
%the probability of $p_I= \prod_{i \in I} (p_i)*\prod_{j \not\in I} (1-p_j)/N$
%on $M_I$,
  The models in $\M_{{\Compat}^{\vec{v}^*}} = \{M_I: I \in \Compat^{\vec{v}^*}\}$ are the
  %dalal23
  %  candidate combined models that the policymaker should consider.
  candidate merged models that the policymaker should consider.
$M_I$ is the ``right'' model provided that exactly the experts in $I$
  are right.
%joe32: moved up from below
  But even if $M_I \in \M_{\Compat^{\vec{v}^*}}$, it may not be
  the ``right'' model, since it may be the case that not all the
  expert in $I$ are right.
The probability that the policymaker should give $M_I$ is $p_I/N$, 
%joe13*: I think think that p_I should be \prod_{i \in I} p_i \prod_{j
%\notin I} (1-p_j)/N.  That is, we want al the experts in I to be
%right and all the other experts to be wrong.  
%  probability $p_I = y$ on $M_I$, where
where  $N = \sum_{I \in \Compat^{\vec{v}^*}} p_I$ is a normalization factor.
%
  %joe2: added intuition
%joe11: line shaving
  %  Intuitively, we are viewing the events ``expert $i$ is right'' as
%joe32: cut the material below, since it's alread said above
%Intuitively, we view the events ``expert $i$ is right'' as
% being mutually independent, for $i = 1, \ldots, n$.  Thus, $p_I$ is
%the probability of the event that 
%  exactly the experts in $I$ are right (and the ones not in $I$ are
%  wrong).  If exactly the experts in $I$ are indeed
%  right, it seems reasonable to view $M_I$ as the ``right''  causal
%  model.  Note that it 
%  is not possible for all the experts in $I$ to be right if there are
%  experts $i, j \in I$ such that $M_i$ and $M_j$ are incompatible.
%  Thus, we consider only models $M_I$ for $I \in \Compat$.  But even
%  if $I \in \Compat$, it is possible that some of the experts in $I$
%  are wrong in their causal judgments.

%joe32: added next line
%Our calculation implicitly
This approach gives the policymaker a distribution over causal models.
This can be used to compute, for each context, which interventions
affect the outcome $\varphi$ of interest, and then  
compute the probability that a particular intervention is effective
(which can be done summing the probability of the models $M_I$ in
$\M_{\Compat^{\vec{v}^*}}$ where it is effective, which in turn can be computed
as described in Section~\ref{sec:intervention}).
Note that
our calculation implicitly
  conditions on the fact that at least one expert is right, but allows
  for the possibility that only some subset of the experts in $I$ is
%joe3
  %  right even if $I \in \Compat$.
%joe11
%  right even if $I \in \Compat$, so we place positive probability on
  right even if $I \in \Compat^{\vec{v}^*}$; we place positive probability on
  $M_{I'}$ even if $I'$ is a strict subset of some $I \in \Compat^{\vec{v}^*}$.
  %hana added connection to witnesses
%joe5: also corrected the FNB bib entry; you should re-bibtex.
%    This method of combining experts' judgments is similar, in spirit, to the
%joe8
% This method of combining experts' judgments is similar, in spirit, to the
 This method of combining experts' judgments is similar in spirit to the
%joe32
 % method proposed by Dawid~\citeyear{Daw87} and Fenton
%  et~al.~\citeyear{FNB16}.
  method proposed by \citet{Daw87} and \citet{FNB16}.

  %joe32: this now seems redundant
  \commentout{
 This completes our description of how to combine experts' causal
judgments.  At a high level, for each subset of experts whose
judgments are compatible (in that the models they are proposing are
%dalal23
%pairwise compatible), we combine the models, and assign the combined
pairwise compatible), we merge the models, and assign the merged
model a probability corresponding the probability of the experts in
the subset.  Of course, once we have a probability on the settings in
$\M_{\Compat}$, we can compute, for each setting,  which interventions
affect the outcome $\varphi$ of interest, and then  
compute the probability that a particular intervention is effective.

%hana added a discussion of the policymaker's strategy and complexity
  The straightforward strategy for a policymaker to compute the most
effective intervention based on 
the experts' opinions and the degree of confidence of the policymaker
in each expert's 
judgment is to compute the set $\M_{\Compat}$ of models and
then to apply 
the computation of interventions as described in
Section~\ref{sec:intervention} to each 
$M_I \in \M_{\Compat}$.
  }
  %joe32: \end{commentout}
%joe3: added
%dalal11:tighten
%The expert can then compute the 

\begin{figure}
%}
%dalal16:
%\shortv{
%\begin{wrapfigure}[10]{r}{0.15\textwidth} 
%}
\begin{center}
%joe18
  %  \setlength{\unitlength}{.12in}
%joe20
  %  \fullv{\setlength{\unitlength}{.12in}}
    \fullv{\setlength{\unitlength}{.18in}}
\shortv{\setlength{\unitlength}{.09in}}
%dalal16:
%\begin{picture}(5,5.6)
%\put(2,0){\circle*{.2}}
%\put(4,6){\circle*{.2}}
%\put(0,6){\circle*{.2}}
%\put(0,3){\circle*{.2}}
%\put(4,3){\circle*{.2}}
%\put(4,6){\vector(0,-4){3}}
%\put(0,3){\vector(2,-3){2}}
%\put(0,6){\vector(0,-4){3}}
%\put(4,3){\vector(-2,-3){2}}
%\put(2.7,-.2){\scriptsize{$F$}}
%\put(-1.2,5.8){\scriptsize{$C$}}
%\put(-1.2,2.8){\scriptsize{$P$}}
%\put(4.45,2.8){\scriptsize{$Y$}}
%\put(4.45,5.8){\scriptsize{$R$}}
%joe35*: the figure is not centered with respect to the caption.  Is
%there an easy way to slide it over?
\begin{picture}(-1,5.6)
\thicklines
\put(5,2.5){\circle*{.4}}
\put(0,5){\circle*{.4}}
\put(0,0){\circle*{.4}}
\put(-5,5){\circle*{.4}}
\put(-5,0){\circle*{.4}}
\put(-5,0){\vector(1,0){5}}
\put(-5,5){\vector(1,0){5}}
\put(0,0){\vector(2,1){5}}
\put(0,5){\vector(2,-1){5}}
\put(-6,5.5){\scriptsize{$R$}}
\put(-5.8,-1){\scriptsize{$C$}}
\put(-.2,-1){\scriptsize{$P$}}
\put(-.2,5.5){\scriptsize{$Y$}}
\put(5.5,2.3){\scriptsize{$F$}}
\end{picture}
\end{center}
%joe8
%\caption{Third expert  on Famine (and  combined model).}\label{fig7}
%dalal16
%\vspace{-4pt}
%dalal23
%\caption{Third expert's (and  combined) model of famine.}\label{fig7}
\caption{Third expert's (and  merged) model of famine.}\label{fig7}
%\fullv{
\end{figure}
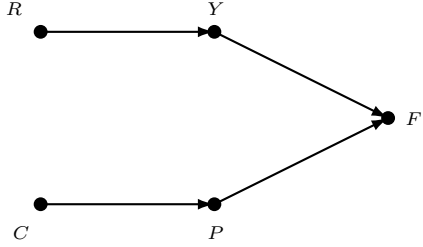

%hana9 changed 0.2 of textwidth to 0.17 of textwidth
%joe5: moved back one paragraph
%joe10*: this figure is in the wrong place when I latex the paper
%dalal9: moved the fig from here to fix formatting
%joe5
%Consider a variant of Example \ref{ex:famine}, in which a third
To get a sense of how this works, consider a variant of Example
\ref{ex:famine}, in which a third 
expert provides her view 
%joe8
%on causes on Famine and thinks that government corruption is an
on causes on famine and thinks that government corruption is an
indirect cause via its effect on   
%joe5
%political conflict (See Figure~\ref{fig7}); we  denote this model as $M_3$.  
%joe8
%political conflict (see Figure~\ref{fig7}); we  denote this model as $M_3$.
political conflict (see Figure~\ref{fig7}); call this model $M_3$.
%joe35
For simplicity, we assume that all models have the same set of
exogenous variables. 
According to the compatibility definition in Section
%joe7
%\ref{sec:combining}, the models $M_2$ and $M_3$ are  compatible,
%dalal23 
%\ref{sec:combining}, the models $M_2$ and $M_3$ are  compatible
\ref{sec:combining}, the models $M_2$ and $M_3$ are fully compatible
(assuming that MI3 holds),  
%joe12*: I think that the calculation below is incorrect!  First, we
%should certainly have {M_1} in Compat.   I can change them, but I
%would appreciate a sanity check to make sure that I'm not missing anything.
%dalal10:
%but $M_1$ and $M_3$ are not.  We have $\Compat= \{\{M_2\},
but $M_1$ and $M_3$ are not.  We have $\M_\Compat= \{\{M_1\}, \{M_2\},
%joe7
%\{M_3\},\{M_2,M_3\}\}$ with $M_{2,3}=M_2 \oplus M_3$  is the same as $M_3$.  
%dalal10:
%\{M_3\},\{M_2,M_3\}\}$ with $M_{2,3}=M_2 \oplus M_3 = M_3$.  
\{M_3\},\{M_{2,3}\}\}$ with $M_{2,3}=M_2 \oplus M_3 = M_3$.  
Suppose that experts  are assigned the confidence values as follows: $(M_1,0.4)$, $(M_2,0.6)$ and  $(M_3,0.5)$ respectively.
%dalal11
Then the probability on $M_{2,3}$  is the probability of  $M_2$ and $M_3$ being right \fullv{(i.e., $0.6*0.5$)}
and $M_1$ being wrong\fullv{~(i.e., $1-0.4=0.6$)}. So we have 
% dalal10:
%$p_{2,3}  = (0.5*0.4+0.5*0.6 +0.5*0.6)/1.1=0.72$,
%joe13*: this will need to change if you agree with my earlier
%comment.  But even with the current definition, I don't understand
%the computation.  is p_{2,3} meant to be the probability of M_{2,3}?
%Why isn't it just .6 * .5 = .3 under the current definition?  With
%the new definition. it should be .6 * .6 *.5 = .18.  I also don't
%understand the denominator (which I think should be .08 + .1 + 12 +
%.18 under the new definition, and .4 + .6 + .5 + .3 = 1.8 under the current
%definition 
%dalal23  commented the below and added the information below
%$p_{2,3}  = (0.6*0.5*0.6)/0.56=0.32$ (where $0.56$ is the normalization factor). 
%\shortv{In a similar way we compute 
%  $p_1 = (0.4*0.4*0.5)/0.56=0.14$.}
%  \fullv{The probabilities on the other models is as follows.
  % dalal23
  
  %dalal25: the probabilities add up to 0.99
  \[
  \begin{array}{lcl}
  %dalal23
  p_1 &= & 0.4*0.4*0.5/0.56=0.14\\ %0.08/0.56
  p_2 &= & 0.6*0.6*0.5/0.56=0.32\\ % 0.18/0.56
  p_3 &= & 0.6*0.4*0.5/0.56=0.21\\ %0.12/0.56
    %dalal23
  p_{2,3}  &= & 0.6*0.5*0.6/0.56=0.32 %0.18/0.56
  \end{array}
  \]
  \noindent
  %dalal23
 % The normalization factor $N$ is simply $0.08+0.18+0.12+0.18=0.56$.
  where $0.08+0.18+0.12+0.18=0.56$ is the normalization factor $N$.
  %dalal23 commented out
  %}
%
%hana12 switched to normal figure for the full version
%dalal16:
%\fullv{
%}
%\shortv{
%\end{wrapfigure}
%}

%dalal23 I wonder whether we should add examples on combining 
% partially compatible models and in the case of decomposition.
% it would be good to refer to more examples here bit in a way
%that shows how we can handle different cases

%dalal25
Let us  consider the Sampson's domestic violence models  as another
point of 
illustration. The model shown in Figure \ref{sampsoncom} is the result
of merging the two fully  
compatible models given in  Figure \ref{fig6}. We thus have
$\M_\Compat= \{\{M_1\}, \{M_2\}, 
\{M_{1,2}\}\}$ with $M_{1,2}=M_1 \oplus M_2$ as given in Figure
\ref{sampsoncom}.  
%joe35
%Assuming the expert 1 is assigned a confidence value $0.6$ and expert
Assuming that expert 1 is assigned a confidence value $0.6$ and expert
2 is assigned $0.7$, then we have 
\[
  \begin{array}{lcl}
  p_1 &= & 0.6*0.3*0.58   /0.44=0.23\\ %0.1/0.44
  p_2 &= & 0.4*0.7*0.58 /0.44=0.36\\ %0.16/0.44
  p_{1,2} &= & 0.6*0.7*0.42 /0.44=0.41\\ %0.18/0.44
  \end{array}
  \]

Note that the number of models in
%joe3
%$\M_{\Compat}$ is exponential
%in the number of experts $n$.  For example, if all experts are compatible,
$\M_{\Compat}$ may be exponential 
%dalal23 in which sense fully/partially
%joe33
%in the number of experts.  For example, if all experts are compatible,
in the number of experts.  For example, if the experts' models are
mutually compatible, then
%joe3
%it corresponds to the number  of subsets of $n$.
$\Compat$ consists of all subsets of $\{1,\ldots, n\}$.
%hana2: agree - the total number is still only single exponent
%joe3*: no!  Each model is small.  The number of variables is at most
%the *sum* of the variables in each one.
%The straightforward computation of interventions per model is exponential
%in the size of the model, resulting in total double-exponential
The straightforward computation of interventions per model is exponential
in the number of variables in the model.  Since the number of 
%dalal23
%variables in a combined model is at most the sum of the variables in
variables in a merged model is at most the sum of the variables in
each one, the problem is exponential in the number of experts and the
total number of variables in the experts' models.  
%blow-up for the whole 
%computation (the parameters are different for each exponent, but both
%are bounded by the size of the input).
In practice, however, we do not expect this to pose a problem.
%joe3
%As we argued
%in Sec.~\ref{sec:intervention}, the policymakers, in practice,
%restrict their attention to small (or cheap) 
%interventions, effectively bounding the size of the interventions
%under consideration by a small 
For the problems we are interested in, there are typically few experts
involved; moreover, as we argued
%joe3
%in Sec.~\ref{sec:intervention}, policymakers, in practice,
in Section~\ref{sec:intervention}, policymakers, in practice,
restrict their attention to interventions on a small set of variables.
%joe12: line shaving
%Thus, we expect that the computation involved will be quite manageable.
Thus, we expect that the computation involved to be manageable.
%constant, which results in a polynomial running time. In addition,
%while the number of models 
%in $\M_{\Compat}$ is exponential in the number of experts $n$, in
%practice $n$ is very small. In fact, 
%it is rare to come across a problem for which there are more than
%$10$ experts' opinions, and usually the number of much smaller. 
%
%joe3: this is true, but I don't think it's significant. We've already
%made our case that it's reasonable.
%Finally, in many cases not all experts are compatible. In particular,
%this can happen 
%if the experts refer to different contexts (see the radicalization
%case study in Sec.~\ref{sec:casestudies}). 
%Since incompatible models cannot be combined into a single model, the
%sets of interventions 
%are computed separately. In some cases the proposed interventions can
%have a (partial) overlap, 
%which can correspond to the best intervention
%overall, though there is no guarantee of this happening.

%joe3
%[[WE SHOULD HAVE A SIMPLE EXAMPLE HERE.]]
%joe5
%Up to now, we have assumed that experts propose only one causal
%Up to now, we have assumed that each expert proposes only one causal
%model
%dalal15:
%and that this model is deterministic.
%dalal15:
Up to now, we have assumed that each expert proposes only one
deterministic causal 
model.
%joe16
%If an expert is uncertain about the model, she may propose
An expert uncertain about the model can propose
%dalal23 incompatible according to full/partial compatibility?
%joe33: I don't think it matters here
several (typically incompatible) models, with a probability
distribution on them.  We can easily extend our framework to handle this.

%dalal23 added a line
Suppose that expert $i$, with probability $p_i$ of being
%joe5
%correct,  $p_i$ of being right,
%joe16
correct,
proposes $m$ models $M_{i1}, \ldots, M_{im}$,
where model $M_{ij}$ has probability $q_j$ of being the right one,
%dalal14:
%according to expert $i$.  We can handle this by simply replacing
%joe16
%according to expert $i$.  To handle this, we simply replace
according to $i$.  To handle this, we simply replace
%joe16
%expert $i$ by $m$ experts, $i1, \ldots, ij$, where expert $ij$
%hana17
expert $i$ by $m$ experts, $i_1, \ldots, i_m$, where expert $i_j$
proposes model $M_{ij}$ with probability $p_iq_j$ of being correct.
%joe5
%As long each of a few experts has a probability on only a few models,
%dalal14:
%joe16: the ``While'' sounds funny to me here; I realize you're doing
%it to shave a line, so I undid this change and made other minor changes.
%While each of a few experts has a probability on only a few models,
As long as each of a few experts has a probability on only a few models,
this will continue to be tractable.

%dalal23 ideally an example of this should be introduced

%hana26 moved to Section 4.3 as examples
%
%%joe10: I don't think we'll have the room for these
%\fullv{
%\section{Case Studies }\label{sec:casestudies}
%
%%dalal17:
%%In this section, we discuss the application of the framework to
%%several case studies, demonstrating 
%In this section, we discuss the application of the framework to two legal cases, demonstrating
%the concepts of compatibility  and combinability and their effect on determining the best
%interventions.

%\input{babyPcase}
%
%\input{victoriaclimbiecase}

%joe20*: Section 6.3 repeats material already in Section 4.2 (on
%partial compatibility) where the prison example is used for
%motivation.  We should think about how we want to restructure this.
%\input{radfull}
%}

%joe4: added
\fullv{
\section{Conclusions}\label{sec:conclusions}
We have provided a method for combining causal models whenever
possible, and used that as a basis for combining experts' causal
judgments in a way that gets around the impossibility result of
%joe10
%Bradley, Dietrich, and List \citet{BDL14}. We provided a gradual weakening
%joe32
%Bradley, Dietrich, and List \citeyear{BDL14}. We provided a gradual weakening
\citet{BDL14}. We provided a gradual weakening
%dalal23
%of our definition of compatibility, allowing us to combine models that only agree
%joe33
%of our definition of compatibility, allowing us to merge models that
of our definition of full compatibility, allowing us to merge models that
only agree 
on some of their parts.
Our approach can be viewed as a formalization of what was
%joe32
%done informally in earlier work \citet{CFKL15,Sampson:2013}.  Our
done informally in earlier work \cite{CFKL15,Sampson:2013}.
%joe32*: slowing down here, in part to address concerns about applicability
%Our
%%dalal20
%%analysis of the case studies suggests that our approach can be applied
%analysis of the two cases suggests that our approach can be applied
%in practice.
While our requirements for compatibility are certainly nontrivial,
the examples that we have considered do suggest that our approach is
quite applicable.  That said, it would be interesting to consider
alternative approaches to combining experts' models.  The approach
considered by \citet{FH18} is one such approach; there may well be others.

%joe32
%We believe that using causal models as a way of
In any case, we believe that using causal models as a way of
formalizing experts' judgments, and then providing a technique for
combining these judgments, will prove to be a powerful tool with which
to approach the problem of finding the best intervention(s) that can
be performed to ameliorate a situation.
}

%hana12 slightly shortened the conclusions to fit in the page limit
%dalal9:
\shortv{\section{Conclusions}
We have provided a method for combining causal models
and used that as a basis for combining experts' causal
judgments in a way that gets around the impossibility result of
%joe10
%Bradley, Dietrich, and List \citet{BDL14}. We provided a gradual weakening
%oe32
%Bradley, Dietrich, and List \citeyear{BDL14}.
\citet{BDL14}.
%joe12: rewrote slightly and left to full paper
%We provided a gradual weakening of our definition
\fullv{We gave successively more general definitions
%dalal23
%of compatibility, allowing us to combine models that
of compatibility, allowing us to merge models that
only agree on some of their parts.}
Our approach can be viewed as a formalization of
%joe32
%an earlier work \citet{CFKL15,Sampson:2013}.
earlier work \cite{CFKL15,Sampson:2013}. 
We believe that using causal models as a way of
formalizing experts' judgments, and then providing a technique for
combining these judgments, will prove to be a powerful tool
%joe12
%with which to approach the problem of finding the best intervention(s)
%that can be performed to ameliorate a situation.
for finding the intervention(s)
best ameliorate a situation.}

%joe14: added. Do either of you want to add anything here.
%joe21: moved  acknowledgements and references to end of paper (after
%the appendix)

\appendix

%joe10
\fullv{
\section{Proof of Proposition~\ref{oplusproperties}}\label{proof-oplusproperties}

\prf
For part (a), suppose that $M_1 \sim_C^{\vec{v}^*}
M_2$ and, by way of contradiction, that $\Pa_{M_1}(C) \ne
\Pa_{M_2}(C)$.  We can assume without loss of generality that there is
some variable $Y \in   \Pa_{M_1}(C) - \Pa_{M_2}(C)$.
Let $\vec{Z} = \Pa_{M_1}(C) - \{Y\}$.  Since $Y$ is a parent of $C$ in
$M_1$, there must be some setting $\vec{z}$ of the variables in
$\vec{Z}$ and values $y$ and $y'$ for $Y$ such that 
$F^1_C(y,\vec{z}) \ne F^1_C(y',\vec{z})$ in $M_1$,  where
$F^1_C = \F_1(C)$.  Suppose that $F^1_C(y,\vec{z}) = c$ and
$F^1_C(y',\vec{z}) =c'$.  Let $\vec{X} = ((\U_1 \cup \V_1) \cap (\U_2
        \cup \V_2))$.  By MI1, $(\Pa_{M_1}(C) \cup \Pa_{M_2}(C))
        \subseteq \vec{X}$.  Let $\vec{x}$ be a setting of the
        variables in $\vec{X}-\{C\}$ such that $\vec{x}$ agrees with
        $\vec{z}$ for the variables in $\vec{Z}$ and $\vec{x}$
        assigns $y$ to $Y$.  Let $\vec{x}'$ be identical to $\vec{x}$
        except that it assigns $y'$ to $Y$.  Since the values of the
        variables in $\Pa_{M_1}(C)$ determine the value of $C$ in
        $M_1$, for all contexts $\vec{u}_1$ for $M_1$, we have
        $(M_1,\vec{u}_1) \satt [\vec{X} \gets \vec{x}](C=c)$ and
        $(M_1,\vec{u}_1) \satt [\vec{X} \gets \vec{x}'](C=c')$.
        Since $\vec{x}$ and $\vec{x}'$ assign the same values to all
        the variables in $\Pa_2(C)$, we must have 
        $(M_2,\vec{u}_2) \satt [\vec{X} \gets \vec{x}](C=c)$ iff
        $(M_2,\vec{u}_2) \satt [\vec{X} \gets \vec{x}'](C=c)$ for all
        contexts $\vec{u}_2$ for $M_2$.  Thus, we get a contradiction
        %joe39
        %to MI4$_{M_1,M_2,C}$.  It follows that $\Pa_{M_1}(C) =
to MI4$_{M_1,M_2,C,\vec{v}^*}$.  It follows that $\Pa_{M_1}(C) =
        \Pa_{M_2}(C)$.
	% dalal3: change $\Pa_1(c)$ => $\Pa_1(C)$
        The fact that $\F_1(C) = \F_2(C)$ also follows from
        %joe39
%        MI4$_{M_1,M_2,C}$.  For suppose that $\vec{z}$ is a setting of
        MI4$_{M_1,M_2,C,\vec{v}^*}$.  For suppose that $\vec{z}$ is a setting of
        the variables in $\Pa_1(C) = \Pa_2(C)$ and $\vec{x}$ is a
        setting of the variables in $\vec{X}' = \vec{X}-\{C\}$ that agrees with
        $\vec{z}$ on the variables in $\Pa_1(C)$.  Then, for all
        contexts $\vec{u}_1$ for $M_1$ and $\vec{u}_2$ for
    $M_2$ such that $\vec{u}_1$ and $\vec{u}_2$ agree on the variables
    in $\U_1 \cap \U_2$, we have 
    $F_C^1(\vec{z}) = c$ iff $(M_1, \vec{u}_1) \satt [\vec{X}'
      \gets \vec{x}](C=c)$ iff $(M_2, \vec{u}_2) \satt [\vec{X}'
%joe39
%      \gets \vec{x}](C=c)$ (by MI4$_{M_1,M_2,C}$) iff $F_C^2(\vec{z})
      \gets \vec{x}](C=c)$ (by MI4$_{M_1,M_2,C,\vec{v}^*}$) iff $F_C^2(\vec{z})
      = c$.  Thus, $\F_1(C) = \F_2(C)$.

%joe35: added lots of \vec{v}^*
    For part (b), note that $M_1 \oplus^{\vec{v}^*} M_2$ is well defined unless
    (i) $\R_1(C) \ne \R_2(C)$ for some $C \in ((\U_1 \cup \V_1) \cap
    (\U_2 \cup \V_2))$ or (ii) $M_1 \sim_C^{\vec{v}^*} M_2$ but $\F_1(C) \ne
    \F_2(C)$ for some $C \in \V_1 \cap \V_2$.  Since $M_1$ and $M_2$
    %dalal23
    % are compatible, (i) cannot happen; by part (a), (ii) cannot happen.
%joe35
%    are fully compatible, (i) cannot happen; by part (a), (ii) cannot happen.
  are fully compatible with respect to $\vec{v}^*$, (i) cannot happen; by part (a), (ii) cannot happen.

For part (c),
%joe12
%we first show that if $A$ and $B$ are both nodes in $M_1$ (i.e., $A$ and
%dalal9: type
%we first part (d): if $A$ and $B$ are both nodes in $M_1$ (i.e., $A$ and
we first show part (d): if $A$ and $B$ are both nodes in $M_1$ (i.e., $A$ and
$B$ are in $\U_1 \cup \V_1$), then (the node labeled) $A$ is an
ancestor of (the node labeled) $B$ in (the causal graph corresponding
to) $M_1$ iff  $A$ is an ancestor of $B$ in $M_1 \oplus^{\vec{v}^*} M_2$, and
similarly for $M_2$.  

%dalal3: change \F => \F_{1,2}
Suppose that $A$ is an ancestor of $B$ in $M_1$,  Then there is a
finite path $A_0, \ldots, A_n$ in the causal graph for $M_1$,
where $A_0 = A$ and $A_n = B$.
%joe7: rewrote argument slightly
%Without loss of generality, we can
%suppose that this is the path that includes the most nodes that are
%also in in $M_2$.  
We first show that if $A_0, \ldots, A_n$ is an arbitrary sequence of nodes in
$M_1$ such that none of the 
intermediate nodes (i.e., $A_1, \ldots,
A_{n-1}$) are in $M_2$, and either $A_0 = A_n$ or at most
one of $A_0$ and $A_n$ 
is in $M_2$, then $A_0, \ldots, A_n$ is a path in $M_1$ iff 
$A_0, \ldots, A_n$ is a path in $M_1 \oplus^{\vec{v}^*} M_2$.
We proceed by induction on $n$, the length of the path.
Since all the nodes in $M_1$ are nodes in $M_1 \oplus^{\vec{v}^*} M_2$, the result
clearly holds if $n=0$.  Suppose that $n > 0$ and the result holds for
$n-1$; we prove it for $n$.
We cannot have $A_n \in \U_1 - \V_2$, since then
$A_n$ has no parents in $M_1$ or $M_1 \oplus^{\vec{v}^*} M_2$.
If $A_n \in \V_1 - \V_2$ or $A_n \in \V_1
\cap \V_2$ and $M_1 \succeq_{A_n}^{\vec{v}^*} M_2$, then $\F_{1,2}(A_n) = \F_1(A_n)$,
so the parents of $A_n$ in $M_1$ are also the
parents of $A_n$ in $M_1 \oplus^{\vec{v}^*} M_2$.  In particular, $A_{n-1}$ is a
parent of $A_n$ in $M_1 \oplus^{\vec{v}^*} M_2$ iff $A_{n-1}$ is a 
parent of $A_n$ in $M_1 \oplus^{\vec{v}^*} M_2$, and the result follows from the
induction hypothesis.  Finally, if $A_n \in (\U \cup \V_1) \cap \V_2$
and $M_2 \succeq_{A_n}^{\vec{v}^*} M_1$, then $\F_{1,2}(A_n) = \F_2(A_n)$,
so $A_{n-1}$ must be in $M_2$.  But this contradicts our assumption,
that no intermediate nodes are in $M_2$ and at most one of $A_0$ and
$A_n$ is in $M_2$.  This completes the argument.  Note that the same
argument applies if we reverse the roles of $M_1$ and $M_2$.

%dalal3: typos
Now suppose that there are $m > 0$ nodes in $M_2$ on the path
from $A$ to $B$ in $M_1$, say $C_1, \ldots, C_m$, in that order.  
We show that (i) $C_m$ is an ancestor of $B$ in $M_1 \oplus^{\vec{v}^*} M_2$, (ii) $A$ is an
ancestor of $C_1$ in $M_1 \oplus^{\vec{v}^*} M_2$, and (iii) $C_1$ is an ancestor of
$C_m$ in $M_1 \oplus^{\vec{v}^*} M_2$.  Parts (i) and (ii) follow from the earlier
argument, since there are no intermediate nodes in $M_2$ on the path
from $C_m$ to $B$ or on the path from $A$ to $C_1$. So it remains to
prove part (iii).  We proceed by induction on $m$.  If $m=1$, the result
is trivially true, since $C_1$ is a node in $M_1 \oplus^{\vec{v}^*} M_2$.  So
%dalal23
%suppose that $m > 1$.  Since $M_1$ and $M_2$ are compatible and $C_2$
%joe35
%suppose that $m > 1$.  Since $M_1$ and $M_2$ are fully compatible and $C_2$
suppose that $m > 1$.  Since $M_1$ and $M_2$ are fully compatible with
respect to $\vec{v}^*$ and $C_2$
is a node in both $M_1$ and $M_2$ for $j > 1$, we must have either
$M_1 \succeq_{C_2}^{\vec{v}^*} M_2$ or $M_2 \succeq_{C_2}^{\vec{v}^*} M_1$.
%joe7: simpler argument
%By MI2, it follows that $C_{1}$ is an ancestor of $C_2$ in
%$M_2$.  This means that there is a path from $C_{1}$ to $C_2$ in
%$M_2$.  Moreover, we claim that none of the intermediate nodes on the
%path can be in $M_1$.  For suppose that there is a path from $C_1$ to
%$C_2$ in $M_2$ going through a node $D$ in $M_1$.  Then, applying MI2
%twice, $D$ is 
%an ancestor of $C_2$ in $M_1$ and $C_1$ is an ancestor of $D$ in $M_1$.  Thus,
%there is a path from $A$ to $B$ in $M_1$ with $m+1$ nodes in $M_2$
%(namely, $C_1, D, C_2, \ldots, C_m$),
%contradicting our assumption that $A_0$ to $A_n$ is the path from $A$
%to $B$ in $M_1$ with the most nodes in $M_2$.  Now we argue that $C_1$
%is an ancestor of $C_2$ in $M_1 \oplus M_2$.  There are two cases.
%If $M_1 \succeq_{C_2}^{\vec{v}^*} M_2$, then $\F(C_2) = \F_1(C_2)$, so
%the parent $D$ of $C_2$ on the path from $C_1$ to $C_2$ is the parent of
%$C_2$ in $M_1 \oplus M_2$.  Since there is a path from $C_1$ to $D$ in
%$M_1$ with no intermediate nodes in $M_2$, our original argument shows
%that $C_1$ is a parent of $D$ in $M_1 \oplus M_2$.  It follows that
%$C_1$ is a parent of $C_2$ in $M_1 \oplus M_2$.  A symmetric argument
%shows that $C_1$ is a parent of $C_2$ in $M_1 \oplus M_2$ if
%$M_2 \succeq_{C_2}^{\vec{v}^*} M_1$.
If $M_1 \succeq_{C_2}^{\vec{v}^*} M_2$ then the parents of $C_2$ in
$M_1$ are the parents of $C_2$ in $M_1 \oplus^{\vec{v}^*} M_2$.  In particular, if
$D$ is the parent of $C_2$ on the path from $C_1$ to $C_2$ in $M_1$, then $D$
is a parent of $C_2$ in $M_1 \oplus^{\vec{v}^*} M_2$.  Since none of the
intermediate nodes on the path from $C_1$ to $D$ in $M_1$
are in $M_2$ except 
for $C_1$, it follows by our earlier argument than the path from $C_1$
to $D$ in $M_1$ is also a path 
from $C_1$ to $D$ in $M_1 \oplus^{\vec{v}^*} M_2$.  Thus, $C_1$ is an ancestor of
$C_2$ in $M_1 \oplus^{\vec{v}^*} M_2$.
If $M_2 \succeq_{C_2}^{\vec{v}^*} M_1$, then the parents of $C_2$ in
$M_1$ must also be in $M_2$ (in fact, they must be $M_1$-immediate
ancestors of $C_2$ in $M_2$).  Since none of the intermediate nodes on
the path from $C_1$ to $C_2$ is in $M_2$, it must be the case that the
path from $C_1$ to $C_2$ has length 1, and $C_1$ is a parent of $C_2$
in $M_1$.  By MI1$_{M_2,M_1,C_2}$, there is a path from $C_1$ to $C_2$
in $M_2$ none of whose intermediate nodes is in $M_1$.  Then the same
argument given for the case that $M_1 \succeq_{C_2}^{\vec{v}^*}
M_2$ shows that this path in $M_2$ also exists in $M_1\oplus^{\vec{v}^*} M_2$.
Thus, $C_1$ is an ancestor of $C_2$ in $M_1 \oplus^{\vec{v}^*} M_2$ in this case
as well.  The fact that $C_2$ is ancestor
of $C_m$ in $M_1 \oplus^{\vec{v}^*} M_2$ follows from the induction
hypothesis.  Thus, $C_1$ is an ancestor of $C_m$ in $M_1 \oplus^{\vec{v}^*} M_2$.

For the converse, suppose that $A$ and $B$ are nodes in $M_1$ and $A$
is an ancestor of $B$ in $M_1 \oplus^{\vec{v}^*} M_2$.  We want to show that $A$
is an ancestor of $B$ in $M_1$.  The argument is similar to that
above, but slightly simpler.   Again, there is a finite path $A_0,
\ldots, A_n$ in the 
causal graph for $M_1 \oplus^{\vec{v}^*} M_2$, where $A_0 = A$ and $A_n = B$.  If 
none of the intermediate nodes on the path are in $M_2$ and at most
one of $A_0$ and $A_n$ is in $M_2$, then our initial argument shows
that this path also exists in $M_1$.

Now suppose that there are $m > 0$ nodes in $M_2$ on the path
from $A$ to $B$ in $M_1 \oplus^{\vec{v}^*} M_2$, say $C_1, \ldots, C_m$, in that order.  
Much like before, 
we show that (i) $C_m$ is an ancestor of $B$ in $M_1$, (ii) $A$ is an
ancestor of $C_1$ in $M_1$, and (iii) $C_1$ is an ancestor of
$C_m$ in $M_1$.  And again, parts (i) and (ii) follow from the earlier
argument, since there are no intermediate nodes in $M_2$ on the path
from $C_m$ to $B$ or the path from $A$ to $C_1$. For part (iii),
we again proceed by induction on $m$.  If $m=1$, the result
is trivially true.  So suppose that $m > 1$.  Since $M_1$ and $M_2$
%dalal23
%are compatible and $C_2$ is a node in both $M_1$ and $M_2$ for $j >
%joe35
%are fully compatible and $C_2$ is a node in both $M_1$ and $M_2$ for $j >
are fully compatible with respect to $\vec{v}^*$ and $C_2$ is a node
in both $M_1$ and $M_2$ for $j > 
1$, we must have either 
$M_1 \succeq_{C_2}^{\vec{v}^*} M_2$ or $M_2 \succeq_{C_2}^{\vec{v}^*}
M_1$.  If $M_1 \succeq_{C_2}^{\vec{v}^*} M_2$, then the parents of
$C_2$ in $M_1$ are just the parents of $C_2$ in $M_1 \oplus^{\vec{v}^*} M_2$, so
if $D$ is the parent of $C_2$ on the path from $C_1$ to $C_2$ in $M_1
\oplus^{\vec{v}^*} M_2$, $D$ is a parent of $C_2$ in $M_1$.  Since the path from
$C_1$ to $D$ in $M_1 \oplus^{\vec{v}^*} M_2$ has no intermediate nodes in $M_2$,
we can apply 
earlier argument to show that there is a path from $C_1$ to $D$ in
$M_1$, and complete the proof as before.  If $M_2
\succeq_{C_2}^{\vec{v}^*} M_1$, then all the parents of $C_2$ in $M_1
\oplus^{\vec{v}^*} M_2$ must be in $M_2$, so the path has length 1 and $C_1$ is a
parent of $C_2$ in $M_1 \oplus^{\vec{v}^*} M_2$ and in $M_2$.
Thus, $C_1$ is an immediate $M_1$-ancestor of $C_2$ in $M_2$.
MI1$_{M_2,M_1,C_2}$
implies that $C_1$ must be a parent of $C_2$ in $M_1$.  Again, we
can complete the proof as before.

%dalal3: update \F => \F_{1,2}
%joe12: this completes the proof of part (d).
The acyclicity of $M_1 \oplus^{\vec{v}^*} M_2$ is now almost immediate.  For suppose that
there is a cycle $A_0, \ldots, A_n$ in the causal graph for $M_1
%joe8: we don't need to worry about n=0; it's not a cycle then
%\oplus^{\vec{v}^*} M_2$, where $A_0 = A_n$.  We cannot have $n=0$, since either
%$\F_{1,2}(A_0) 
%= \F_1(A_0)$ or $\F_{1,2}(A_0) = \F_2(A_0)$, and either way, $A_0$ cannot be
%its own parent.  If $n> 0$, then either $A_n$ and $A_{n-1}$ are both
\oplus^{\vec{v}^*} M_2$, where $A_0 = A_n$ and $n> 0$. Either $A_n$ and $A_{n-1}$ are both
in $M_1$ (if $\F_{1,2}(A_n) = \F_1(A_n)$) or they are both in $M_2$ (if
$\F_{1,2}(A_n) = \F_2(A_n)$).  Suppose that they are both in $M_1$.  Then,
since $A_{n-1}$ is an ancestor of $A_n$ in $M_1 \oplus^{\vec{v}^*} M_2$ and $A_n$
is an ancestor of $A_{n-1}$ in $M_1 \oplus^{\vec{v}^*} M_2$, by the preceding
argument, $A_{n-1}$ is an ancestor of $A_n$ in $M_1$ and $A_n$
is an ancestor of $A_{n-1}$ in $M_1$, contradicting the acyclicity of
$M_1$.  A similar argument applies if both $A_{n-1}$ and $A_n$ are in
$M_2$.

%joe12    
%For part (d), suppose that $\vec{u}$ and $\vec{u}_1$ agree on the
For part (e), suppose that
%joe39
%$\vec{u}$ and $\vec{u}_1$ agree on the
%  variables in $\U_1  \cap \U_2$, $\vec{u}$ agrees with $\vec{v}^*$ on
%the variables in $\U - (\U_1 \cap \U_2)$, and $\vec{u}_1$ agrees with
%$\vec{v}^*$ on the variables in $\U_1 - \U_2$.
$\vec{u}_1$ and $\vec{u}_2$ are compatible with $\vec{v}^*$.
It clearly suffices to
  show that $(M_1, \vec{u}_1) \satt \varphi$ iff $(M_1 \oplus^{\vec{v}^*} M_2,
  \vec{u}) \satt \varphi$ if $\varphi$ has the form $[\vec{X} \gets
    \vec{x}](Y=y)$, where $(\vec{X} \cup \{Y\}) \subseteq \V_1$.  To
    show this, it suffices to show that
    $((M_1)_{\vec{X}= \vec{x}},\vec{u}_1) \satt (Y=y)$ iff
    $((M_1 \oplus^{\vec{v}^*} M_2)_{\vec{X}= \vec{x}},\vec{u}) \satt (Y=y)$.
Define the \emph{depth} of a variable $Y$ in a causal graph to be the
length of the longest path from an exogenous variable to $Y$ in the
graph.
    We prove, by induction on the depth of the variable $Y$ in the
    causal graph of $M_1 \oplus^{\vec{v}^*} M_2$, that for all contexts
    $\vec{u}_1$ in $M_1$, $\vec{u}_2$ in $M_2$, and $\vec{u}$ in $M_1
    \oplus^{\vec{v}^*} M_2$,
    (i) if $Y \in \U_1 \cup \V_1$, $\vec{X} \subseteq \V_1$,
%joe39
%    $\vec{u}$ and $\vec{u}_1$ agree on the
%  variables in $\U_1  \cap \U_2$, $\vec{u}$ agrees with $\vec{v}^*$ on
%the variables in $\U - (\U_1 \cap \U_2)$, and $\vec{u}_1$ agrees with
    %$\vec{v}^*$ on the variables in $\U_1 - \U_2$, then
    and $\vec{u}$ and $\vec{u}_1$ are compatible with $\vec{v}^*$, then
      $((M_1)_{\vec{X}= \vec{x}},\vec{u}_1) \satt (Y=y)$ iff
    $((M_1 \oplus^{\vec{v}^*} M_2)_{\vec{X}= \vec{x}},\vec{u}) \satt (Y=y)$, and
(ii) if $Y \in \U_2 \cup \V_2$, $\vec{X} \subseteq \V_2$,
%joe39
%    $\vec{u}$ and $\vec{u}_2$ agree on the
%  variables in $\U_1  \cap \U_2$, $\vec{u}$ agrees with $\vec{v}^*$ on
%the variables in $\U - (\U_1 \cap \U_2)$, and $\vec{u}_2$ agrees with
    %$\vec{v}^*$ on the variables in $\U_2 - \U_1$, then
    $\vec{u}$ and $\vec{u}_2$ are compatible with $\vec{v}^*$, then
      $((M_2)_{\vec{X}= \vec{x}},\vec{u}_2) \satt (Y=y)$ iff
  $((M_1 \oplus^{\vec{v}^*} M_2)_{\vec{X}= \vec{x}},\vec{u}) \satt (Y=y)$.
  (Note that if $Y \in (\U_1 \cup \V_1) \cap (\U_2 \cup \V_2)$, then
  it must satisfy both (i) and (ii).)
  
If $Y$ has depth 0, then $Y$ is an exogenous variable,   and the
result is immediate.  Suppose that $Y$ has depth $d > 0$.  If $Y \in
\V_1 - (\U_2 \cup \V_2)$, then the parents of $Y$ in $M_1 \oplus^{\vec{v}^*} M_2$
are the same as the parents of $Y$ in $M_1$; (i) is then
immediate from the induction hypothesis and (ii) is vacuously true. 
Similarly, if $Y \in  \V_2 - (\U_1 \cup \V_1)$, then (ii) is immediate
from the induction hypothesis and (i) is vacuously true.  
If $Y \in (\U_1 \cup \V_1) \cap (\U_2
\cup \V_2)$ and $M_1 \succeq_Y^{\vec{v}^*} M_2$, then again,
the parents of $Y$ in $M_1 \oplus^{\vec{v}^*} M_2$
are the same as the parents of $Y$ in $M_1$, so (i) is
immediate from the induction hypothesis.  To show that (ii) holds,
fix appropriate contexts $\vec{u}_2$ and $\vec{u}$.  Now the parents
of $Y$ in $M_2$ are the immediate $M_2$-ancestors of $Y$ in $M_1$.
Let $\vec{Z} = \Pa_{M_2}(Y)$.
It follows from the arguments for part (c) that for all $Z \in
\Pa_{M_2}(Y)$, all the paths from $Z$ to $Y$ in $M_1$ also exist in
$M_1 \oplus^{\vec{v}^*} M_2$ and the parents of $Y$ in $M_2$ are exactly the immediate
$M_2$-ancestors of $Y$ in $M_1 \oplus^{\vec{v}^*} M_2$.    That is, $\vec{Z}$
screens $Y$ from all other variables in $M_2$ not only in $M_2$, but
also in $M_1$ and $M_1 \oplus^{\vec{v}^*} M_2$.  Suppose that $((M_2)_{\vec{X} \gets
  {x}}, \vec{u}_2) \satt \vec{Z} = \vec{z}$.  It follows from the
induction hypothesis that
$((M_1 \oplus^{\vec{v}^*} M_2)_{\vec{X}= \vec{x}}, \vec{u}) \satt \vec{Z} =
\vec{z}$.  Let $\vec{W} = ((\U_1 \cup \V_1) \cap (\U_2
   \cup \V_2)) - \{Y\}$.   Let $\vec{w}$ be a setting for $\vec{W}$
   that agrees with $\vec{z}$ on the variables in $\vec{Z}$.  Then
we have the following chain of equivalences:
$$\begin{array}{lll}
&((M_2)_{\vec{X} = \vec{x}},\vec{u}) \satt Y = y\\
\mbox{ iff } & ((M_2)_{\vec{X} = \vec{x}},\vec{u}_2) \satt [\vec{Z} \gets
  \vec{z}](Y = y)\\
\mbox{ iff } & ((M_2)_{\vec{X} = \vec{x}},\vec{u}_2) \satt [\vec{W} \gets
  \vec{w}](Y = y)\\ 
\mbox{ iff } & (M_2,\vec{u}_2) \satt [\vec{W} \gets   \vec{w}](Y = y)\\ 
\mbox{ iff } & (M_1,\vec{u}_1) \satt [\vec{W} \gets   \vec{w}](Y = y)
%joe39
%&\hspace{-45pt}\mbox{[by MI4$_{M_1,M_2,Y}$]}\\
&\hspace{-45pt}\mbox{[by MI4$_{M_1,M_2,Y,\vec{v}^*}$]}\\ 
\mbox{ iff } & (M_1,\vec{u}_1) \satt [\vec{Z} \gets   \vec{z}](Y = y)\\
\mbox{ iff } & ((M_1)_{\vec{Z} = \vec{z}},\vec{u}_1) \satt (Y = y)\\
\mbox{ iff } & ((M_1 \oplus^{\vec{v}^*} M_2)_{\vec{Z} = \vec{z}},\vec{u}_1) \satt
(Y = y) & \hspace{-35pt}\mbox{[already shown]}\\
\mbox{ iff } & ((M_1 \oplus^{\vec{v}^*} M_2),\vec{u}_1) \satt
[\vec{Z} \gets \vec{z}](Y = y)\\
\mbox{ iff } & ((M_1 \oplus^{\vec{v}^*} M_2)_{\vec{X} = \vec{x}},\vec{u}_1) \satt
     [\vec{Z} \gets \vec{z}](Y = y)\\
%     \mbox{[since $\vec{Z}$ screens $Y$ from variables in $M_2$]}\\
\mbox{ iff } & ((M_1 \oplus^{\vec{v}^*} M_2)_{\vec{X} = \vec{x}},\vec{u}_1) \satt
Y = y \\
& \ \ \mbox{[since $(M_1 \oplus^{\vec{v}^*} M_2)_{\vec{X} = \vec{x}},\vec{u}_1)
    \satt \vec{Z} = \vec{z}$]}
\end{array}$$
This completes the proof of (d).

%joe12: 
%Part (e) is immediate from the definitions.
Part (f) is immediate from the definitions.

%dalal3: changed 123 to 1,2,3 and so forth.
%joe12: 
%For part (f), suppose that $M_1 = ((\U_1, \V_1, \R_1), \F_1)$,
For part (g), suppose that $M_1 = ((\U_1, \V_1, \R_1), \F_1)$,
$M_2 = ((\U_2, \V_2, \R_2), \F_2)$,
$M_3 = ((\U_3, \V_3, \R_3), \F_3)$,
%joe20
$M_1 \oplus^{\vec{v}^*} M_2 = ((\U_{1,2}, \V_{1,2}, \R_{1,2}), \F_{1,2})$,
$M_2 \oplus^{\vec{v}^*} M_3 = ((\U_{2,3}, \V_{2,3}, \R_{2,3}), \F_{2,3})$,
$M_1 \oplus^{\vec{v}^*} (M_2 \oplus^{\vec{v}^*} M_3) = ((\U_{1,2,3}, \V_{1,2,3}, \R_{1,2,3}),
\F_{1,2,3})$, and
$(M_1 \oplus^{\vec{v}^*} M_2) \oplus^{\vec{v}^*} M_3 = ((\U_{1,2,3}', \V_{1,2,3}', \R_{1,2,3}'),
\F_{1,2,3}')$.  We want to show that $M_1 \oplus^{\vec{v}^*} (M_2 \oplus^{\vec{v}^*} M_3) = 
(M_1 \oplus^{\vec{v}^*} M_2) \oplus^{\vec{v}^*} M_3$.  It is almost immediate from the
definitions that $\U_{1,2,3} = \U_{1,2,3}'$, $\V_{1,2,3} = \V_{1,2,3}'$,
and $\R_{1,2,3} = \R_{1,2,3}'$.  To show that $\F_{1,2,3} = \F_{1,2,3}'$, we
show that for all variables $C \in \V_{1,2,3}$, $\F_{1,2,3}(C)
=\F_{1,2,3}'(C)$.  We proceed  by cases.  First suppose that 
%joe20: since we now allow to be exogenous
%$C = \V_1 - (\V_2 \cup \V_3)$, then $C \notin \V_{2,3}$, so it is easy
%then $C \notin \V_{2,3}$, so it is easy 
$C$ is in exactly one of the models.  Say, for example,
$C$ is in $M_1$ but not $M_2$ or $M_3$ (i.e.,
$C = (\U_1 \cup \V_1) - (\U_2 \cup \V_2 \cup \U_3 \cup \V_3)$.
If $C \in \U_1$, there is nothing further to prove.  If $C \in \V_1$, 
then it is easy to check that 
$\F_{1,2,3}(C) = \F_{1,2,3}' = \F_{1}(C)$.  
%joe20
%Similarly, if  $C \in \V_2 - (\V_1 \cup \V_3)$, then $\F_{1,2,3}(C) =
%$\F_{1,2,3}(C) = \F_{1,2,3}' = \F_{2}(C)$, and if 
%$C \in \V_3 - (\V_1 \cup \V_2)$, then $\F_{1,2,3}(C) = \F_{1,2,3}' =
%\F_{3}(C)$.
The same argument works if $C$ is just in $M_2$ or just in $M_3$.

%joe20
%If $C \in (\V_1 \cap \V_2) - \V_3$, since $M_1$ and $M_2$
If $C$ is in two of the three models, suppose without loss of
generality that $C$ is in $M_1$ and 
$M_2$ but not $M_3$.  Since $M_1$ and $M_2$
%dalal23
%are compatible, either $M_1 \succeq_{C}^{\vec{v}^*} M_2$ or
%joe35
%are fully compatible, either $M_1 \succeq_{C}^{\vec{v}^*} M_2$ or
are fully compatible with respect to $\vec{v}^*$, either $M_1
\succeq_{C}^{\vec{v}^*} M_2$ or 
$M_2 \succeq_{C}^{\vec{v}^*} M_1$ (or both). If $M_1 \succeq_{C}^{\vec{v}^*} M_2$,
%joe20
then either $C \in \U_1 \in \U_2$, in which case there is nothing
further to prove, or $C\in \V_1$.  In that case,
$\F_{1,2}(C) = \F_1(C)$, so $\F_{1,2,3}(C) = \F_1(C)$.  Since $C
\notin \V_3$, we have $\F_{2,3}(C) = \F_2(C)$. If we also have
$M_2 \succeq_{C}^{\vec{v}^*} M_1$, then by (a), $\F_1(C) = \F_2(C)$,
and it is easy to see that $\F_{1,2,3}'(C) = \F_1(C)$.  Now suppose that
$M_2 \not\succeq_{C}^{\vec{v}^*} M_1$.
%dalal23
%$M_1$ is compatible with $M_2 \oplus M_3$, we must have either
%joe35
%$M_1$ is fully compatible with $M_2 \oplus^{\vec{v}^*} M_3$, we must
$M_1$ is fully compatible with $M_2 \oplus^{\vec{v}^*} M_3$ with
respect to $\vec{v}^*$, we must
have either 
$M_1 \succeq_{C}^{\vec{v}^*} M_2 \oplus^{\vec{v}^*} M_3$ or
$M_2 \oplus^{\vec{v}^*} M_3 \succeq_{C}^{\vec{v}^*} M_1$.  It is easy to see that
since $M_2 \not\succeq_{C}^{\vec{v}^*} M_1$, we cannot have
$M_2 \oplus^{\vec{v}^*} M_3 \succeq_{C}^{\vec{v}^*} M_1$, so we must have
$M_1 \succeq_{C}^{\vec{v}^*} M_2 \oplus^{\vec{v}^*} M_3$.  It follows that
$\F_{1,2,3}'(C) = \F_1(C)$.
%joe20
%The argument is similar if $C \in (\V_1
%\cap V_3) - \V_2$ or $C \in (V_2 \cap V_3) - \V_2$.

%joe20
%Finally, suppose that $C \in (\V_1 \cap \V_2 \cap \V_3)$.
Finally, suppose that $C$ is in all three models.
We first show that $\succeq^{\vec{v}^*}_C$ is transitive when restricted
to $M_1$, $M_2$, and $M_3$.  For suppose that
$M_1 \succeq_{C}^{\vec{v}^*} M_2$  and
$M_2 \succeq_{C}^{\vec{v}^*} M_3$. If
$M_1 \sim_{C}^{\vec{v}^*} M_2$ or
$M_2 \sim_{C}^{\vec{v}^*} M_3$, then it is easy to see
that $M_1 \succeq_{C}^{\vec{v}^*} M_3$.  So suppose that 
$M_1 \succ_{C}^{\vec{v}^*} M_2$  and
$M_2 \succ_{C}^{\vec{v}^*} M_3$.  Since $M_1$ and $M_3$ are
%dalal23 
%compatible, we must have either $M_1 \succeq_{C}^{\vec{v}^*} M_3$ or
%joe35
%fully compatible, we must have either $M_1 \succeq_{C}^{\vec{v}^*} M_3$ or
fully compatible with respect to $\vec{v}^*$, we must have either $M_1 \succeq_{C}^{\vec{v}^*} M_3$ or
$M_3 \succeq_{C}^{\vec{v}^*} M_1$.  Suppose by way of contradiction
that $M_3 \succ_{C}^{\vec{v}^*} M_1$.  Let $\vec{X}_1 =
\Pa_{M_1}(C)$, $\vec{X}_2 = \Pa_{M_2}(C)$, and $\vec{X}_3 =
\Pa_{M_3}(C)$.
We now construct an infinite sequence of variables $A_0,
A_1, \ldots$ such that each variable in the sequence is either in
$\vec{X}_2 - \vec{X}_1$, $\vec{X}_3 - \vec{X}_2$, or $\vec{X}_1 -
\vec{X}_3$, and if variable $A_n$ is in $\vec{X}_i - \vec{X}_j$, then
the next variable is in $\vec{X}_j$ and there is a path in $M_j$ from
$A_n$ to $A_{n+1}$.  We proceed by induction.
Since $M_1 \succ_{C}^{\vec{v}^*} M_2$, by MI1$_{M_1,M_2,C}$
there must be at least one variable in $A_0 \in \vec{X}_2 - \vec{X}_1$ and a
path from $Z_1$ to $C$ in $M_1$ that does not go through any other
variables in $\vec{X}_2$.  Since $\vec{X}_1$ screens $C$ from all
ancestors in $M_1$, this path must go through a variable $A_1 \in
\vec{X}_1 - \vec{X}_2$.  If $A_1 \in \vec{X}_3$, then it is in
$\vec{X}_3 - \vec{X}_2$; if $A_1 \notin \vec{X}_3$, it is in
$\vec{X}_1 - \vec{X}_3$.  Either way, $A_1$ is an appropriate
successor of $A_0$ in the sequence.  The inductive step of the
argument is identical; if $A_n \in \vec{X}_i - \vec{X}_j$, we use the
fact that $M_j \succ_{C}^{\vec{v}^*} M_i$ to construct $A_{n+1}$.
Note that, for all $n \ge 0$, since $A_n \in \vec{X}_i - \vec{X}_j$
and $A_{n+1} \in \vec{X}_j$, we must have $A_n \ne A_{n+1}$.
Moreover, by the argument in the proof of (c) since there is a path from $A_n$
to $A_{n+1}$ in $M_j$, there must also be such a path in $M_1 \oplus^{\vec{v}^*}
(M_2 \oplus^{\vec{v}^*} M_3)$.  Since there are only finitely many variables
altogether, there must be some $N_1$ and $N_2$ such that $A_{N_1} =
A_{N_2}$.  That means we have a cycle in $M_1 \oplus^{\vec{v}^*} (M_2 \oplus^{\vec{v}^*}
M_3)$, contradicting (c).

%dalal3: typo
Since $\succeq_{C}^{\vec{v}^*}$ is transitive and complete on
$\{M_1,M_2,M_3\}$ (completeness says that for each pair, one of the
two must be dominant), one of $M_1$, $M_2$, and $M_3$ must dominate
the other two with respect to $\succeq_C^{\vec{v}^*}$.  Suppose it is
$M_1$.  It is easy to see that $M_1 \oplus^{\vec{v}^*} M_2 \succeq_C^{\vec{v}^*}
M_3$ and $M_1 \succeq_C^{\vec{v}^*} (M_2 \oplus^{\vec{v}^*} M_3)$.  
It then easily follows that $\F_{1,2,3}(C) = \F_{1,2,3}'(C) = \F_1(C)$.  A
similar argument holds if $M_2$ or $M_3$ is the model that dominates
with respect to $\succeq_C^{\vec{v}^*}$.
%joe33
%$\Box$
\eprf
}

%joe21: moved here
\smallskip
\noindent {\bf Acknowledgments:} We thank Noemie Bouhana, Frederick
Eberhardt, 
%joe17
%and the anonynous reviewers for useful comments.  Joe Halpern's work
%was supported in part by NSF grants IIS-1703846 and IIS-1718108, 
%joe20: since Meir pointed out that we have problems with acyclicity
Meir Friedenberg,
and anonymous reviewers for useful comments.  Joe Halpern's work
was supported by NSF grants IIS-1703846 and IIS-1718108, 
AFOSR grant FA9550-12-1-0040, 
%joe17
%and ARO grant W911NF-17-1-0592. Dalal Alrajeh's work was supported by
ARO grant W911NF-17-1-0592, and the Open Philanthropy project.
%joe17: Are you sure that this is the grant number?  It doesn't look
%right to me as a grant number
%Dalal Alrajeh's work was supported by MRI grant ONR BAA 14-01. 
Dalal Alrajeh's work was supported by MRI grant FA9550-16-1-0516. 
%dalal16
%\vspace{-.05in}
%\input{appendix}
%joe13
%\newpage
%joe1
%hana17
%\bibliographystyle{aaai}
%\bibliography{fv,joe,z,dalal,jk}
%
%\bibliographystyle{theapa}
%dalal22
\bibliographystyle{elsarticle-harv}
\bibliography{fv,joe,z,dalal,jk}

\end{document}